\documentclass{article} 
\usepackage[preprint]{neurips_2026}

\usepackage{amsmath,amsfonts,bm}

\def\eqref#1{equation~\ref{#1}}

\def\1{\bm{1}}

\DeclareMathAlphabet{\mathsfit}{\encodingdefault}{\sfdefault}{m}{sl}
\SetMathAlphabet{\mathsfit}{bold}{\encodingdefault}{\sfdefault}{bx}{n}

\usepackage{hyperref}
\usepackage{url}
\usepackage{amsmath}
\usepackage{amssymb}
\usepackage{amsthm}
\usepackage{mathtools}
\usepackage{graphicx}
\usepackage{subcaption}
\usepackage[most]{tcolorbox}
\usepackage{wrapfig}
\usepackage{xspace}
\usepackage{tabularx}
\usepackage{booktabs}
\usepackage{xcolor}
\usepackage{multirow}
\usepackage[table]{xcolor}

\definecolor{lightgreen}{RGB}{223,240,216}
\definecolor{myblue}{RGB}{31, 119, 180}    
\definecolor{myorange}{RGB}{255, 127, 14}  
\definecolor{mygreen}{RGB}{44, 160, 44}

\newtheorem{theorem}{Theorem}[section]
\newtheorem{lemma}[theorem]{Lemma}

\newtcolorbox{principlebox1}[1][]{
  colback=blue!5!white,
  colframe=blue!60!black,
  fonttitle=\bfseries,
  title=Principle,
  coltitle=black,
  boxed title style={colback=blue!15!white},
  sharp corners,
  enhanced,
  attach boxed title to top left={xshift=0.5cm, yshift=-2mm},
  #1
}

\title{\method: Query-Adaptive KV Cache \\Compression via Attention-Error Bounds
}

\author{Kewei Lian, Jiayi Zhai, Anji Liu \\
National University of Singapore\\
}

\newcommand{\method}{\textsc{RefineKV}\xspace}

\begin{document}

\maketitle

\begin{abstract}
Repeatedly accessing the full key--value (KV) cache is a major bottleneck in long-context language model decoding. We find that small query-specific subsets can accurately approximate full attention, yet their union covers nearly the entire cache, motivating query-adaptive access for the KV cache. A natural approach is to split the KV cache into clusters, using compact representatives for clusters that incur negligible error while expanding the rest to their original KV entries. 
We derive a selection criterion from an upper bound on attention-output error, which reveals two complementary factors governing the error: \textit{importance} measures the impact of a cluster, while \textit{distortion} measures how faithfully the cluster can be represented. Based on this criterion, we propose \method, a decoding-time query-adaptive KV compression method. Experiments on two models and three benchmarks demonstrate that \method retains near-full-attention quality while achieving up to $7.6\times$ speedup in the attention phase and $1.66\times$ faster decoding, establishing a new quality–latency Pareto frontier.\footnote{Project page: \url{https://kevin-lkw.github.io/RefineKV-release/}}
\end{abstract}

\section{Introduction}
Context has emerged as a fundamental resource for modern language model inference in both reasoning LLMs~\citep{deepseekr1} and interactive agent systems~\citep{yao2023reactsynergizingreasoningacting}. During autoregressive decoding with full attention, each token computes attention weights between its query vector and the key vectors of preceding context tokens, then uses these weights to aggregate their value vectors~\citep{vaswani2023attentionneed}. To avoid redundant computation, the key and value vectors for context tokens are stored as the Key-Value (KV) cache~\citep{li2024snapkvllmknowslooking}. However, repeatedly accessing this massive KV cache at every generation step incurs severe overhead, making attention a substantial decoding bottleneck in long-context regimes~\citep{tang2024questqueryawaresparsityefficient}.

KV cache compression aims to reduce the amount of data accessed by every decoding query while minimizing the deviation of compressed output from full attention output, typically through eviction~\citep{zhang2023h2oheavyhitteroracleefficient,li2024snapkvllmknowslooking,xiao2024efficientstreaminglanguagemodels,oren2024tovatransformersmultistaternns}, merging~\citep{wang2024modeltellsmergeadaptivekvmerger,CAM} or quantization~\citep{kivi}. To examine the potential for reducing KV access, we ask a key question: how much of the KV cache is necessary for each query to approximate attention output with high fidelity? Our findings reveal that a relatively small query-specific subset is already sufficient. However, these subsets vary substantially across queries, and their union covers nearly the entire KV cache. This gap reveals that different queries attend to sparse but different parts of the KV cache, which suggests that, rather than compressing the KV cache into a fixed subset for all future queries, the accessed subset should adapt to the current query~\citep{tang2024questqueryawaresparsityefficient,hooper2025multipoleattentionefficientlong,lu2025mobamixtureblockattention,ni2026doublephierarchicaltoppsparse,colaco2026keepforgetratedistortionview}. 

Query-adaptive access raises a central challenge: given a query, how can we decide which part should be accessed without reading the full KV cache? We move access decisions to a coarser granularity, using a compact summary to assess a group of KV entries without inspecting its individual members. Specifically, we partition the cache into clusters and precompute a compact summary for each cluster to guide query-dependent access. Selected clusters are expanded to expose their original KV entries, while the remaining clusters are approximated by compact representatives. Within this paradigm, the next question is: how can we identify clusters whose representative approximation introduces negligible attention-output error?

We address this question with a cluster-wise risk score. Starting from a decomposition of the attention-output error, we derive an upper bound on the overall approximation error and use its cluster-wise terms as risk scores. The cluster-wise score captures the joint effect of \textit{global importance}, measured by a cluster's approximate attention mass, and \textit{local distortion}, characterized by the spread of its attention logits under the current query. Under this criterion, high-importance clusters require tighter local approximation to remain within the same risk tolerance, while lower-importance clusters can tolerate greater distortion. Empirically, the resulting score correlates strongly with the actual attention-output error caused by cluster approximation.

Based on this risk score, we propose \method, a query-adaptive KV
compression method for efficient decoding. After prefill, \method
partitions the KV cache into contiguous clusters and precomputes their
representatives and auxiliary metadata for computing risk score, while retaining the original KV entries for selective expansion. At each decoding step, cluster risk scores are computed from the current query and the
cached summaries. Clusters whose risk score exceeds a predefined threshold are expanded to their original KV entries, while the remaining clusters are each compressed into a single representative. An attention operation over this mixed representation produces the approximate output. Experiments on two models and three benchmarks demonstrate that \method retains near-full-attention quality while achieving up to $7.6\times$ speedup in the attention phase and $1.66\times$ faster decoding, establishing a new quality–latency Pareto frontier.
\section{Motivation: From Eviction to Query-Adaptive Access}

Autoregressive Transformers generate each token by conditioning on all previous tokens.
Within an attention head, a query vector \(\boldsymbol{q} \in \mathbb{R}^d\) representing the current decoding position interacts with the key vectors \(\boldsymbol{k}_i \in \mathbb{R}^d\) of the context tokens at position \(i\) to determine how their value vectors \(\boldsymbol{v}_i \in \mathbb{R}^{d_v}\) are aggregated. For a context of \(N\) tokens and query \(\boldsymbol{q}\), the full-attention output \(\boldsymbol{o} \in \mathbb{R}^{d_v}\):
\begin{equation}
\label{eq:full-attention}
\boldsymbol{o}=\sum_{i=1}^{N}\alpha_i\boldsymbol{v}_i, \qquad 
\alpha_i= \frac{\exp\!\left(\boldsymbol{q}^\top \boldsymbol{k}_i/\sqrt d\right)} {\sum_{j=1}^{N}\exp\!\left(\boldsymbol{q}^\top \boldsymbol{k}_j/\sqrt d\right)},
\end{equation}
During decoding, the keys and values of previous tokens are reused across generation steps. The Key-Value cache (KV cache) stores these representations as \(\{(\boldsymbol{k}_i,\boldsymbol{v}_i)\}_{i=1}^{N}\), avoiding repeated computation. However, the cache grows linearly with context length \(N\), and full attention reads the entire cache for every query, which still incurs substantial decoding latency in long-context regimes. To mitigate this, KV cache compression methods~\citep{li2025surveykvlargelanguagemodel,zhang2023h2oheavyhitteroracleefficient,xiao2024efficientstreaminglanguagemodels} aim to reduce KV data accessed by each query while minimizing the deviation of the compressed output $\boldsymbol{\hat{o}}$ from full attention output $\boldsymbol{o}$ in Equation~\ref{eq:full-attention}.

\begin{figure}
    \centering
    \includegraphics[width=\linewidth]{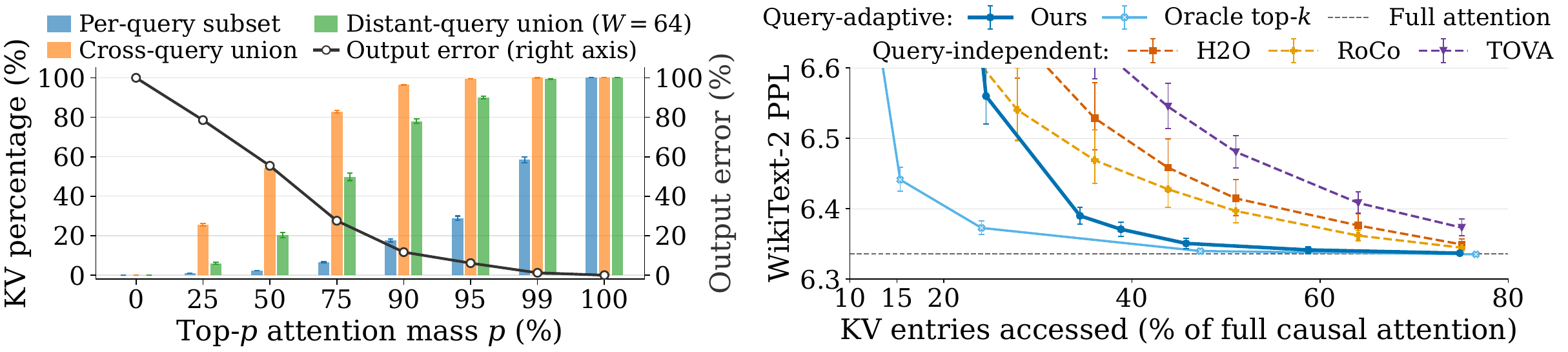}
    \caption{\textbf{Left: Each query needs only a small KV subset, but together these subsets cover nearly the entire cache.} For each query, we select the highest-weight KV entries needed to retain $p$\% of its total attention weight. The figure compares sizes of these per-query subsets and their union across queries, alongside the resulting attention output error. At roughly 10\% output error, each query needs only a small subset of the cache, yet their union covers most KV entries. \textbf{Right: Query-adaptive methods surpass query-independent methods.} 
    Oracle query-adaptive top-\(k\) attention achieves a markedly better perplexity--access trade-off than representative eviction baselines. Our method recovers a substantial portion of this advantage without oracle access.}
    \label{fig:query-adapt}
\end{figure}

A fundamental question is: \textit{For any query, how much of the KV cache needs to be accessed to accurately approximate full attention?} 
To examine this, we select the highest-weight KV entries for each query until their cumulative attention mass reaches $p$\%. Figure~\ref{fig:query-adapt} (left) shows how the percentage of KVs selected (bars) and the attention output error (line) vary with the amount of attention mass retained. The \textbf{\textcolor{myblue}{blue bars}} show the average fraction of KV items retained for each query. At an attention mass of $p = 90\%$, retaining merely $17.6\%$ of the KV cache per query is sufficient to keep the approximation error down to $10\%$. This suggests that only a relatively small subset is crucial for accurately approximating the output for a particular query.

However, compressing the cache into a fixed subset requires that the same subset remains sufficient to accurately approximate the full-attention output across different queries. This raises a second question: \textit{Are these crucial subsets shared across queries, such that a fixed subset of KV pairs could serve all future queries?} 
The \textbf{\textcolor{myorange}{orange bars}} show the size of union of these query-specific subsets. Although an individual query requires only 17.6\% of the cache at $p = 90\%$, their union spans 96.4\% of the KV cache. One potential confounding factor is attention locality: each query tends to attend strongly to its immediately preceding tokens. Thus, the union may appear large even under a purely local attention pattern. To control for this effect, we remove the most recent 64 tokens for each query before computing the union. The gap remains substantial, as shown by the \textbf{\textcolor{mygreen}{green bars}}. This gap shows that different queries rely on different KV pairs, therefore a fixed subset of cache simply cannot work for all queries. Detailed experimental setups are provided in Appendix~\ref{app:kvimportant}.

These observations suggest a rethinking in the compression paradigm: \textbf{rather than committing to a fixed subset of KV entries for all future queries, the accessed subset should adapt to the current query.} We next examine whether such query-adaptive access indeed translates into a better end-to-end quality--efficiency trade-off. Figure \ref{fig:query-adapt} (right) illustrates a perplexity evaluation on WikiText-2. We consider three query-independent methods, H2O~\citep{zhang2023h2oheavyhitteroracleefficient}, RoCo~\citep{ren2024rocoefficacyevictionpolicykeyvalue}, and TOVA~\citep{oren2024tovatransformersmultistaternns}, and contrast them with query-adaptive approaches: oracle top-\(k\) selection, and our practical query-adaptive approach introduced in the following section. The query-independent methods exhibit a substantially less favorable quality–budget trade-off. This provides end-to-end evidence supporting query-adaptive access as a more effective compression paradigm.
\section{Deciding What to Access: From Attention Error to Risk Score}
Query-adaptive access leaves a central question: which KV entries
should be accessed for the current query? Answering this question
involves two requirements: efficiency and accuracy. \textit{Efficiency} requires making access decisions without scanning every token-level KV entry, which would otherwise defeat the purpose of compression. \textit{Accuracy} requires selecting tokens that incur minimal output deviation from full attention.  

\subsection{Cluster-based Access}
For \textit{efficiency}, as KV entries cannot be visited individually, we need to move access decisions to a coarser granularity, using a compact summary to assess a group of KV entries, to decide whether they should be accessed. A natural realization of this idea is to partition KV entries into clusters, and use only cluster-level summaries to decide which clusters require token-level access. Selected clusters are expanded to compute full attention, while unselected ones are either pruned entirely~\citep{tang2024questqueryawaresparsityefficient,lu2025mobamixtureblockattention} or coarsely approximated by their representative tokens~\citep{hooper2025multipoleattentionefficientlong,ni2026doublephierarchicaltoppsparse}.

In this work, we adopt the representative-based approximation route: clusters whose coarse-grained representation introduces negligible error in the attention output are represented by their compact representatives, whereas those that cannot be faithfully approximated are expanded to full attention. Intuitively, this paradigm allocates a ``high-resolution'' view to clusters that require precise access. 
Then we focus on the \textit{accuracy} requirement, which is: \textit{how can we identify clusters whose approximation incurs negligible error in the attention output?}

\subsection{Upper Bounding Attention Error}
To answer this question, we seek an efficiently computable, query-dependent risk score that can reflect the output error caused by approximating a particular cluster. Specifically, we start from decomposing the single-head attention-output error. Let $\mathcal{C} = \{C_1,...,C_m\}$ be a partition of the KV cache. For each cluster $C$, we use its mean key and value $(\bar{\boldsymbol{k}}_C, \bar{\boldsymbol{v}}_C)$ as a shared representative for every KV entry in $C$.
For a query $\boldsymbol{q}$, under full attention, let $p_C$ denote the total attention weight assigned to cluster $C$, and let $\boldsymbol{v}_C$ denote the attention-weighted average of its token values. Under approximation, the cluster receives total attention weight $\widehat p_C$ and contributes its mean value $\bar{\boldsymbol{v}}_C$. Formally,
\begin{equation} 
p_C = \frac{\sum_{i \in C} e^{\boldsymbol{q}^\top \boldsymbol{k}_i / \sqrt{d}}}{\sum_{j=1}^N e^{\boldsymbol{q}^\top \boldsymbol{k}_j / \sqrt{d}}}, \qquad
\boldsymbol{v}_C = \frac{\sum_{i \in C} e^{\boldsymbol{q}^\top \boldsymbol{k}_i / \sqrt{d}} \boldsymbol{v}_i}{\sum_{j \in C} e^{\boldsymbol{q}^\top \boldsymbol{k}_j / \sqrt{d}}}, \qquad
\widehat{p}_C = \frac{|C| e^{\boldsymbol{q}^\top \bar{\boldsymbol{k}}_C / \sqrt{d}}}{\sum_{C' \in \mathcal{C}} |C'| e^{\boldsymbol{q}^\top \bar{\boldsymbol{k}}_{C'} / \sqrt{d}}}.
\label{eq:def-pc}
\end{equation}

The error between exact attention output $\boldsymbol{o} = \sum_{C \in \mathcal{C}} p_C \boldsymbol{v}_C$ and approximate attention output $\widehat{\boldsymbol{o}}=\sum_{C \in \mathcal{C}} \widehat{p}_C \bar{\boldsymbol{v}}_C$ can be decomposed as:
\begin{equation}
    \boldsymbol{o} - \widehat{\boldsymbol{o}}
    = \sum_{C \in \mathcal{C}} p_C \boldsymbol{v}_C - \sum_{C \in \mathcal{C}} \widehat{p}_C \bar{\boldsymbol{v}}_C
    = \underbrace{
        \sum_{C \in \mathcal{C}} (p_C - \widehat{p}_C) \boldsymbol{v}_C
    }_{\text{inter-cluster mass redistribution}}
    + \underbrace{
        \sum_{C \in \mathcal{C}} \widehat{p}_C (\boldsymbol{v}_C - \bar{\boldsymbol{v}}_C)
    }_{\text{intra-cluster value distortion}}.
\label{eq:output-decomposition}
\end{equation}
This decomposition separates the output error into two sources: The first term captures the redistribution of attention mass across clusters, whereas the second term captures the error caused by approximating $\boldsymbol{v}_C$ with $\bar{\boldsymbol{v}}_C$. After regrouping the first term (Lemma~\ref{lem:source-mass}), both terms are related to the dispersion the logits $\boldsymbol{q}^\top \boldsymbol{k}_i$ within a cluster: when all $\boldsymbol{q}^\top \boldsymbol{k}_i$ are close to $\boldsymbol{q}^\top \bar{\boldsymbol{k}}_C$, the cluster is well-approximated and both terms get close to zero.

Neither error term is directly computable from compact cluster summaries. We therefore bound these two terms using the maximum logit deviation: $\delta_C = \max_{i\in C} |\boldsymbol{q}^\top(\boldsymbol{k}_i-\bar{\boldsymbol{k}}_C)/\sqrt d|$. However, computing $\delta_C$ exactly would require scanning all keys in the cluster. We therefore use an efficiently computable upper bound $\delta_C^{\mathrm{bound}} \geq \delta_C$, constructed from compact cluster summaries (Details in Appendix~\ref{app:deltabound}). Letting $V_C = \max_{i \in C} \lVert \boldsymbol{v}_i \rVert_2$ and $V_{\max} = \max_{C\in \mathcal{C}} V_C$, we have:
\begin{equation}
\lVert \boldsymbol{o}-\widehat{\boldsymbol{o}} \rVert_2
\leq
2 \sum_{C\in\mathcal C}
\underbrace{\widehat p_C \left( V_{\max} \frac{\cosh(\delta_C^{\mathrm{bound}})-1}{\sum_{C'\in\mathcal C} \widehat p_{C'} \cosh(\delta_{C'}^{\mathrm{bound}})}
+ V_C \tanh\left(\frac{\delta_C^{\mathrm{bound}}}{2}\right) \right)}_{\text{Risk Score $R_C$ for cluster \(C\)}}.
\label{eq:risk-score}
\end{equation}
In Eq.~\ref{eq:risk-score}, we write the resulting upper bound in a clusterwise form and use each clusterwise term as its risk score $R_C$. The detailed derivation is provided in Appendix~\ref{app:derivation}. In the following section, we provide an analysis of this risk score: what it actually measures and whether it faithfully reflects the output error.

\subsection{Importance and Distortion}
\begin{figure}
    \centering
    \includegraphics[width=\linewidth]{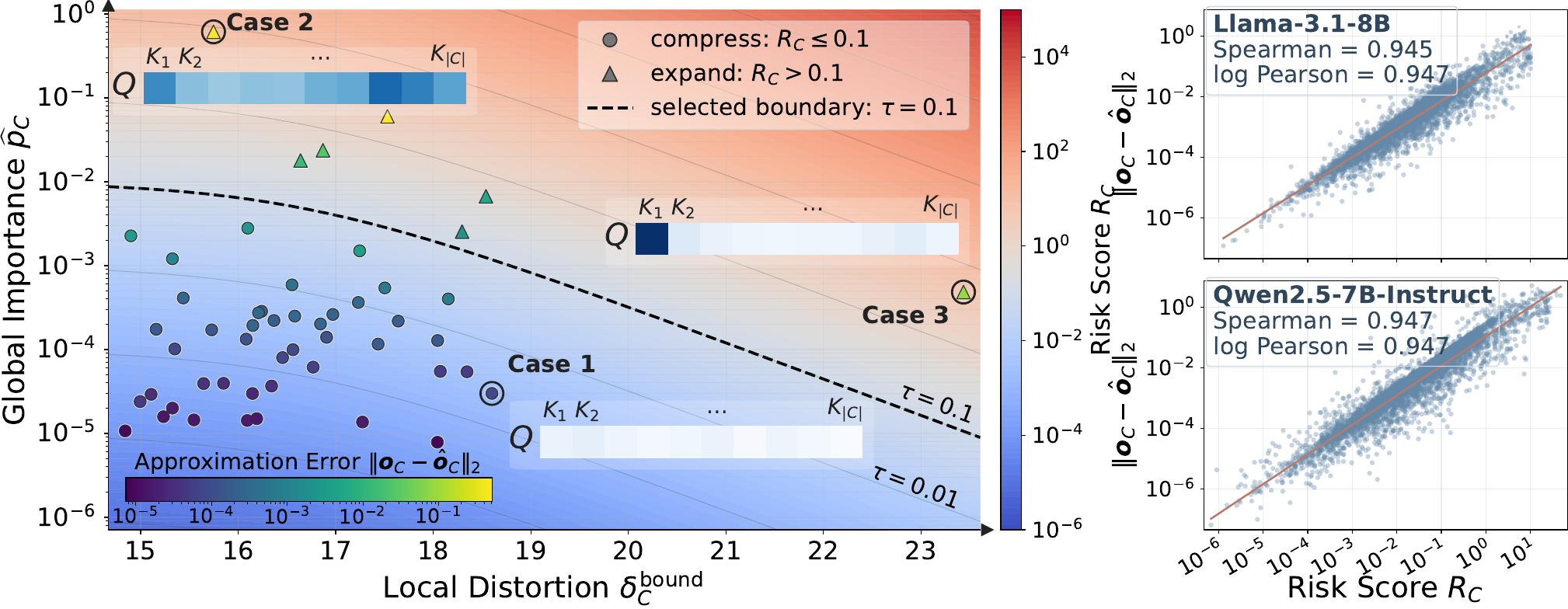}
    \caption{
    \textbf{Left: Cluster selection is jointly determined by importance and distortion.}
    Each point represents a cluster positioned by its global importance $\widehat{p}_C$ and local distortion $\delta_C^{\mathrm{bound}}$. Background color encodes the risk score $R_C$, with the dashed line denoting the decision boundary at $R_C = \tau$. The tilted boundary captures their trade-off: important clusters tolerate only small local distortion, whereas low-importance clusters can tolerate larger distortion. Point colors indicate the error incurred on approximating cluster. Three cases show the within-cluster attention-score patterns.
    \textbf{Right: Risk score $R_C$ strongly correlated with approximation error.} These figures illustrate the correlation between Risk Score and cluster approximation error $\lVert \boldsymbol{o}_C - \hat {\boldsymbol{o}}_C\rVert_2$, where $\boldsymbol{o}_C$ and $\hat{\boldsymbol o}_C$ denote original output and approximate output for cluster $C$.
    }
    \label{fig:main}
\end{figure}
The risk score derived from Eq.~\ref{eq:risk-score} captures two query-dependent factors that influence the approximation error: global importance, measured by approximated cluster attention mass \(\widehat p_C\), and local distortion, measured by $\delta_C$ and upper-bounded by $\delta_C^{\mathrm{bound}}$. 

\textbf{Global Importance \(\widehat p_C\).} Approximation error depends on how much the cluster contributes to the overall attention computation. If a cluster receives only a negligible fraction of the attention mass relative to the rest of the context, even a relatively coarse approximation may have little effect on the final attention output.

\textbf{Local Distortion $\delta_C^{\mathrm{bound}}$.} For a given query, a cluster is easy to approximate when its tokens behave similarly under the query. In particular, if all $\boldsymbol{q}^\top \boldsymbol{k}_i$ are close to $\boldsymbol{q}^\top \bar{\boldsymbol{k}}_C$, replacing them by the representative introduces little error. Importantly, this distortion is query-dependent: a cluster may be dispersed in the original key space while remaining tightly concentrated along the direction relevant to the current query.

Crucially, the score captures the joint effect of both factors. A highly important cluster can only tolerate a small local approximation error, whereas a low-importance cluster can tolerate substantially larger internal variation. Figure~\ref{fig:main} (Left) illustrates a sample from real data: Each point represents a cluster positioned by Local Distortion $\delta_C^{\mathrm{bound}}$ and Global Importance $\widehat p_C$. The background color indicates the Risk score, and the tilted decision boundary $R_C = \tau$ adaptively balances $\widehat{p}_C$ against $\delta_C^{\mathrm{bound}}$.

Beyond this theoretical intuition, the next question is whether risk score $R_C$ faithfully reflects the true output error caused by approximation. Figure~\ref{fig:main} (Right)  plots the risk score against the cluster approximation error $\lVert \boldsymbol{o}_C - \hat {\boldsymbol{o}}_C\rVert_2$, where $\boldsymbol{o}_C$ and $\hat{\boldsymbol o}_C$ denote original output and approximate output for cluster $C$. The results demonstrate a clear, robust correlation across different model architectures, confirming that our theoretical bound serves as an effective surrogate for actual approximation error. Appendix \ref{app:correlation} reports the partial correlation of score with the cluster error given $\hat p_C$.

\section{\method: Efficient Risk-Guided KV Access}
\label{sec:method}
The preceding analysis provides a principled Risk Score for deciding which clusters require token-level access. Turning this criterion into an efficient decoding method requires both making access decisions without scanning the full KV cache and ensuring that these decisions reduce actual KV reads. We address these requirements with \method, a query-adaptive KV access method that preserves the full cache while selectively accessing its original entries.

\begin{figure}
    \centering
    \includegraphics[width=\linewidth]{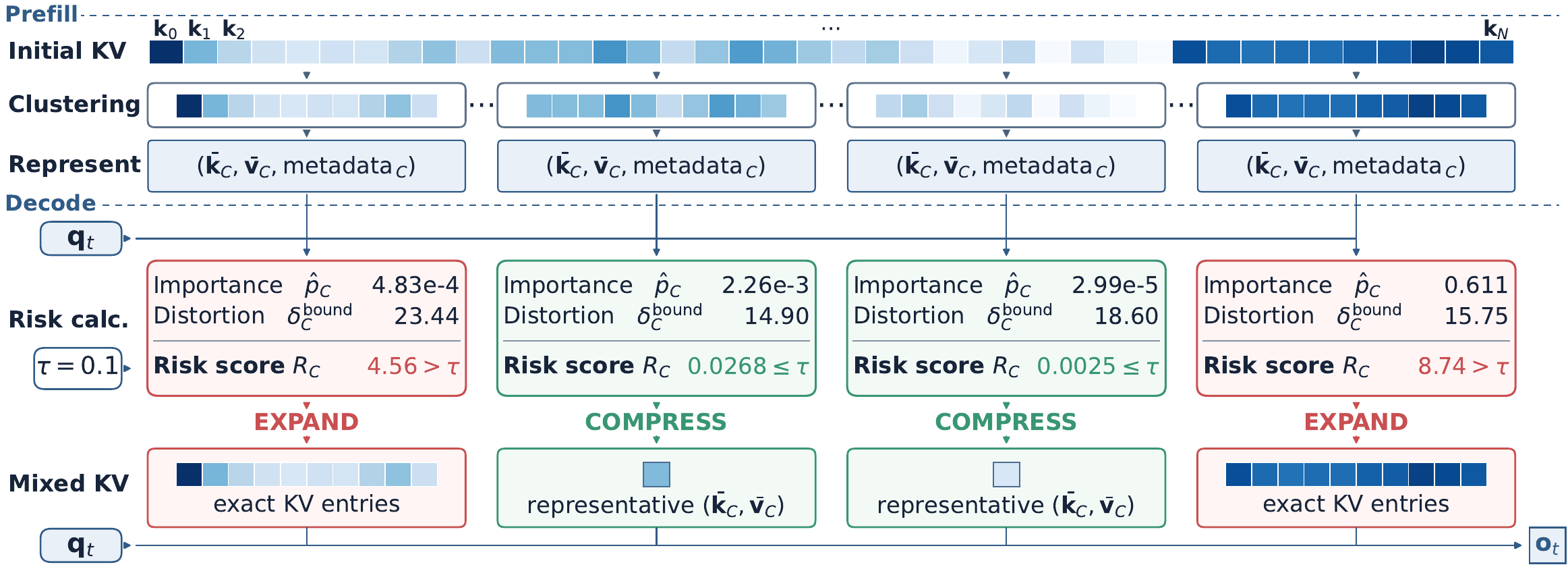}
    \caption{\textbf{\method Pipeline.} Prefill phase: The KV cache is partitioned into contiguous clusters, and a representative key-value pair together with auxiliary metadata is computed for each cluster. Decode phase: Each query $\boldsymbol{q}_t$ uses the cached summaries to calculate risk score $R_C$. Clusters with $R_C>\tau$ are expanded, while clusters with $R_C\leq\tau$ retain only their representatives. The resulting mixed KV sequence is used to compute the output $\mathbf{o}_t$. The color depth of each block indicates attention score.}
    \label{fig:pipeline}
\end{figure}

\textbf{Pipeline Overview.} We illustrate our pipeline in Figure~\ref{fig:pipeline}. In prefill phase, we first compute the entire KV cache and then partition it into contiguous clusters and precompute cluster representative mean key and value \((\bar{\boldsymbol{k}}_C, \bar{\boldsymbol{v}}_C)\) as well as auxiliary metadata. At each decoding step $t$, the incoming query $\boldsymbol{q_t}$ interacts only with these cluster-level summaries to estimate the approximation risk score $R_C$. Clusters with risk above a predefined threshold \(\tau\) are expanded for token-level attention, while the remaining clusters are represented by their cached means. A single attention operation over this mixed representation produces the approximate output.

\textbf{Efficient Estimation of Risk Score.} Evaluating the risk score $R_C$ in Eq.~\ref{eq:risk-score} requires four quantities: $\widehat{p}_C$, $V_{\max}$, $V_C$, and $\delta_C^{\mathrm{bound}}$. Among these, $\widehat{p}_C$ is computed directly from the cluster size $\vert{}C\vert{}$ and centroid $\bar{\boldsymbol{k}}_C$ via Eq.~\ref{eq:def-pc}, while $V_C$ and $V_{\max}$ are precomputed cluster-level statistics. Finally, $\delta_C^{\mathrm{bound}}$ is derived from compact cluster summaries, with details deferred to Appendix~\ref{app:deltabound}. 

\textbf{Determining the Expansion Set.} Once the risk scores are available, we compare them with a threshold \(\tau\) to determine which clusters should be expanded for token-level attention. For a single query head, the rule is straightforward: expand clusters with $R_C>\tau$ and approximate the remaining clusters by their representatives.
Modern large language models adopt Grouped-Query Attention (GQA)~\citep{ainslie2023gqatraininggeneralizedmultiquery}, where multiple query heads share a single key-value head. Because selection is executed at the KV-cache level, all query heads sharing a KV group must operate on a unified set of expanded clusters. We designate \textbf{GQA-any} as our default strategy: a cluster $C$ is expanded if the risk score evaluated by \emph{any} query head within the group exceeds the threshold. We also consider an alternative aggregation heuristic, \textbf{GQA-mean}, which expands a cluster only when the average risk across the group exceeds $\tau$; we systematically ablate both approaches in Section~\ref{sec:exp-ablation}.
\label{sec:gqa}

\textbf{Efficient kernel implementation.} Realizing the algorithmic sparsity of \method requires reducing actual KV-cache traffic. We first compute and cache the representative and bound statistics of each prompt block. At each decoding step, a lightweight block-selection kernel evaluates the expansion criterion of \method using the current query and these cached summaries, producing a mask that identifies the blocks requiring exact attention. The sparse-attention kernel consumes this mask directly on the GPU: it loads token-level K/V only from the selected blocks, while approximating the remaining blocks with their representatives. Appendix~\ref{app:kernel-implementation} provides the implementation details.
\section{Experiments}


\subsection{Experiment Setup}

\textbf{Models.} We conduct experiments on two open-sourced long context models: Llama-3.1-8B-Instruct~\citep{grattafiori2024llama3herdmodels} and Qwen2.5-7B-Instruct~\citep{qwen2025qwen25technicalreport}. Both models support up to 128K context, which is the default setting we use. We add experiments for Qwen2.5-14B-Instruct in Appendix \ref{app:qwen14b}, and context length ablation in Appendix \ref{app:length}.

\textbf{Benchmarks.}
We evaluate long-context generation on three representative benchmarks: LongBench~\citep{bai2024longbenchbilingualmultitaskbenchmark}, RULER~\citep{hsieh2024rulerwhatsrealcontext}, and HELMET~\citep{yen2025helmetevaluatelongcontextlanguage}, which cover complementary application scenarios. LongBench measures real-world long-document understanding across question answering, summarization, retrieval, and code completion. RULER uses controlled synthetic tasks to evaluate retrieval, association, and aggregation as the context length increases. HELMET covers diverse application-oriented workloads, testing whether the quality--efficiency trade-off generalizes across different patterns of long-context usage. 

\textbf{Baselines.} We compare \method with Full Attention and five representative efficient-attention baselines. 
SnapKV~\citep{li2024snapkvllmknowslooking} compresses the prompt KV cache by using attention statistics from a recent observation window. StreamingLLM~\citep{xiao2024efficientstreaminglanguagemodels} preserves attention-sink tokens together with a sliding window. 
PyramidKV~\citep{cai2025pyramidkvdynamickvcache} adaptively distributes the KV budget across layers. 
Quest~\citep{tang2024questqueryawaresparsityefficient} partitions the KV cache into pages and uses query-dependent key bounds to select pages that may receive high attention.
Double-P~\citep{ni2026doublephierarchicaltoppsparse} clusters prompt KVs and applies two-stage probability-mass pruning. 

\textbf{Metric.} We use measured latency, rather than the retained KV budget, as our primary efficiency metric because methods with similar token budgets can have substantially different runtimes. Since prefill is performed only once and contributes a relatively small fraction of the total cost (Analysis in Appendix~\ref{app:main-exp-metric}), we focus on decode time per output token (TPOT), which reflects end-to-end decode performance rather than isolated attention phase latency. For quality, we follow the official metric of each task and report the unweighted macro-average across tasks.

\textbf{Ours.} We partition the KV sequence into contiguous clusters of 32 tokens and report the quality--latency trade-off using two thresholds, \(\tau \in \{0.01, 0.1\}\). For GQA aggregation, we use GQA-any introduced in Section~\ref{sec:gqa}. All main results use \textbf{no} query anchors: at every decoding step, clusters are selected by \(R_C\) alone. Query anchors, are an optional technique that further improves quality; we describe and evaluate them in Appendix~\ref{app:qanchor}.

Detailed model configurations, benchmark setups, baseline implementations, evaluation metrics, and hardware settings are provided in Appendix~\ref{app:main-exp}.

\subsection{Main Results}
\begin{figure}
    \centering
    \includegraphics[width=\linewidth]{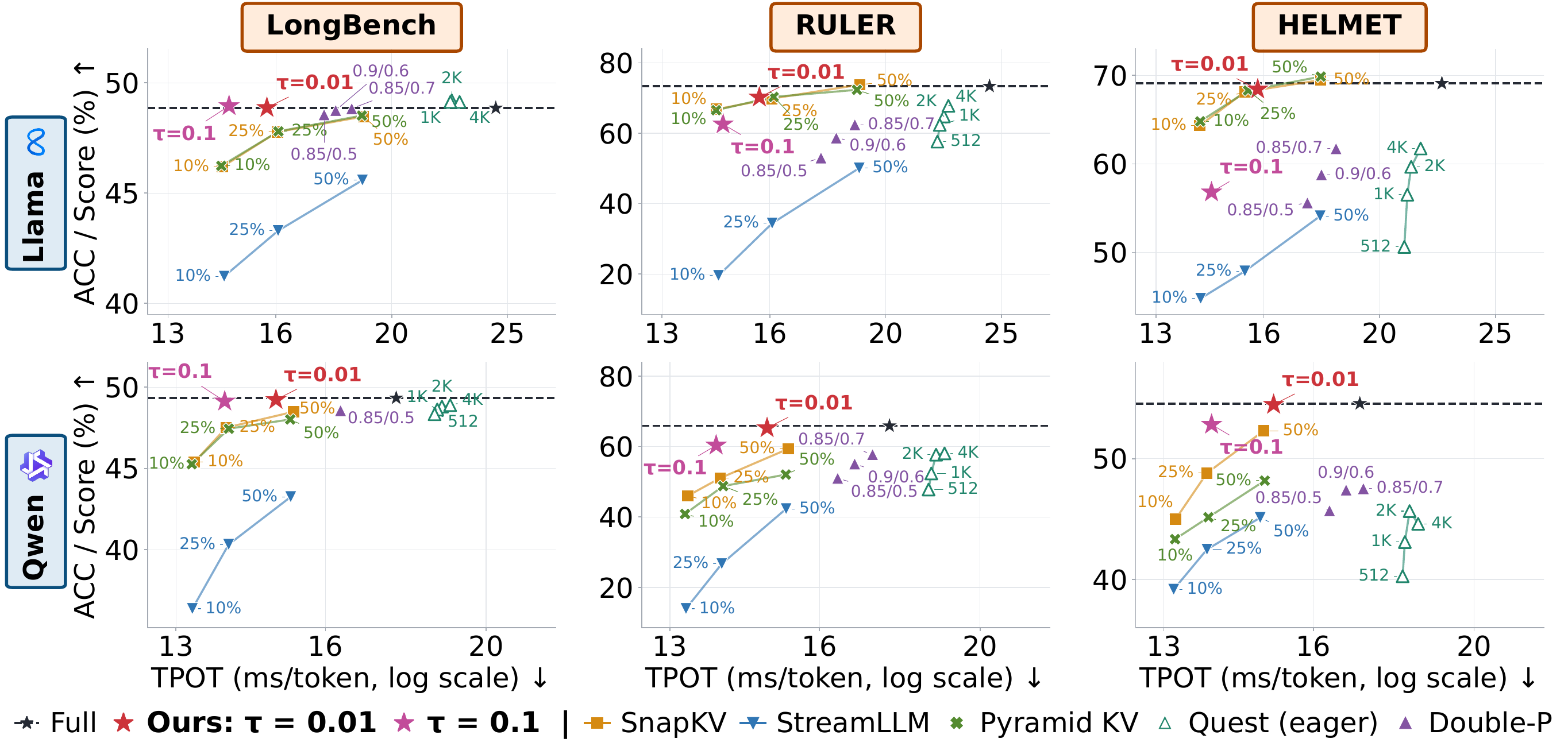}
    \caption{
    \textbf{Quality-latency trade-off of \method and existing KV-cache compression baselines.} Experiments are evaluated across two models (Llama-3.1 and Qwen-2.5) on three long-context benchmarks (LongBench, RULER, and HELMET), with results averaged across all tasks. We present a conservative threshold $\tau=0.01$, which matches full attention performance across all settings while achieving up to $1.56\times$ decode speedup as measured by Time Per Output Token (TPOT). Our method extends the quality–latency Pareto frontier.}
    \label{fig:main-exp}
\end{figure}

Figure~\ref{fig:main-exp} shows that \method consistently provides a favorable quality--latency trade-off across both models and all three benchmarks. With the conservative threshold $\tau=0.01$, \method nearly matches the quality of Full Attention in all six model--benchmark settings while reducing TPOT, achieving
up to a $1.55\times$ decode speedup over Full Attention. Compared with both eviction-based and query-adaptive baselines, \method introduces new Pareto-optimal operating points, extending the empirical frontier toward lower latency without sacrificing quality. Increasing the threshold to $\tau=0.1$ enables more aggressive sparsification and further improves the maximum speedup to $1.66\times$, although its effect on quality varies across tasks. Table~\ref{tab:main-task-group-acc-speedup} reports quality–latency results by task group, while  Appendix~\ref{app:maintable} provides the per-task results.

\providecolor{lightgreen}{RGB}{239,248,239}
\begin{table*}[t]
\centering
\caption{Task-group quality--latency comparison. Each cell reports $\mathrm{Accuracy}/\mathrm{Acceleration}$, where $\mathrm{Accuracy}$ is the task-macro benchmark score and $\mathrm{Acceleration}$ is latency acceleration relative to Full, higher is better for both. \protect\colorbox{lightgreen}{Shaded} cells are Pareto-optimal. \underline{Underlined} cells have an absolute $\mathrm{Accuracy}$ drop versus Full of less than $0.5$ point. Baselines use SnapKV/StreamLLM/PyramidKV.}
\label{tab:main-task-group-acc-speedup}
\setlength{\tabcolsep}{1.15pt}
\renewcommand{\arraystretch}{1.12}
\small
\resizebox{\textwidth}{!}{%
\begin{tabular}{@{}c@{\hspace{0.5pt}}l@{\hspace{0.5pt}}c@{\hspace{0.5pt}}*{11}{c}@{}}
\toprule
 & Method & \multicolumn{5}{c}{LongBench} & \multicolumn{3}{c}{RULER} & \multicolumn{4}{c}{HELMET} \\[-2pt]
\cmidrule(lr){3-7}\cmidrule(lr){8-10}\cmidrule(l){11-14}
 & & Doc QA & Summ. & Few-shot & Synth. & Code & NIAH & A\&T & QA & RAG & LC-QA & ICL & Rec./Rank. \\
\midrule
\multirow{9}{*}{\rotatebox[origin=c]{90}{\textbf{Llama}}} & Full & $43.6${\scriptsize $/1.0$} & $29.2${\scriptsize $/1.0$} & $69.2${\scriptsize $/1.0$} & $53.0${\scriptsize $/1.0$} & $59.6${\scriptsize $/1.0$} & $88.6${\scriptsize $/1.0$} & $45.0${\scriptsize $/1.0$} & $55.0${\scriptsize $/1.0$} & $60.4${\scriptsize $/1.0$} & $48.6${\scriptsize $/1.0$} & $77.0${\scriptsize $/1.0$} & $76.4${\scriptsize $/1.0$} \\
 & Snap $10\%$ & $41.2${\scriptsize $/1.7$} & $23.4${\scriptsize $/1.7$} & $65.9${\scriptsize $/1.7$} & \cellcolor{lightgreen}\strut \underline{$53.4${\scriptsize $/1.7$}} & \cellcolor{lightgreen}\strut $58.5${\scriptsize $/1.7$} & \cellcolor{lightgreen}\strut $79.5${\scriptsize $/1.7$} & $40.7${\scriptsize $/1.7$} & \underline{$55.0${\scriptsize $/1.7$}} & \cellcolor{lightgreen}\strut \underline{$61.0${\scriptsize $/1.7$}} & \cellcolor{lightgreen}\strut \underline{$48.4${\scriptsize $/1.6$}} & \cellcolor{lightgreen}\strut \underline{$77.0${\scriptsize $/1.6$}} & $60.7${\scriptsize $/1.5$} \\
 & Snap $25\%$ & $42.8${\scriptsize $/1.5$} & $26.3${\scriptsize $/1.5$} & $67.5${\scriptsize $/1.5$} & \underline{$53.2${\scriptsize $/1.5$}} & \underline{$59.8${\scriptsize $/1.5$}} & \cellcolor{lightgreen}\strut $84.2${\scriptsize $/1.5$} & $41.3${\scriptsize $/1.5$} & \underline{$55.0${\scriptsize $/1.5$}} & \underline{$60.6${\scriptsize $/1.5$}} & \cellcolor{lightgreen}\strut \underline{$49.4${\scriptsize $/1.5$}} & \cellcolor{lightgreen}\strut \underline{$78.0${\scriptsize $/1.5$}} & $71.8${\scriptsize $/1.4$} \\
 & Str. $10\%$ & $32.7${\scriptsize $/1.7$} & $22.2${\scriptsize $/1.7$} & $63.3${\scriptsize $/1.7$} & $47.5${\scriptsize $/1.7$} & $56.1${\scriptsize $/1.7$} & $11.2${\scriptsize $/1.7$} & $29.8${\scriptsize $/1.7$} & $37.5${\scriptsize $/1.7$} & $53.5${\scriptsize $/1.6$} & $42.4${\scriptsize $/1.6$} & $75.0${\scriptsize $/1.6$} & $8.6${\scriptsize $/1.5$} \\
 & Str. $25\%$ & $35.3${\scriptsize $/1.5$} & $24.7${\scriptsize $/1.5$} & $66.1${\scriptsize $/1.5$} & $46.2${\scriptsize $/1.5$} & $58.1${\scriptsize $/1.5$} & $30.6${\scriptsize $/1.5$} & $34.2${\scriptsize $/1.5$} & $50.0${\scriptsize $/1.5$} & $57.9${\scriptsize $/1.5$} & $41.5${\scriptsize $/1.5$} & $73.0${\scriptsize $/1.5$} & $17.3${\scriptsize $/1.4$} \\
 & Pyr. $10\%$ & \cellcolor{lightgreen}\strut $41.3${\scriptsize $/1.7$} & $23.6${\scriptsize $/1.7$} & $66.0${\scriptsize $/1.7$} & \cellcolor{lightgreen}\strut \underline{$53.3${\scriptsize $/1.7$}} & \cellcolor{lightgreen}\strut $58.4${\scriptsize $/1.7$} & \cellcolor{lightgreen}\strut $79.7${\scriptsize $/1.7$} & $39.2${\scriptsize $/1.7$} & \cellcolor{lightgreen}\strut \underline{$55.0${\scriptsize $/1.7$}} & \underline{$61.0${\scriptsize $/1.7$}} & \cellcolor{lightgreen}\strut \underline{$49.0${\scriptsize $/1.6$}} & $75.0${\scriptsize $/1.6$} & $63.9${\scriptsize $/1.5$} \\
 & Pyr. $25\%$ & $43.0${\scriptsize $/1.5$} & $26.2${\scriptsize $/1.5$} & $67.5${\scriptsize $/1.5$} & \underline{$53.0${\scriptsize $/1.5$}} & \cellcolor{lightgreen}\strut \underline{$59.8${\scriptsize $/1.5$}} & \cellcolor{lightgreen}\strut $84.8${\scriptsize $/1.5$} & $41.8${\scriptsize $/1.5$} & \underline{$55.0${\scriptsize $/1.5$}} & \underline{$60.4${\scriptsize $/1.5$}} & \underline{$49.4${\scriptsize $/1.5$}} & $76.0${\scriptsize $/1.5$} & \cellcolor{lightgreen}\strut $74.4${\scriptsize $/1.4$} \\
 & {\bfseries\boldmath Our $\tau.01$} & \underline{$\mathbf{43.3}${\scriptsize $\mathbf{/1.6}$}} & \cellcolor{lightgreen}\strut \underline{$\mathbf{29.3}${\scriptsize $\mathbf{/1.5}$}} & \cellcolor{lightgreen}\strut \underline{$\mathbf{69.4}${\scriptsize $\mathbf{/1.5}$}} & \underline{$\mathbf{53.3}${\scriptsize $\mathbf{/1.6}$}} & \underline{$\mathbf{59.7}${\scriptsize $\mathbf{/1.6}$}} & \cellcolor{lightgreen}\strut $\mathbf{82.8}${\scriptsize $\mathbf{/1.6}$} & \cellcolor{lightgreen}\strut \underline{$\mathbf{46.7}${\scriptsize $\mathbf{/1.6}$}} & \underline{$\mathbf{55.0}${\scriptsize $\mathbf{/1.5}$}} & \cellcolor{lightgreen}\strut \underline{$\mathbf{61.7}${\scriptsize $\mathbf{/1.5}$}} & \underline{$\mathbf{48.9}${\scriptsize $\mathbf{/1.5}$}} & $\mathbf{75.0}${\scriptsize $\mathbf{/1.4}$} & \cellcolor{lightgreen}\strut $\mathbf{75.1}${\scriptsize $\mathbf{/1.4}$} \\
 & {\bfseries\boldmath Our $\tau.1$} & \cellcolor{lightgreen}\strut \underline{$\mathbf{43.8}${\scriptsize $\mathbf{/1.7}$}} & \cellcolor{lightgreen}\strut \underline{$\mathbf{28.9}${\scriptsize $\mathbf{/1.7}$}} & \cellcolor{lightgreen}\strut \underline{$\mathbf{69.2}${\scriptsize $\mathbf{/1.7}$}} & \underline{$\mathbf{53.2}${\scriptsize $\mathbf{/1.7}$}} & \cellcolor{lightgreen}\strut \underline{$\mathbf{59.7}${\scriptsize $\mathbf{/1.7}$}} & $\mathbf{72.2}${\scriptsize $\mathbf{/1.7}$} & $\mathbf{40.3}${\scriptsize $\mathbf{/1.7}$} & \cellcolor{lightgreen}\strut \underline{$\mathbf{57.5}${\scriptsize $\mathbf{/1.7}$}} & $\mathbf{57.7}${\scriptsize $\mathbf{/1.6}$} & $\mathbf{47.0}${\scriptsize $\mathbf{/1.6}$} & $\mathbf{76.0}${\scriptsize $\mathbf{/1.6}$} & $\mathbf{40.8}${\scriptsize $\mathbf{/1.5}$} \\
\midrule
\multirow{9}{*}{\rotatebox[origin=c]{90}{\textbf{Qwen}}} & Full & $43.2${\scriptsize $/1.0$} & $27.8${\scriptsize $/1.0$} & $69.9${\scriptsize $/1.0$} & $54.5${\scriptsize $/1.0$} & $64.1${\scriptsize $/1.0$} & $74.8${\scriptsize $/1.0$} & $57.6${\scriptsize $/1.0$} & $42.5${\scriptsize $/1.0$} & $48.5${\scriptsize $/1.0$} & $50.8${\scriptsize $/1.0$} & $76.0${\scriptsize $/1.0$} & $39.5${\scriptsize $/1.0$} \\
 & Snap $10\%$ & \cellcolor{lightgreen}\strut $41.3${\scriptsize $/1.3$} & $21.3${\scriptsize $/1.3$} & $62.2${\scriptsize $/1.3$} & \underline{$54.5${\scriptsize $/1.3$}} & $59.5${\scriptsize $/1.3$} & $45.3${\scriptsize $/1.3$} & $53.6${\scriptsize $/1.3$} & $37.5${\scriptsize $/1.3$} & $46.7${\scriptsize $/1.3$} & \cellcolor{lightgreen}\strut $49.9${\scriptsize $/1.3$} & \cellcolor{lightgreen}\strut $74.0${\scriptsize $/1.3$} & $12.7${\scriptsize $/1.3$} \\
 & Snap $25\%$ & \underline{$42.8${\scriptsize $/1.3$}} & $23.9${\scriptsize $/1.3$} & $66.2${\scriptsize $/1.3$} & \underline{$54.5${\scriptsize $/1.3$}} & $62.0${\scriptsize $/1.3$} & $51.2${\scriptsize $/1.3$} & \cellcolor{lightgreen}\strut \underline{$57.9${\scriptsize $/1.3$}} & \cellcolor{lightgreen}\strut $40.0${\scriptsize $/1.3$} & \cellcolor{lightgreen}\strut \underline{$49.2${\scriptsize $/1.3$}} & $49.1${\scriptsize $/1.2$} & \cellcolor{lightgreen}\strut $75.0${\scriptsize $/1.2$} & $22.2${\scriptsize $/1.2$} \\
 & Str. $10\%$ & $30.1${\scriptsize $/1.3$} & $19.7${\scriptsize $/1.3$} & $62.8${\scriptsize $/1.3$} & $23.2${\scriptsize $/1.3$} & $53.8${\scriptsize $/1.3$} & $6.9${\scriptsize $/1.3$} & $27.1${\scriptsize $/1.3$} & $22.5${\scriptsize $/1.3$} & $38.1${\scriptsize $/1.3$} & $46.3${\scriptsize $/1.3$} & $70.0${\scriptsize $/1.3$} & $6.3${\scriptsize $/1.3$} \\
 & Str. $25\%$ & $33.0${\scriptsize $/1.3$} & $22.5${\scriptsize $/1.3$} & $65.9${\scriptsize $/1.3$} & $33.1${\scriptsize $/1.3$} & $57.8${\scriptsize $/1.3$} & $22.0${\scriptsize $/1.3$} & $30.3${\scriptsize $/1.3$} & $40.0${\scriptsize $/1.3$} & $39.8${\scriptsize $/1.3$} & $47.3${\scriptsize $/1.2$} & $75.0${\scriptsize $/1.2$} & $10.2${\scriptsize $/1.2$} \\
 & Pyr. $10\%$ & \cellcolor{lightgreen}\strut $40.5${\scriptsize $/1.3$} & $22.2${\scriptsize $/1.3$} & $61.9${\scriptsize $/1.3$} & \cellcolor{lightgreen}\strut \underline{$54.5${\scriptsize $/1.3$}} & $60.0${\scriptsize $/1.3$} & $38.1${\scriptsize $/1.3$} & $52.0${\scriptsize $/1.3$} & $35.0${\scriptsize $/1.3$} & \cellcolor{lightgreen}\strut $47.3${\scriptsize $/1.3$} & $49.0${\scriptsize $/1.3$} & $71.0${\scriptsize $/1.3$} & $10.2${\scriptsize $/1.3$} \\
 & Pyr. $25\%$ & $42.1${\scriptsize $/1.3$} & $24.2${\scriptsize $/1.3$} & $66.9${\scriptsize $/1.3$} & \underline{$54.5${\scriptsize $/1.3$}} & $62.1${\scriptsize $/1.3$} & $48.8${\scriptsize $/1.3$} & $54.7${\scriptsize $/1.3$} & $40.0${\scriptsize $/1.3$} & $45.6${\scriptsize $/1.2$} & $48.7${\scriptsize $/1.2$} & $75.0${\scriptsize $/1.2$} & $13.5${\scriptsize $/1.2$} \\
 & {\bfseries\boldmath Our $\tau.01$} & \cellcolor{lightgreen}\strut \underline{$\mathbf{43.0}${\scriptsize $\mathbf{/1.2}$}} & \underline{$\mathbf{27.8}${\scriptsize $\mathbf{/1.2}$}} & \cellcolor{lightgreen}\strut \underline{$\mathbf{69.8}${\scriptsize $\mathbf{/1.2}$}} & \underline{$\mathbf{54.5}${\scriptsize $\mathbf{/1.2}$}} & \cellcolor{lightgreen}\strut \underline{$\mathbf{64.0}${\scriptsize $\mathbf{/1.2}$}} & \cellcolor{lightgreen}\strut $\mathbf{73.3}${\scriptsize $\mathbf{/1.2}$} & \cellcolor{lightgreen}\strut \underline{$\mathbf{59.1}${\scriptsize $\mathbf{/1.2}$}} & \cellcolor{lightgreen}\strut \underline{$\mathbf{42.5}${\scriptsize $\mathbf{/1.2}$}} & \underline{$\mathbf{49.2}${\scriptsize $\mathbf{/1.1}$}} & \cellcolor{lightgreen}\strut \underline{$\mathbf{50.8}${\scriptsize $\mathbf{/1.2}$}} & \underline{$\mathbf{76.0}${\scriptsize $\mathbf{/1.1}$}} & \cellcolor{lightgreen}\strut $\mathbf{38.7}${\scriptsize $\mathbf{/1.1}$} \\
 & {\bfseries\boldmath Our $\tau.1$} & \cellcolor{lightgreen}\strut \underline{$\mathbf{43.0}${\scriptsize $\mathbf{/1.3}$}} & \cellcolor{lightgreen}\strut \underline{$\mathbf{27.9}${\scriptsize $\mathbf{/1.3}$}} & \cellcolor{lightgreen}\strut \underline{$\mathbf{69.7}${\scriptsize $\mathbf{/1.3}$}} & \underline{$\mathbf{54.2}${\scriptsize $\mathbf{/1.3}$}} & \cellcolor{lightgreen}\strut $\mathbf{63.5}${\scriptsize $\mathbf{/1.3}$} & $\mathbf{67.2}${\scriptsize $\mathbf{/1.3}$} & \cellcolor{lightgreen}\strut \underline{$\mathbf{57.3}${\scriptsize $\mathbf{/1.3}$}} & $\mathbf{37.5}${\scriptsize $\mathbf{/1.3}$} & \cellcolor{lightgreen}\strut \underline{$\mathbf{49.4}${\scriptsize $\mathbf{/1.2}$}} & \cellcolor{lightgreen}\strut \underline{$\mathbf{50.7}${\scriptsize $\mathbf{/1.2}$}} & \cellcolor{lightgreen}\strut \underline{$\mathbf{76.0}${\scriptsize $\mathbf{/1.2}$}} & $\mathbf{33.3}${\scriptsize $\mathbf{/1.2}$} \\
\bottomrule
\end{tabular}%
}
\end{table*}

\textbf{Outperform full attention in certain tasks.} Interestingly, our method can outperform full attention on tasks dominated by irrelevant context like LongBench-GovReport or RULER-NIAH-MultiKey. One possible explanation is that preserving salient blocks exactly while representing less relevant blocks acts as an adaptive information filter, allowing the model to focus more on task-relevant evidence. 

\paragraph{Latency Breakdown.}
\begin{wrapfigure}{r}{0.50\textwidth}
  \centering
  \includegraphics[width=0.50\textwidth]{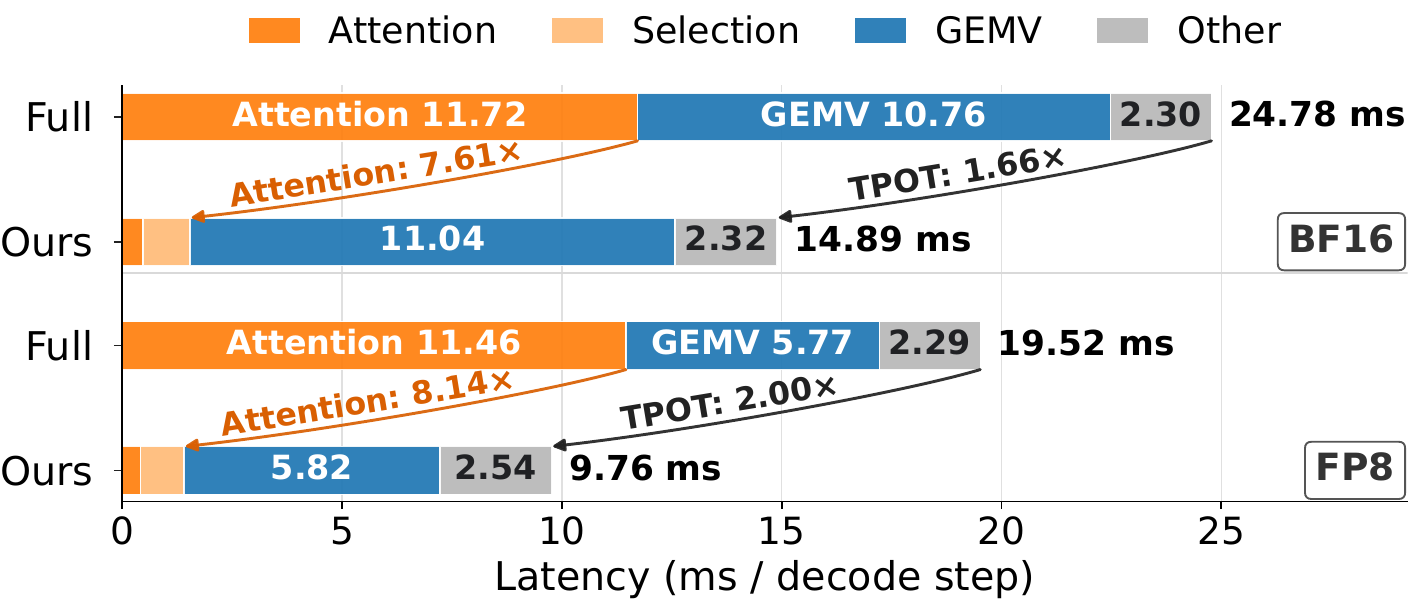} 
  \setlength{\belowcaptionskip}{-25pt}
  \caption{\textbf{Latency Breakdown. } }
  \label{fig:latency}
\end{wrapfigure}
Figure~\ref{fig:latency} shows that \method substantially reduces decoding latency. Under BF16, it achieves a \(7.6\times\) attention speedup and a \(1.66\times\) overall decoding speedup. The overall gain is limited by non-attention operations, particularly the GEMV kernels. FP8 cuts GEMV from 10.8 ms to 5.8 ms per step, under FP8, the overall decoding speedup increases to \(2.00\times\). Results are measured on Llama-3.1-8B-Instruct with 128K input.

\subsection{Ablation Study.}
\begin{wraptable}{r}{0.50\textwidth}
\centering
\begin{minipage}{0.50\textwidth}
\centering
\caption{Selection-score ablation at a matched per-head token budget
(Llama-3.1-8B-Instruct, RULER-128K, $Q{=}0$). Only the score that ranks clusters changes.
$\Delta$ is relative to Ours. Exact $\delta_C$ is an oracle that scans all keys.}
\label{tab:ablation-score}
\small
\setlength{\tabcolsep}{4pt}
\begin{tabular}{lcccc}
\toprule
 & \multicolumn{2}{c}{$\tau{=}0.01$} & \multicolumn{2}{c}{$\tau{=}0.1$} \\
\cmidrule(lr){2-3}\cmidrule(lr){4-5}
Score & Acc. & $\Delta$ & Acc. & $\Delta$ \\
\midrule
\textbf{Ours} ($R_C$, box) & \textbf{70.27} & --- & \textbf{62.58} & --- \\
$R_C$, ellipsoid      & 68.90 & $-1.37$ & 57.29 & $-5.28$ \\
Quest-style           & 66.38 & $-3.88$ & 58.87 & $-3.71$ \\
$\hat p_C V_C$        & 58.78 & $-11.49$ & 41.38 & $-21.19$ \\
$\hat p_C$            & 59.35 & $-10.92$ & 42.87 & $-19.71$ \\
$\delta_C$            & 47.23 & $-23.04$ & 20.68 & $-41.90$ \\
\midrule
$R_C$, exact $\delta_C$ & 72.00 & $+1.73$ & 69.06 & $+6.49$ \\
Full attention        & 73.37 & $+3.10$ & 73.37 & $+10.79$ \\
\bottomrule
\end{tabular}
\end{minipage}
\end{wraptable}

We ablate on Llama-3.1-8B-Instruct with RULER-128K; \textbf{Ours} uses $\tau=0.01$, $|C|=32$, the box summary, GQA-any and $Q=0$.

\textbf{Selection score.} Table~\ref{tab:ablation-score} fixes the per-head token budget to that of Ours and changes only the score that ranks clusters.
Importance alone ($\hat p_C$, $\hat p_C V_C$) or deviation alone ($\delta_C$) falls far behind $R_C$, and the Quest-style logit bound is also lower.
The exact-$\delta_C$ oracle shows that a tighter summary would help most at small budgets.

\textbf{Design choices.} Table~\ref{tab:ablation} changes one setting at a time.
$\tau$ is the main quality--latency knob; $|C|$ barely affects accuracy; GQA-mean is slightly faster but less accurate; the box summary beats the ellipsoid in every setting; and prompt-query anchors further improve accuracy at some cost in speed (Appendix~\ref{app:qanchor}).












\begin{table}[ht]
\centering
\caption{Ablation study of Llama-3.1-8B-Instruct on RULER-128K .
\textbf{Ours} is the default configuration: $\tau{=}0.01$, $|C|{=}32$, box deviation summary,
GQA-any, no prompt-query anchors ($Q{=}0$). Each other column changes one setting.
TPOT is steady-state decode latency at 128K input under CUDA Graph; speedup is relative to Full Attention.}
\label{tab:ablation}
\small
\setlength{\tabcolsep}{3.5pt}
\begin{tabular}{l cc ccc cc c c c}
\toprule
 & \multicolumn{2}{c}{Ref.} & \multicolumn{3}{c}{Threshold} & \multicolumn{2}{c}{Cluster size} & GQA & Summary & Anchors \\
\cmidrule(lr){2-3}\cmidrule(lr){4-6}\cmidrule(lr){7-8}\cmidrule(lr){9-9}\cmidrule(lr){10-10}\cmidrule(lr){11-11}
Metric & Full & \textbf{Ours} & $\tau{=}0.001$ & $\tau{=}0.1$ & $\tau{=}1$
       & $|C|{=}16$ & $|C|{=}64$ & mean & ellipsoid & $Q{=}32$ \\
\midrule
Acc.$\uparrow$   & 73.37 & \textbf{70.19} & 73.63 & 62.58 & 34.67 & 70.36 & 70.34 & 68.82 & 68.90 & 72.99 \\
TPOT (ms)$\downarrow$ & 24.43 & \textbf{15.68} & 19.08 & 14.62 & 14.42 & 15.86 & 16.20 & 15.06 & 15.14 & 17.14 \\
Speedup$\uparrow$ & 1.00$\times$ & \textbf{1.56$\times$} & 1.28$\times$ & 1.67$\times$ & 1.69$\times$
                  & 1.54$\times$ & 1.51$\times$ & 1.62$\times$ & 1.61$\times$ & 1.43$\times$ \\
\bottomrule
\end{tabular}
\end{table}
\label{sec:exp-ablation}

\section{Related Works}

\textbf{Eviction-based KV Cache Compression.} A prominent line of work reduces KV-cache overhead by permanently retaining only a subset of tokens deemed important~\citep{zhang2023h2oheavyhitteroracleefficient,xiao2024efficientstreaminglanguagemodels,cai2025pyramidkvdynamickvcache,guo2024vatpattentionscoreneedtoken,oren2024tovatransformersmultistaternns,ren2024rocoefficacyevictionpolicykeyvalue,li2026momentkvclosingdirectionalgap}. H$_2$O~\citep{zhang2023h2oheavyhitteroracleefficient} identifies \emph{heavy-hitter} tokens from accumulated attention statistics and dynamically preserves them together with recent tokens. StreamingLLM~\citep{xiao2024efficientstreaminglanguagemodels} exploits the attention-sink phenomenon and maintains a small set of initial sink tokens alongside a sliding window of recent tokens. Another line of work compresses the cache by merging multiple KV states into fewer representatives~\citep{CAM,wan2025d2odynamicdiscriminativeoperations,wang2024modeltellsmergeadaptivekvmerger,yuan2025weightedkvattentionscoresweighted,hu2026centroidkvefficientlongcontextllm}; while retaining more aggregate information than hard eviction, such merging is still generally lossy because the original token-level states cannot be exactly recovered once combined. Despite their different compression mechanisms, these methods make irreversible decisions about the KV cache: discarded or merged token-level states are no longer fully accessible to future queries.

\textbf{Query-Aware Cluster-Based Attention.} 
To avoid the irreversible information loss, another line of work preserves recoverable access to the original KV states and performs query-conditioned sparse attention at inference time. Block-based methods~\citep{tang2024questqueryawaresparsityefficient,lu2025mobamixtureblockattention,tian2026cobscumulantorderblock} partition the KV cache into contiguous pages and use compact page-level statistics to identify regions relevant to the current query . Quest~\citep{tang2024questqueryawaresparsityefficient} estimates the importance of each page using coordinate-wise key extrema and loads only the highest-scoring pages for exact attention. More recent methods, including Multipole Attention~\citep{hooper2025multipoleattentionefficientlong}, and Double-P~\citep{ni2026doublephierarchicaltoppsparse}, further adopt a coarse-to-fine, mixed-resolution design. These approaches establish block and cluster summaries as an efficient interface for query-dependent access to long contexts. However, their refinement decisions are primarily driven by estimated attention scores, rather than by a direct characterization of the attention-output error introduced when a cluster remains at low resolution.

\textbf{Adaptive Sparse-Attention Selection.}
Recent KV compression and sparse attention methods have increasingly explored adaptive computation budgets to accommodate the heterogeneous attention patterns across heads, layers, and queries~\citep{feng2025adakvoptimizingkvcache,cai2025pyramidkvdynamickvcache}.
A fixed top-$k$ budget cannot adapt to the substantial
variation in attention concentration across queries, heads, and layers.
Twilight~\citep{lin2025twilightadaptiveattentionsparsity} addresses this limitation by introducing top-$p$ pruning, which dynamically adjusts the retained set.
Tactic~\citep{zhu2025tacticadaptivesparseattention} uses function fitting to estimate the weight distributions under top-p clustering. Our method provides a complementary route to adaptive budgeting: instead of explicitly prescribing or allocating a budget, we expand clusters according to their query-conditioned approximation risk. Consequently, the KV access budget emerges naturally from error control, allocating more computation only when a coarse representation is insufficient to faithfully approximate the attention output.
\section{Limitations and Future Work}
Our work has several limitations that suggest promising directions for future research. 

\textbf{End-to-end error guarantee.} Our theoretical analysis focuses on the approximation error of a single-head attention output and does not characterize how such errors propagate through Transformer layers. Extending the analysis toward end-to-end guarantees remains a challenging yet important direction. 

\textbf{Mixed-state re-scoring.} RefineKV scores all clusters once, in the state where every cluster is approximated. Eq. \ref{eq:risk-score} also bounds the output once some clusters are expanded, but expanding a cluster changes the softmax normalizer, so scores computed before expansion no longer match the error terms afterward. A natural extension is to re-score the remaining clusters after each round of expansion. This would track the error of the mixed output more closely, at the cost of extra rounds of KV access during decoding. Making this cheap, for example by updating the normalizer incrementally or limiting it to a few rounds, is left for future work.

\textbf{Better estimation of $\delta$.} Our selection rule relies on a computable upper bound on the logit deviation $\delta$ of each block, which is cheap to evaluate but can be loose. When we replace this bound with the \emph{oracle} $\delta$, computed exactly from the keys and values inside each block, the method achieves a consistently better quality--compression trade-off under the same budget. We leave the design of tighter yet efficient approximations to it as future work.

\textbf{Hierarchical clustering}. our current method adopts contiguous clustering. A natural extension is hierarchical clustering, where decoding starts from coarse cluster summaries and progressively expands only the regions that require higher-resolution access. Such a hierarchy could enable flexible allocation of KV access budgets and provide a better trade-off between accuracy and decoding efficiency.

\textbf{Other modalities.} RefineKV is modality-agnostic: its summaries, score and kernels operate only on cached keys and values. Extending it to vision-language models raises two questions we leave to future work. First, contiguous blocks of raster-ordered patches may group spatially distant tokens, so 2D-tile or spatio-temporal clustering may yield tighter summaries; since decode attention does not depend on the order in which keys are stored, this only requires reordering tokens once at prefill. Second, fine-grained visual tasks such as OCR or small-object grounding may need a smaller $\tau$, and differing value scales across modalities may call for per-modality thresholds.

\section{Conclusion}
In this work, we revisit KV compression from the perspective of query-adaptive access. Starting from attention output error, we derive a query-dependent cluster-wise risk score that characterizes whether each cluster can be faithfully approximated. Based on this criterion, we present \method, a query-adaptive KV access method that preserves the full cache while selectively expanding clusters that require fine-grained attention. Experiments show that \method achieves a favorable trade-off between quality and efficiency.

\section*{Acknowledgments}
This work was funded by the National University of Singapore under its Start-up Grant (Award No: SUG-251RES2505).

\bibliography{iclr2027_conference}
\bibliographystyle{iclr2027_conference}

\appendix
\section{Derivation for cluster selection criterion}

\label{app:derivation}
We provide a detailed derivation of our selection criterion in the following section. We present the formulation for a single attention head and omit layer and head indices for clarity.

\subsection{Setup.}

Let $\{(\boldsymbol{k}_i,\boldsymbol{v}_i)\}_{i=1}^{N}$ denote the KV cache, where $\boldsymbol{k}_i \in \mathbb{R}^d, \boldsymbol{v}_i \in \mathbb{R}^{d_v}$. Let $\mathcal{C}=\{C_1,\ldots,C_m\}$ be a partition of the KV cache. 
For each cluster $C\in\mathcal{C}$, let $n_C=|C|$ and define
\begin{equation}
\bar{\boldsymbol{k}}_C=\frac{1}{n_C}\sum_{i\in C}\boldsymbol{k}_i,
\qquad
\bar{\boldsymbol{v}}_C=\frac{1}{n_C}\sum_{i\in C}\boldsymbol{v}_i.
\end{equation}
For a query $\boldsymbol{q}\in\mathbb{R}^d$, define the token and representative logits as
\begin{equation}
s_i=\frac{\boldsymbol{q}^\top \boldsymbol{k}_i}{\sqrt d},
\qquad
s_C=\frac{\boldsymbol{q}^\top\bar{\boldsymbol{k}}_C}{\sqrt d}.
\end{equation}
The exact and approximated unnormalized masses of cluster $C$ are
\begin{equation}
z_C=\sum_{i\in C}e^{s_i},
\qquad
\widehat z_C=n_Ce^{s_C}.
\end{equation}
Equivalently, the representative of $C$ has logit $s_C+\log n_C$.

The exact and approximated total mass are:
\begin{equation}
Z=\sum_{C\in \mathcal{C}}z_C,
\qquad
\widehat Z=\sum_{C\in \mathcal{C}}\widehat z_C.
\end{equation}

Define the exact within-cluster output
\begin{equation}
\boldsymbol{v}_C=\frac{\sum_{i\in C}e^{s_i}\boldsymbol{v}_i}{z_C},
\end{equation}
and the exact and approximated cluster probabilities
\begin{equation}
p_C=\frac{z_C}{Z},
\qquad
\widehat p_C=
\frac{\widehat z_C}{\widehat Z}.
\end{equation}
The exact and approximated attention outputs are therefore
\begin{equation}
\boldsymbol{o}=\sum_{C\in\mathcal{C}}p_C\boldsymbol{v}_C,
\qquad
\widehat{\boldsymbol{o}}=\sum_{C\in\mathcal{C}}\widehat p_C\bar{\boldsymbol{v}}_C.
\end{equation}

Let
\begin{equation}
V_{\max}=\max_i\lVert \boldsymbol{v}_i\rVert_2,
\qquad
V_C=\max_{i\in C}\lVert \boldsymbol{v}_i\rVert_2.
\end{equation}

\subsection{Decomposition of attention output.}

We first decompose the single-head attention output into \textbf{inter-cluster mass redistribution} and \textbf{intra-cluster value distortion}.
Defining the auxiliary output \(\widetilde{\boldsymbol{o}} = \sum_{C\in\mathcal C} \widehat p_C \boldsymbol{v}_C\), we have:
\begin{align}
    \boldsymbol{o}-\widehat{\boldsymbol{o}}
    &= (\boldsymbol{o}-\widetilde{\boldsymbol{o}}) + (\widetilde{\boldsymbol{o}} -\widehat{\boldsymbol{o}})\\ 
    &= \underbrace{
    \sum_{C\in\mathcal{C}}(p_C-\widehat p_C)\boldsymbol{v}_C
    }_{\text{inter-cluster mass redistribution}}
    +
    \underbrace{
    \sum_{C\in\mathcal{C}} \widehat p_C(\boldsymbol{v}_C-\bar{\boldsymbol{v}}_C)
    }_{\text{intra-cluster value distortion}}.
\end{align}

This decomposition separates the output error into two sources: inter-cluster mass redistribution and intra-cluster value distortion. It reveals that the approximation error arises through two mechanisms: 
The first term, \((p_C-\widehat p_C)\), reflects how the estimated cluster probability mass deviates from the real mass, capturing the redistribution of attention mass \textit{across clusters} caused by approximating with cluster representatives. 

The second term describes the error caused by approximating $\boldsymbol{v}_C$ with $\bar{\boldsymbol{v}}_C$. It satisfies a desired property: \textit{For a cluster $C$, when all $\boldsymbol{q}^\top \boldsymbol{k}_i$ are close to $\boldsymbol{q}^\top \bar{\boldsymbol{k}}_C$, the cluster is well-approximated and this term gets close to zero. In particular, if $\boldsymbol{q}^\top \boldsymbol{k}_i = \boldsymbol{q}^\top \bar{\boldsymbol{k}}_C,\forall i \in C$, the term becomes exactly zero.}

The first term, $p_C-\widehat p_C$, does not satisfy this property. It reflects shift in probability assigned to cluster $C$, which is coupled with approximation of all clusters due to softmax normalization. Instead, we want to measure how approximating cluster $C$ impacts the overall output. Therefore, we regroup the first term in a source-wise way. 

\subsection{Regrouping Inter-cluster mass redistribution}
\begin{lemma}[Core Lemma: Source-wise regroup of inter-cluster mass redistribution]
\label{lem:source-mass}

Define the auxiliary output
\begin{equation}
    \widetilde{\boldsymbol{o}}
    =
    \sum_{C\in\mathcal C}\widehat p_C \boldsymbol{v}_C,
\end{equation}
the cluster mass correction factor
\begin{equation}
    m_C
    =
    \frac{z_C}{\widehat z_C},
\end{equation}
Then the inter-cluster mass redistribution error admits the exact source-wise decomposition
\begin{equation}
\boxed{
    \boldsymbol{o}-\widetilde{\boldsymbol{o}}
    =
    \sum_{C\in\mathcal C}
    \widehat p_C\frac{m_C-1}{\sum_{C'\in\mathcal C}\widehat p_{C'} m_{C'}}
    (\boldsymbol{v}_C-\widetilde{\boldsymbol{o}}).
}
\label{eq:source-wise-mass}
\end{equation}
In particular, if cluster $C$ has no mass approximation error, i.e., $z_C=\widehat z_C$ (equivalently, $m_C=1$), its source-wise contribution to the mass-redistribution error is exactly zero.
\end{lemma}
Using Lemma~\ref{lem:source-mass}, we first rewrite \( \boldsymbol{o}-\widetilde{\boldsymbol{o}} \) as:
\begin{equation}
    \boldsymbol{o}-\widetilde{\boldsymbol{o}}
    =
    \sum_{C\in\mathcal C}
    \widehat p_C\frac{m_C-1}{\sum_{C'\in\mathcal C}\widehat p_{C'} m_{C'}}
    (\boldsymbol{v}_C-\widetilde{\boldsymbol{o}}).
\end{equation}
With \(\left\| \boldsymbol{v}_C-\widetilde{\boldsymbol{o}} \right\|_2 \leq 2V_{\max}\):
\begin{equation}
    \boxed{
    \left\|
    \boldsymbol{o}-\widetilde{\boldsymbol{o}}
    \right\|_2
    \leq
    2V_{\max}
    \frac{
    \sum_{C\in\mathcal C}\widehat p_C(m_C-1)
    }{
    \sum_{C'\in\mathcal C}\widehat p_{C'} m_{C'}
    }.
    }
\end{equation}
To connect the local mass correction $m_C$ with the approximation fidelity of cluster $C$, we define the query-dependent logit deviation
\begin{equation}
    \delta_C
    =
    \max_{i\in C}|s_i-s_C|
    =
    \max_{i\in C}
    \left|
    \frac{\boldsymbol{q}^\top(\boldsymbol{k}_i-\bar{\boldsymbol{k}}_C)}{\sqrt d}
    \right|.
\end{equation}

By technical lemma \ref{lem:mass_bound}, $m_C\leq \cosh(\delta_C)$. Note that $\sum_{C\in\mathcal C}\widehat p_C = 1$, which gives
\begin{equation}
    \frac{\sum_{C\in\mathcal C}\widehat p_C(m_C-1)}{\sum_{C\in\mathcal C}\widehat p_C m_C} = 1 - \frac{1}{\sum_{C\in\mathcal C}\widehat p_C m_C}.
\end{equation}
Since $t\mapsto 1-1/t$ is monotonically increasing for $t\ge 1$, we obtain:
\begin{equation}
    \boxed{
    \left\|
    \boldsymbol{o} - \widetilde{\boldsymbol{o}}
    \right\|_2
    \leq
    2V_{\max}
    \frac{
    \sum_{C\in\mathcal C}
    \widehat p_C
    \left(
    \cosh(\delta_C)-1
    \right)
    }{
    \sum_{C\in\mathcal C}
    \widehat p_C
    \cosh(\delta_C)
    }.
    }
    \label{eq:mass}
\end{equation}

\subsection{Intra-cluster value distortion}
For the intra-cluster value distortion term, define the exact within-cluster attention weights as
\begin{equation}
    w_{i\mid C}
    =
    \frac{e^{s_i}}{z_C},
    \qquad
    \sum_{i\in C} w_{i\mid C}=1.
\end{equation}
Then
\begin{align}
    \boldsymbol{v}_C-\bar{\boldsymbol{v}}_C
    &=
    \sum_{i\in C}w_{i\mid C}\boldsymbol{v}_i
    -
    \sum_{i\in C}\frac{1}{n_C}\boldsymbol{v}_i \\
    &=
    \sum_{i\in C}
    \left(
    w_{i\mid C}-\frac{1}{n_C}
    \right)\boldsymbol{v}_i.
\end{align}

Therefore,
\begin{align}
    \|\boldsymbol{v}_C-\bar{\boldsymbol{v}}_C\|_2
    &\leq
    \sum_{i\in C}
    \left|
    w_{i\mid C}-\frac{1}{n_C}
    \right|
    \|\boldsymbol{v}_i\|_2 \\
    &\leq
    V_C
    \sum_{i\in C}
    \left|
    w_{i\mid C}-\frac{1}{n_C}
    \right|.
\end{align}

By Lemma~\ref{lem:softmax-uniform}, since $|s_i-s_C|\leq\delta_C$, and
\begin{equation}
    w_{i\mid C}
    =
    \frac{e^{s_i-s_C}}
    {\sum_{j\in C}e^{s_j-s_C}},
\end{equation}
we have
\begin{equation}
    \sum_{i\in C}
    \left|
    w_{i\mid C}-\frac{1}{n_C}
    \right|
    \leq
    2\tanh\left(\frac{\delta_C}{2}\right).
\end{equation}
Hence,
\begin{equation}
    \|\boldsymbol{v}_C-\bar{\boldsymbol{v}}_C\|_2
    \leq
    2V_C
    \tanh\left(\frac{\delta_C}{2}\right).
\end{equation}

Finally,
\begin{align}
    \left\|
    \sum_{C\in\mathcal C}
    \widehat p_C(\boldsymbol{v}_C-\bar{\boldsymbol{v}}_C)
    \right\|_2
    &\leq
    \sum_{C\in\mathcal C}
    \widehat p_C
    \|\boldsymbol{v}_C-\bar{\boldsymbol{v}}_C\|_2 \\
    &\leq
    \boxed{
    2
    \sum_{C\in\mathcal C}
    \widehat p_C V_C
    \tanh\left(\frac{\delta_C}{2}\right).
    }
    \label{eq:value}
\end{align}
Combining Eq.~\ref{eq:mass} and Eq.~\ref{eq:value} via the triangle inequality, we have
\begin{align}
    \left\|
    \boldsymbol{o}-\widehat{\boldsymbol{o}} 
    \right\|_2
    &\leq \left\|\boldsymbol{o}-\widetilde{\boldsymbol{o}}\right\|_2 + \left\|\widetilde{\boldsymbol{o}}-\widehat{\boldsymbol{o}}\right\|_2 \\
    &\leq 2V_{\max}
    \frac{
    \sum_{C\in\mathcal C}
    \widehat p_C
    \left(
    \cosh(\delta_C)-1
    \right)
    }{
    \sum_{C\in\mathcal C}
    \widehat p_C
    \cosh(\delta_C)
    } + 2
    \sum_{C\in\mathcal C}
    \widehat p_C V_C
    \tanh\left(\frac{\delta_C}{2}\right) \\
    &\boxed{= 2 \sum_{C\in\mathcal C}\widehat p_C \left ( V_{\max}\frac{
    \cosh(\delta_C)-1
    }{
    \sum_{C'\in\mathcal C}
    \widehat p_{C'}
    \cosh(\delta_{C'})
    }
    + V_C
    \tanh\left(\frac{\delta_C}{2}\right) 
    \right )}.
\end{align}

The maximum logit deviation $\delta_C$, however, cannot be used directly for access decisions, since evaluating it requires inspecting all keys within cluster $C$, defeating the purpose of cluster-level selection. We therefore construct an efficiently computable query-dependent upper bound $\delta_C^{\mathrm{bound}}$ from compact cluster statistics, such that
\begin{equation}
\delta_C \leq \delta_C^{\mathrm{bound}}.
\end{equation}
The specific construction of $\delta_C^{\mathrm{bound}}$ is deferred to Appendix~\ref{app:deltabound}. Since the above error bound is monotonically non-decreasing with respect to each $\delta_C$, replacing $\delta_C$ with $\delta_C^{\mathrm{bound}}$ yields a computable upper bound:
\begin{equation}
\lVert \boldsymbol{o}-\widehat{\boldsymbol{o}} \rVert_2
\leq
2 \sum_{C\in\mathcal C}
\underbrace{\widehat p_C \left( V_{\max} \frac{\cosh(\delta_C^{\mathrm{bound}})-1}{\sum_{C'\in\mathcal C} \widehat p_{C'} \cosh(\delta_{C'}^{\mathrm{bound}})}
+ V_C \tanh\left(\frac{\delta_C^{\mathrm{bound}}}{2}\right) \right)}_{\text{Risk Score $R_C$ for cluster \(C\)}}.
\end{equation}

We write the resulting upper bound in a clusterwise form and use each clusterwise term as its risk score $R_C$.

\subsection{Proof of core lemma}

\begin{proof}
We define absolute mass correction of cluster \(C\):
\begin{equation}
    \Delta z_C = z_C - \widehat z_C \geq 0 
\end{equation}
Proof of inequality can be done using Jensen's inequality with $1/n_C \sum s_i = s_C$. This revels that approximation underestimate the mass of cluster.

We have:
\begin{align}
        p_C - \widehat p_C 
        =& \frac{z_C}{Z} - \widehat p_C \\
        =& \frac{\Delta z_C + \widehat z_C - \widehat p_C Z}{Z} \\
        =& \frac{\Delta z_C + \widehat p_C \widehat Z - \widehat p_C Z}{Z} \\
        =& \underbrace{\frac{\Delta z_C}{Z}}_{\text{effect of cluster } C} - \underbrace{\frac{\widehat p_C \sum_{C'}\Delta z_{C'}}{Z}}_{\text{effect of changed gobal normlization}}.
\end{align}

If we define \(T_{C,C'}\) as the effect of approximating \(C'\) on cluster \(C\):

\begin{equation}
    T_{C,C'} = \frac{\Delta z_{C'}}{Z}(\mathbf{1}_{[C=C']} - \widehat p_C).
\end{equation}

Naturally we have:
\begin{equation}
    p_C - \widehat p_C  = \sum_{C'} T_{C,C'}
\end{equation}
And:
\begin{equation}
     \boldsymbol{o} - \widetilde{\boldsymbol{o}} = \sum_{C}(p_C-\widehat p_C)\boldsymbol{v}_C = \sum_{C}\sum_{C'} T_{C,C'} \boldsymbol{v}_C
\end{equation}

As analysed before, we want to know how much output distortion originates from approximating a particular cluster:
\begin{align}
    \boldsymbol{o} - \widetilde{\boldsymbol{o}} 
    =& \sum_{C'}\sum_{C} T_{C,C'} \boldsymbol{v}_C\\
    =& \sum_{C'}\sum_{C}\frac{\Delta z_{C'}}{Z}(\mathbf{1}_{[C=C']} - \widehat p_C)\boldsymbol{v}_C \\
    =&\sum_{C'}\frac{\Delta z_{C'}}{Z}\left(\boldsymbol{v}_{C'} - \sum_C \widehat p_C \boldsymbol{v}_C\right)\\
    =&\sum_{C'}\frac{\Delta z_{C'}}{Z}(\boldsymbol{v}_{C'} - \widetilde{\boldsymbol{o}})
\end{align}

We define mass correction factor $m_C$ as
\begin{equation}
    m_C = \frac{z_C}{\widehat z_C}=\frac{1}{n_C}\sum_{i\in C}e^{s_i-s_C}  \geq 1
\end{equation}

The total mass correction factor \(M\) can be defined as:
\begin{equation}
        M
    =\frac{Z}{\widehat Z} 
    =\frac{\sum_{C\in\mathcal C} z_C}{\widehat Z} 
    =\frac{\sum_{C\in\mathcal C} m_C \widehat z_C}{\widehat Z}
    =\sum_{C\in\mathcal C} \widehat p_C m_C
\end{equation}

We have:
\begin{align}
    \boldsymbol{o} - \widetilde{\boldsymbol{o}} 
    =&\sum_{C}\frac{\Delta z_{C}}{Z}(\boldsymbol{v}_{C} - \widetilde{\boldsymbol{o}})\\
    =&\sum_{C}\frac{\widehat z_{C}(m_C-1)}{\widehat Z M}(\boldsymbol{v}_{C} - \widetilde{\boldsymbol{o}})\\
    =&\sum_C \frac{\widehat p_C(m_C-1)}{M}(\boldsymbol{v}_C-\widetilde{\boldsymbol{o}})
\end{align}
\end{proof}

\subsection{Proof of technical lemmas}
\begin{lemma}[Cluster Mass Correction Bound]
\label{lem:mass_bound}
For any cluster $C \in \mathcal{C}$, let the query-dependent logit deviation be defined as
\begin{equation}
    \delta_C \coloneqq \max_{i\in C} |s_i - s_C| = \max_{i\in C} \left| \frac{\boldsymbol{q}^\top (\boldsymbol{k}_i - \bar{\boldsymbol{k}}_C)}{\sqrt{d}} \right|.
\end{equation}
Then, the local mass correction factor $m_C \coloneqq \frac{1}{n_C} \sum_{i\in C} e^{s_i - s_C}$ satisfies
\begin{equation}
    m_C \leq \cosh(\delta_C).
\end{equation}
\end{lemma}

\begin{proof}
Let $x_i = s_i - s_C$. By definition of $s_C$, the mean of $\{x_i\}_{i\in C}$ vanishes:
\begin{equation}
    \frac{1}{n_C} \sum_{i\in C} x_i = \frac{1}{n_C} \sum_{i\in C} s_i - s_C = 0.
\end{equation}
Moreover, by definition of $\delta_C$, we have $x_i \in [-\delta_C, \delta_C]$ for all $i \in C$. By the convexity of the exponential function $x \mapsto e^x$, any $x \in [-\delta_C, \delta_C]$ can be upper-bounded via the secant line between the endpoints:
\begin{equation}
    e^x \leq \frac{\delta_C + x}{2\delta_C} e^{\delta_C} + \frac{\delta_C - x}{2\delta_C} e^{-\delta_C}.
\end{equation}
Averaging this inequality over all tokens in cluster $C$, we obtain
\begin{align}
    m_C &= \frac{1}{n_C} \sum_{i\in C} e^{x_i} \\
    &\leq \frac{1}{n_C} \sum_{i\in C} \left[ \frac{\delta_C + x_i}{2\delta_C} e^{\delta_C} + \frac{\delta_C - x_i}{2\delta_C} e^{-\delta_C} \right] \\
    &= \frac{\delta_C + \frac{1}{n_C}\sum_{i\in C} x_i}{2\delta_C} e^{\delta_C} + \frac{\delta_C - \frac{1}{n_C}\sum_{i\in C} x_i}{2\delta_C} e^{-\delta_C} \\
    &= \frac{e^{\delta_C} + e^{-\delta_C}}{2} \\
    &= \cosh(\delta_C),
\end{align}
which completes the proof.
\end{proof}

\begin{lemma}[Softmax deviation from uniform]
\label{lem:softmax-uniform}
Let $x_1,\ldots,x_n\in\mathbb{R}$ satisfy
\begin{equation}
    |x_i|\leq \delta,
    \qquad i=1,\ldots,n,
\end{equation}
and define
\begin{equation}
    w_i
    =
    \frac{e^{x_i}}{\sum_{j=1}^n e^{x_j}}.
\end{equation}
Then
\begin{equation}
    \boxed{
    \sum_{i=1}^n
    \left|
    w_i-\frac{1}{n}
    \right|
    \leq
    2\tanh\left(\frac{\delta}{2}\right).
    }
\end{equation}
\end{lemma}

\begin{proof}
Since $x_i\in[-\delta,\delta]$, for any $i,j$,
\begin{equation}
    \frac{w_i}{w_j}
    =
    e^{x_i-x_j}
    \leq
    e^{2\delta}.
\end{equation}
Let
\begin{equation}
    r_i=nw_i,
\end{equation}
so that
\begin{equation}
    \frac{1}{n}\sum_{i=1}^n r_i=1.
\end{equation}
Denote
\begin{equation}
    a=\min_i r_i,
    \qquad
    b=\max_i r_i.
\end{equation}
Then
\begin{equation}
    0<a\leq 1\leq b,
    \qquad
    \frac{b}{a}\leq e^{2\delta}.
\end{equation}

For any random variable $R\in[a,b]$ with $\mathbb{E}[R]=1$,
the convexity of $|R-1|$ implies
\begin{equation}
    \frac{1}{2}\mathbb{E}|R-1|
    \leq
    \frac{(1-a)(b-1)}{b-a}.
\end{equation}
Applying this to the empirical distribution over
$\{r_i\}_{i=1}^n$ gives
\begin{equation}
    \frac{1}{2}
    \sum_{i=1}^n
    \left|
    w_i-\frac{1}{n}
    \right|
    \leq
    \frac{(1-a)(b-1)}{b-a}.
\end{equation}

For a fixed ratio $R=b/a$, the right-hand side is maximized at
\begin{equation}
    a=\frac{1}{\sqrt{R}},
    \qquad
    b=\sqrt{R},
\end{equation}
which yields
\begin{equation}
    \frac{(1-a)(b-1)}{b-a}
    \leq
    \frac{\sqrt{R}-1}{\sqrt{R}+1}.
\end{equation}
Since $R\leq e^{2\delta}$,
\begin{equation}
    \frac{\sqrt{R}-1}{\sqrt{R}+1}
    \leq
    \frac{e^\delta-1}{e^\delta+1}
    =
    \tanh\left(\frac{\delta}{2}\right).
\end{equation}
Therefore,
\begin{equation}
    \sum_{i=1}^n
    \left|
    w_i-\frac{1}{n}
    \right|
    \leq
    2\tanh\left(\frac{\delta}{2}\right).
\end{equation}
\end{proof}

\subsection{Computable upper bound for $\delta_C$.}
\label{app:deltabound}
The quantity $\delta_C$ directly measures the worst-case logit deviation
within cluster $C$ under the current query:
\begin{equation}
    \delta_C
    =
    \max_{i\in C}
    \left|
    \frac{\boldsymbol{q}^\top(\boldsymbol{k}_i-\bar{\boldsymbol{k}}_C)}{\sqrt d}
    \right|.
\end{equation}
While $\delta_C$ provides a natural measure of local fidelity, computing it
exactly requires accessing all keys in the cluster, which defeats the purpose
of efficient selection. We therefore seek a query-dependent upper bound
$D_C(\boldsymbol{q})$ such that
\begin{equation}
    \boxed{
    \delta_C^{\mathrm{bound}}(\boldsymbol{q})
    \geq
    \delta_C.
    }
\end{equation}
\subsubsection{Axis-aligned box bound.}

Let \(\boldsymbol{k}_C^{\min}\) and \(\boldsymbol{k}_C^{\max}\) be the coordinate-wise minimum and
maximum keys in cluster \(C\), and define
\begin{equation}
    s_C=\frac{\boldsymbol{q}^\top\bar{\boldsymbol{k}}_C}{\sqrt d}.
\end{equation}
Since every \(\boldsymbol{k}_i\), \(i\in C\), lies inside the box
\([\boldsymbol{k}_C^{\min},\boldsymbol{k}_C^{\max}]\), its query-key score is bounded by
\begin{align}
    s_C^{\mathrm{upper}}
    &=
    \frac{1}{\sqrt d}
    \sum_{j=1}^{d}
    \max\left\{
        q_j k_{C,j}^{\min},
        q_j k_{C,j}^{\max}
    \right\},\\
    s_C^{\mathrm{lower}}
    &=
    \frac{1}{\sqrt d}
    \sum_{j=1}^{d}
    \min\left\{
        q_j k_{C,j}^{\min},
        q_j k_{C,j}^{\max}
    \right\}.
\end{align}
Therefore,
\begin{equation}
    \boxed{
    \delta_C^{\mathrm{box}}(\boldsymbol{q})
    =
    \max\left\{
        |s_C^{\mathrm{upper}}-s_C|,
        |s_C^{\mathrm{lower}}-s_C|
    \right\}
    \geq \delta_C.
    }
\end{equation}
This bound requires two \(d\)-dimensional vectors per cluster and \(O(d)\)
query-time computation. It can be loose because its coordinate-wise extrema
may come from different keys.

\subsubsection{Diagonal-Ellipsoid Bound}
\label{app:diag-ell-bound}

For a centered key $\boldsymbol{r}_i=\boldsymbol{k}_i-\bar{\boldsymbol{k}}_C$, define the coordinate-wise radii
\begin{equation}
    w_{C,j}=\max_{i\in C}|r_{i,j}|.
    \label{eq:diag-ell-width}
\end{equation}
For clarity, we assume $w_{C,j}>0$; a zero-width coordinate has zero residual
for every key and can be omitted without affecting the bound.  We further
define
\begin{equation}
    \rho_C
    =
    \max_{i\in C}
    \left(
        \sum_{j=1}^{d_k}\frac{r_{i,j}^2}{w_{C,j}^2}
    \right)^{1/2}.
    \label{eq:diag-ell-rho}
\end{equation}
The vector $\boldsymbol{w}_C$ normalizes the scale of each coordinate, while $\rho_C$
measures the largest residual after this normalization.  Equivalently, all
centered keys lie inside the diagonal ellipsoid
\begin{equation}
    \sum_{j=1}^{d_k}\frac{r_j^2}{w_{C,j}^2}\le \rho_C^2.
\end{equation}
The resulting query-dependent bound is
\begin{equation}
    \boxed{
    \delta_C^{\mathrm{ell}}(\boldsymbol{q})
    =
    \frac{\rho_C}{\sqrt{d_k}}
    \left(
        \sum_{j=1}^{d_k}q_j^2w_{C,j}^2
    \right)^{1/2}}
    .
    \label{eq:diag-ell-delta}
\end{equation}
Indeed, for every $i\in C$, weighted Cauchy--Schwarz gives
\begin{align}
    |\boldsymbol{q}^\top \boldsymbol{r}_i|
    &=
    \left|
        \sum_{j=1}^{d_k}
        (q_j w_{C,j})\frac{r_{i,j}}{w_{C,j}}
    \right| \\
    &\le
    \left(\sum_{j=1}^{d_k}q_j^2w_{C,j}^2\right)^{1/2}
    \left(\sum_{j=1}^{d_k}\frac{r_{i,j}^2}{w_{C,j}^2}\right)^{1/2} \\
    &\le
    \rho_C
    \left(\sum_{j=1}^{d_k}q_j^2w_{C,j}^2\right)^{1/2}.
\end{align}
Therefore,
\begin{equation}
    \delta_C^{\mathrm{ell}}(\boldsymbol{q})
    \ge
    \max_{i\in C}
    \frac{|\boldsymbol{q}^\top(\boldsymbol{k}_i-\bar{\boldsymbol{k}}_C)|}{\sqrt{d_k}}.
\end{equation}
Both $\boldsymbol{w}_C$ and $\rho_C$ are query-independent and can be precomputed once per
block.  The bound stores one vector and one scalar per block and takes
$O(d_k)$ operations to evaluate for each query--block pair.

\section{\method Details}
\subsection{Q-anchor}
\label{app:qanchor}
\textbf{Prompt-Query Anchors.} During prefill, we use the query vectors from the last \(Q\) prompt positions as anchors and cache the clusters they select. These anchor-selected clusters remain expanded throughout decoding, in addition to those selected by the current query. We use no Q anchors in main text, but we provide result for Q-anchors on Appendix \ref{app:maintable}. The anchor mask is computed once during prefill and reused throughout decoding. Table~\ref{tab:q-anchor-ablation} quantifies the resulting quality–latency trade-off.

\paragraph{Ablation for Q-anchor.} 
We additionally perform ablation on the number of query anchors. The results are shown in Table~\ref{tab:q-anchor-ablation}.


\begin{table*}[ht]
\centering
\caption{Query-anchor ablation of Llama-3.1-8B-Instruct on RULER-128K (260 samples).
All variants use the default configuration ($\tau{=}0.01$, $|C|{=}32$, box summary, GQA-any) and differ only in
the number of query anchors $Q$; $Q{=}0$ (no query anchors) is our default and is shown in bold.
Latency is steady-state TPOT with CUDA Graph (130{,}944 input / 128 output tokens).}
\label{tab:q-anchor-ablation}

\renewcommand{\arraystretch}{1.12}
\begin{tabular}{@{}cccccc@{}}
\toprule
\multirow{2}{*}{Metric}
& Ref.
& \multicolumn{4}{c}{Number of query anchors $Q$} \\
\cmidrule(lr){2-2}
\cmidrule(lr){3-6}
& Full & $\mathbf{0}$ (default) & $16$ & $32$ & $64$ \\
\midrule
Acc.$\uparrow$
& $73.37$ & $\mathbf{70.19}$ & $72.49$ & $72.99$ & $72.91$ \\
Steady TPOT (ms)$\downarrow$
& $24.43$ & $\mathbf{15.68}$ & $16.65$ & $17.14$ & $17.84$ \\
Speedup$\uparrow$
& $1.00\times$ & $\mathbf{1.56\times}$ & $1.47\times$ & $1.42\times$ & $1.37\times$ \\
\bottomrule
\end{tabular}
\end{table*}
\subsection{Kernel Implementation}
\label{app:kernel-implementation}

We implement \method using custom Triton kernels. After prefill, the prompt K/V cache is partitioned into blocks of 32 tokens. For each block, we compute and cache its key and value representatives, diagonal-ellipsoid bound statistics, maximum value norm, and valid-token count. Prompt-anchor queries are captured during prefill, and their selected blocks are computed once and stored as a compact per-KV-head mask. These summaries and masks are reused throughout generation.

At each decoding step, the block-selection kernel jointly processes the query heads associated with each KV head. It computes the representative attention score and deviation bound for every block, applies the expansion criterion of \method, and combines the resulting decisions with the cached prompt-anchor mask. The output is a compact selection mask that remains on the GPU. The sparse-attention kernel then reads token-level K/V directly from the original cache only for selected blocks; unselected blocks contribute through their cached representatives, while generated suffix tokens are always processed exactly. The three types of contributions are merged using online softmax, with BF16 Tensor-Core operations for matrix products and FP32 accumulation for numerical stability. We use a preallocated static KV cache and reuse all intermediate workspaces across tokens. For fixed-length latency evaluation, the complete decoding step is captured in a CUDA Graph, reducing Python and kernel-launch overhead to one graph replay per token. For fair comparison, full Attention, and baselines are timed under the same CUDA Graph protocol. We provide additional eager results on Appendix \ref{app:eager}.

\subsection{Correlation Analysis}
\label{app:correlation}

\paragraph{Definition of the target.} For a cluster $C$ and query $q$, let

$$o_C = \frac{1}{Z}\sum_{i\in C} e^{q^\top k_i/\sqrt d}\, v_i, \qquad \hat o_C = \hat p_C\, \bar v_C, \qquad \hat p_C = \frac{|C|\, e^{q^\top \bar k_C/\sqrt d}}{\sum_{C'} |C'|\, e^{q^\top \bar k_{C'}/\sqrt d}},$$

where $Z$ is the Full Attention softmax normalizer. $o_C$ is $C$’s contribution to the Full Attention output, and $\hat o_C$ is its contribution to the output in which every cluster is replaced by its representative. Hence $\sum_C (o_C - \hat o_C)$ is exactly the error of that approximation, and $\lVert o_C - \hat o_C\rVert$ is $C$’s share of it.

\paragraph{Correlation analysis.} 
We use Llama-3.1-8B on RULER-128K (26 prompts) in the main setting ($B=32$, box, $\tau=.01$). We take every layer and head at decode steps 0, 31 and 126. For each (step, layer, head), we compute the Spearman correlation between a score and the target across clusters, and we report the median. $\hat p_C$ alone reaches .975, so the target is dominated by attention mass. We therefore also report the \textbf{partial Spearman correlation given $\hat p_C$}.

Within each instance, we rank-transform the score $s$, the target $y$ and $\hat p_C$ across clusters. We then regress the ranks of $s$ and of $y$ on the rank of $\hat p_C$ by least squares, and take the Pearson correlation of the two residuals. Equivalently,

$$\rho_{sy\cdot \hat p} = \frac{\rho_{sy} - \rho_{s\hat p}\,\rho_{y\hat p}}{\sqrt{(1-\rho_{s\hat p}^2)(1-\rho_{y\hat p}^2)}},$$

where each $\rho$ is a Spearman correlation. This measures how much a score explains about the error beyond attention mass. On this target, $R_C$ carries clear information beyond mass (partial $\rho = .49$), more than a Quest-style bound (.32). The mass-weighted value norm $\hat p_C \lVert \bar v_C\rVert$ carries almost none (.06). With the exact deviation in place of the box summary, the partial correlation rises to .81. The deviation term is therefore what separates $R_C$ from an importance score, and a tighter deviation summary would sharpen it further. 
\begin{table}[h]
\centering
\caption{Spearman correlation between each score and $\lVert o_C - \hat o_C\rVert$ (Llama-3.1-8B, RULER-128K, $B=32$, box, $\tau=.01$). Values are medians over (step, layer, head). ``Partial'' is the correlation given $\hat p_C$.}
\label{tab:partial-corr}
\begin{tabular}{lcc}
\toprule
Score & Raw & Partial \\
\midrule
$\hat p_C$                          & .975 & --  \\
$\hat p_C \lVert \bar v_C \rVert$   & .975 & .06 \\
Quest-style bound                   & .864 & .32 \\
$R_C$ (box $\delta$)                & .975 & .49 \\
$R_C$ (exact $\delta$, oracle)      & .990 & .81 \\
\bottomrule
\end{tabular}
\end{table} 
\section{Experiment Details}
\subsection{Critical KV Analysis}
\label{app:kvimportant}
\paragraph{Top-$p$ KV analysis.}
For the top-$p$ analysis, we evaluate all 32 layers of Llama-3.1-8B on 20 non-overlapping 1,024-token segments from the WikiText-2 test split. For each query head, we select the smallest KV subset that accumulates attention mass $p\in\{0,0.25,0.5,0.75,0.9,0.95,0.99,1\}$. Because the model uses GQA, the four query heads sharing a KV head access the union of their selected subsets. We report the average per-query KV percentage, the union across all queries, and the union across distant queries satisfying $q-k\geq64$. Output error is measured after renormalizing attention over the selected KVs as
$\lVert O_p-O_{\mathrm{full}}\rVert_2/\lVert O_{\mathrm{full}}\rVert_2$.

\paragraph{Perplexity evaluation.}
For the perplexity experiment, we use 20 non-overlapping 2,048-token WikiText-2 segments, totaling 40,940 next-token predictions. We compare our query-adaptive method against oracle top-$k$ and the query-independent H2O, RoCo, and TOVA eviction methods. The horizontal axis reports the measured fraction of KV entries accessed relative to full causal attention, rather than the nominal cache budget. All results are averaged across samples, and error bars show 95\% confidence intervals.

\subsection{Main Experiment Setup}
\label{app:main-exp}

\subsubsection{Models}
\label{app:main-exp-model}
We evaluate \texttt{Llama-3.1-8B-Instruct}~\citep{grattafiori2024llama3herdmodels} and \texttt{Qwen2.5-7B-Instruct}~\citep{qwen2025qwen25technicalreport} without additional training or fine-tuning. Llama-3.1-8B contains 32 transformer layers with 32 query heads and 8 KV heads, corresponding to four query heads per KV head. It uses the rotary position encoding~\citep{su2023roperoformerenhancedtransformerrotary} with a native context length of 128K. 

Qwen2.5-7B contains 28 layers with 28 query heads and 4 KV heads, resulting in a larger GQA group of seven query heads per KV head. Qwen2.5 natively
supports a 32K context window with its default RoPE configuration. For the
64K and 128K evaluations, we follow the officially recommended long-context
setting by enabling static YaRN\citep{peng2026yarnefficientcontextwindow} with a scaling factor of four and an original
context length of 32K. This positional-encoding extension is applied
identically to all methods. Both models have a head dimension of 128 and are evaluated in BF16 with batch size one and greedy decoding.

\subsubsection{Benchmarks}
\label{app:main-exp-benchmark}
For LongBench~\citep{bai2024longbenchbilingualmultitaskbenchmark}, we use the 16-task English/code subset consisting of 14 English tasks, \texttt{lcc}, and \texttt{repobench-p}, with 3,750 test examples per configuration. Quality is evaluated on LongBench using the official prompts, generation limits, and task-specific metrics, while latency is measured separately using standardized 128K prompts. 

For RULER~\citep{hsieh2024rulerwhatsrealcontext}, we evaluate all 13 official tasks at context length of 128K, using the same 20 source examples per task and length, for a total of 260 examples per configuration. RULER data are generated separately for each model and tokenizer. 

For HELMET~\citep{yen2025helmetevaluatelongcontextlanguage}, we include all tasks with objective automatic evaluation and exclude only subjective tasks that require an LLM-as-a-judge. We follow the official task-specific formatting, generation limits, stopping rules, and evaluation metrics. For all benchmarks, we compute each task score independently and report the unweighted macro-average across tasks.

\subsubsection{Baselines}
\label{app:main-exp-Baselines}
\paragraph{Full Attention.}
Full Attention retains the complete visible KV cache and serves as the reference for both quality and latency. We use the model's native GQA layout and optimized scaled dot-product attention without replicating KV heads. The latency implementation uses a preallocated static KV cache and captures each decoding step in a CUDA Graph, under the same BF16 and batch-size-one protocol as RefineKV and the eviction baselines.

\paragraph{SnapKV.}
We use the KVPress~\citep{devoto2025kvpressexpectedattention} implementation of SnapKV with an observation window of 64 tokens and a pooling kernel size of 5. We evaluate target KV-retention ratios of 10\%, 25\%, and 50\%, where the percentage denotes the fraction of prompt KV states retained after prefill compression. Decode attention operates directly on the compressed cache, while newly generated tokens are retained.

\paragraph{StreamLLM.}
We use the KVPress~\citep{devoto2025kvpressexpectedattention} implementation of StreamLLM with four attention-sink tokens. After preserving these sink tokens, the remaining budget is assigned to the most recent prompt tokens. We evaluate prompt-KV retention ratios of 10\%, 25\%, and 50\%. The prompt cache is compressed after prefill, and all newly generated tokens are retained during decoding.

\paragraph{PyramidKV.}
We configure PyramidKV with an observation window of 64 tokens, a pooling kernel size of 5, and $\beta=20$. We evaluate mean retention targets of 10\%, 25\%, and 50\%; because PyramidKV distributes its budget non-uniformly across layers, these percentages denote the configured average retention rather than an identical per-layer budget. The selected prompt KVs are stored in a visible-prefix static cache and processed with native-GQA attention; the decoding step is captured in a CUDA Graph.

\paragraph{Quest.}
We use a page size of 16 tokens and evaluate absolute token budgets of 512, 1,024, 2,048, and 4,096 tokens. Following the official setting, the first two transformer layers use Full Attention, while the remaining layers apply query-dependent page selection. The mandatory tail page is included in the reported token budget. We use paged CUDA decoding for Llama and a native-head parallel page-scoring implementation for Qwen, without physically replicating GQA KV heads. 

\paragraph{Double-P.}
We use clusters of 32 prompt tokens, four iterations of per-KV-head $k$-means, four exact sink tokens, and a 64-token exact local window. We evaluate the threshold pairs $(p_1,p_2)\in\{(0.85,0.50),(0.85,0.70),(0.90,0.60),(0.95,0.70)\}$ when complete measurements are available. Clusters outside the $p_1$ probability prefix are pruned, clusters between the $p_1$ and $p_2$ prefixes are represented by their centroids, and clusters inside the $p_2$ prefix are expanded to exact tokens. Prompt clustering is performed once after prefill, and generated tokens remain exact. Since the authors' production kernels are \textit{not publicly available}, we use our optimized Triton reimplementation and do not describe it as the official Double-P implementation. 

\subsubsection{Metrics.}
\label{app:main-exp-metric}
We focus on steady-state decoding because \method primarily targets the repeated KV-cache access during autoregressive generation. Prefill processes all prompt tokens in parallel through regular, highly optimized matrix--matrix multiplications, whereas batch-one decoding executes sequentially for each generated token and is often dominated by memory-intensive KV-cache access. Prefill is therefore a one-time, highly parallelized cost, while decode latency accumulates with the output length. As shown in Table~\ref{tab:prefill-decode-profile}, our auxiliary experiment on Llama-3.1-8B shows that decoding accounts for 82.2\% with 32K  prefill/512 decode setting, and 90.3\% with 32K/1024 setting. We thus evaluate serving efficiency using decode TPOT and treat prefill latency and time to first token as separate costs. Unless noted otherwise, all methods and Full Attention are timed with CUDA Graph; eager-mode results are in Appendix \ref{app:eager}.

\begin{table}[ht]
    \centering
    \small
    \caption{
    Latency breakdown of prefill and decoding on Llama-3.1-8B in BF16.
    Each result is averaged over three runs.
    }
    \label{tab:prefill-decode-profile}
    \begin{tabular}{ccrrrr}
        \toprule
        \multicolumn{2}{c}{Sequence length} &
        \multicolumn{2}{c}{Latency (s)} &
        \multicolumn{2}{c}{Fraction (\%)} \\
        \cmidrule(lr){1-2}
        \cmidrule(lr){3-4}
        \cmidrule(lr){5-6}
        Prefill & Decode &
        Prefill & Decode &
        Prefill & Decode \\
        \midrule
        16K & 512  & 1.01 & 9.20  & 9.9  & 90.1 \\
        32K & 512  & 2.42 & 11.15 & 17.8 & 82.2 \\
        32K & 1024 & 2.41 & 22.37 & 9.7  & 90.3 \\
        \bottomrule
    \end{tabular}
\end{table}

For every method, we disable early stopping and generate exactly 128 output tokens. We exclude the output token produced during prefill, as it does not require an autoregressive decoding step, and discard the first post-prefill iteration to remove one-time initialization overhead. Let $T_{\mathrm{steady}}$ denote the elapsed time of the remaining 126 decoding iterations. We compute
\[
    \mathrm{TPOT} = \frac{T_{\mathrm{steady}}}{126}\ \text{ms/token}.
\]
The timed region includes all method-specific operations, including token selection or routing, attention computation, Transformer execution, KV-cache updates, and generation-loop overhead, while excluding model loading, compilation, warm-up, and prefill.

\begin{table}[ht]
    \centering
    \small
    \caption{Sensitivity of steady-state TPOT to the generation length.
    Values in parentheses denote the relative change from 128 tokens.
    All lengths use identical prompts, cache capacities, and method configurations.}
    \label{tab:generation-length}
    \begin{tabular}{lccc}
        \toprule
        & \multicolumn{3}{c}{TPOT (ms/token)} \\
        \cmidrule(lr){2-4}
        Benchmark
        & 128 tokens
        & 256 tokens
        & 512 tokens \\
        \midrule
        LongBench
        & 14.76
        & 14.61 {\scriptsize($-1.03\%$)}
        & 14.55 {\scriptsize($-1.43\%$)} \\
        RULER
        & 16.35
        & 16.20 {\scriptsize($-0.92\%$)}
        & 16.14 {\scriptsize($-1.27\%$)} \\
        HELMET
        & 17.20
        & 17.06 {\scriptsize($-0.80\%$)}
        & 17.01 {\scriptsize($-1.07\%$)} \\
        \bottomrule
    \end{tabular}
\end{table}
\paragraph{Choice of generation length.}
We report latency using 128 generated tokens. As shown in
Table~\ref{tab:generation-length}, increasing the generation length to 256 or
512 tokens changes steady-state TPOT by at most 1.43\% on average across all
three benchmarks; the largest per-prompt change is 1.69\%, within our
predefined 3\% practical threshold. Thus, 128 tokens are sufficient to obtain
a stable estimate of steady-state decoding efficiency, while longer
generations substantially increase evaluation cost without materially
changing the conclusion. We therefore use 128 tokens in the main experiments.

\subsubsection{Hardware Settings.}
\label{app:main-exp-hardware}
All experiments are conducted on NVIDIA RTX PRO 6000 Blackwell Server Edition GPUs. Each latency measurement uses a single exclusive GPU. Unless noted otherwise, all methods use bfloat16 and batch size one.

\subsection{Extra Experiment Results}
\subsubsection{main table tasks}
\label{app:maintable}
For compact presentation, we aggregate semantically related tasks into task groups; no tasks are cherry-picked or omitted, and every task is included exactly once. For LongBench, Doc QA combines single-document and multi-document question answering, Summ. denotes long-document summarization, Few-shot denotes few-shot learning, Synth. covers synthetic counting and retrieval, and Code evaluates long-context code understanding and completion. For RULER, NIAH combines single-key, multi-key, multi-value, and multi-query needle-in-a-haystack retrieval, A\&T denotes aggregation and tracking, and QA denotes question answering. For HELMET, RAG denotes retrieval-augmented generation, LC-QA denotes long-context question answering, ICL denotes in-context learning, and Rec./Rank. combines recall and reranking tasks. We put the detail for subtask result in: Table~\ref{tab:longbench-per-task-acc-accl}, Table~\ref{tab:ruler-per-task-acc-accl} and Table~\ref{tab:helmet-per-task-acc-accl}.

\providecolor{lightgreen}{RGB}{239,248,239}
\begin{table*}[t]
\centering
\caption{LongBench per-task quality--latency comparison. Each cell reports $\mathrm{Accuracy}/\mathrm{Acceleration}$, where $\mathrm{Accuracy}$ is the task-macro benchmark score and $\mathrm{Acceleration}$ is latency acceleration relative to Full, higher is better for both. \protect\colorbox{lightgreen}{Shaded} cells are Pareto-optimal. \underline{Underlined} cells have an absolute $\mathrm{Accuracy}$ drop versus Full of less than $0.5$ point. Baselines use SnapKV/StreamLLM/PyramidKV.}
\label{tab:longbench-per-task-acc-accl}
\setlength{\tabcolsep}{1.0pt}
\renewcommand{\arraystretch}{1.10}
\small
\resizebox{\textwidth}{!}{%
\begin{tabular}{@{}c@{\hspace{0.5pt}}l@{\hspace{0.5pt}}c@{\hspace{0.5pt}}*{15}{c}@{}}
\toprule
 & Method & \multicolumn{3}{c}{S-Doc QA} & \multicolumn{3}{c}{M-Doc QA} & \multicolumn{3}{c}{Summ.} & \multicolumn{3}{c}{Few-shot} & \multicolumn{2}{c}{Synth.} & \multicolumn{2}{c}{Code} \\[-2pt]
\cmidrule(lr){3-5}\cmidrule(lr){6-8}\cmidrule(lr){9-11}\cmidrule(lr){12-14}\cmidrule(lr){15-16}\cmidrule(l){17-18}
 & & \rotatebox{60}{\scriptsize NarrativeQA} & \rotatebox{60}{\scriptsize Qasper} & \rotatebox{60}{\scriptsize MultiFieldQA} & \rotatebox{60}{\scriptsize HotpotQA} & \rotatebox{60}{\scriptsize 2WikiMQA} & \rotatebox{60}{\scriptsize MuSiQue} & \rotatebox{60}{\scriptsize GovReport} & \rotatebox{60}{\scriptsize QMSum} & \rotatebox{60}{\scriptsize MultiNews} & \rotatebox{60}{\scriptsize TREC} & \rotatebox{60}{\scriptsize TriviaQA} & \rotatebox{60}{\scriptsize SAMSum} & \rotatebox{60}{\scriptsize Count} & \rotatebox{60}{\scriptsize Retrieval} & \rotatebox{60}{\scriptsize LCC} & \rotatebox{60}{\scriptsize RepoBench-P} \\
\midrule
\multirow{12}{*}{\rotatebox[origin=c]{90}{\textbf{Llama}}} & Full & $29.8${\scriptsize $/1.0$} & $45.5${\scriptsize $/1.0$} & $54.7${\scriptsize $/1.0$} & $55.2${\scriptsize $/1.0$} & $45.0${\scriptsize $/1.0$} & $31.2${\scriptsize $/1.0$} & $35.0${\scriptsize $/1.0$} & $25.3${\scriptsize $/1.0$} & $27.2${\scriptsize $/1.0$} & $72.5${\scriptsize $/1.0$} & $91.7${\scriptsize $/1.0$} & $43.4${\scriptsize $/1.0$} & $6.5${\scriptsize $/1.0$} & $99.5${\scriptsize $/1.0$} & $63.1${\scriptsize $/1.0$} & $56.1${\scriptsize $/1.0$} \\
 & Snap $10\%$ & \cellcolor{lightgreen}\strut \underline{$30.3${\scriptsize $/1.7$}} & $36.5${\scriptsize $/1.7$} & $52.4${\scriptsize $/1.7$} & \cellcolor{lightgreen}\strut \underline{$54.9${\scriptsize $/1.7$}} & $42.7${\scriptsize $/1.7$} & \underline{$30.8${\scriptsize $/1.7$}} & $27.0${\scriptsize $/1.7$} & $23.4${\scriptsize $/1.7$} & $19.8${\scriptsize $/1.7$} & $63.0${\scriptsize $/1.7$} & \cellcolor{lightgreen}\strut \underline{$91.9${\scriptsize $/1.7$}} & \cellcolor{lightgreen}\strut $42.8${\scriptsize $/1.7$} & \cellcolor{lightgreen}\strut \underline{$7.2${\scriptsize $/1.7$}} & \cellcolor{lightgreen}\strut \underline{$99.5${\scriptsize $/1.7$}} & $60.8${\scriptsize $/1.7$} & \cellcolor{lightgreen}\strut \underline{$56.3${\scriptsize $/1.7$}} \\
 & Snap $25\%$ & \underline{$30.0${\scriptsize $/1.5$}} & $42.2${\scriptsize $/1.5$} & $54.2${\scriptsize $/1.5$} & \underline{$55.1${\scriptsize $/1.5$}} & $44.4${\scriptsize $/1.5$} & \underline{$31.1${\scriptsize $/1.5$}} & $30.4${\scriptsize $/1.5$} & \underline{$25.0${\scriptsize $/1.5$}} & $23.4${\scriptsize $/1.5$} & $68.5${\scriptsize $/1.5$} & \underline{$91.6${\scriptsize $/1.5$}} & $42.4${\scriptsize $/1.5$} & \underline{$6.9${\scriptsize $/1.5$}} & \underline{$99.5${\scriptsize $/1.5$}} & \underline{$62.8${\scriptsize $/1.5$}} & \underline{$56.8${\scriptsize $/1.5$}} \\
 & Str. $10\%$ & $26.3${\scriptsize $/1.7$} & $26.2${\scriptsize $/1.7$} & $33.7${\scriptsize $/1.7$} & $47.6${\scriptsize $/1.7$} & $37.0${\scriptsize $/1.7$} & $25.2${\scriptsize $/1.7$} & $25.4${\scriptsize $/1.7$} & $21.4${\scriptsize $/1.7$} & $19.7${\scriptsize $/1.7$} & $58.5${\scriptsize $/1.7$} & $88.8${\scriptsize $/1.7$} & $42.5${\scriptsize $/1.7$} & \underline{$6.4${\scriptsize $/1.7$}} & $88.5${\scriptsize $/1.7$} & $59.4${\scriptsize $/1.7$} & $52.9${\scriptsize $/1.7$} \\
 & Str. $25\%$ & $28.1${\scriptsize $/1.5$} & $31.3${\scriptsize $/1.5$} & $35.6${\scriptsize $/1.5$} & $50.4${\scriptsize $/1.5$} & $40.9${\scriptsize $/1.5$} & $25.3${\scriptsize $/1.5$} & $29.1${\scriptsize $/1.5$} & $22.0${\scriptsize $/1.5$} & $23.0${\scriptsize $/1.5$} & $66.0${\scriptsize $/1.5$} & $90.0${\scriptsize $/1.5$} & $42.4${\scriptsize $/1.5$} & \cellcolor{lightgreen}\strut \underline{$7.9${\scriptsize $/1.5$}} & $84.5${\scriptsize $/1.5$} & $60.9${\scriptsize $/1.5$} & $55.3${\scriptsize $/1.5$} \\
 & Pyr. $10\%$ & \cellcolor{lightgreen}\strut \underline{$30.3${\scriptsize $/1.7$}} & $36.0${\scriptsize $/1.7$} & $52.3${\scriptsize $/1.7$} & \cellcolor{lightgreen}\strut \underline{$54.8${\scriptsize $/1.7$}} & $42.7${\scriptsize $/1.7$} & \cellcolor{lightgreen}\strut \underline{$31.6${\scriptsize $/1.7$}} & $26.7${\scriptsize $/1.7$} & $24.2${\scriptsize $/1.7$} & $19.9${\scriptsize $/1.7$} & $63.0${\scriptsize $/1.7$} & \cellcolor{lightgreen}\strut \underline{$91.9${\scriptsize $/1.7$}} & \cellcolor{lightgreen}\strut \underline{$43.0${\scriptsize $/1.7$}} & \cellcolor{lightgreen}\strut \underline{$7.2${\scriptsize $/1.7$}} & \cellcolor{lightgreen}\strut \underline{$99.5${\scriptsize $/1.7$}} & $60.6${\scriptsize $/1.7$} & \cellcolor{lightgreen}\strut \underline{$56.1${\scriptsize $/1.7$}} \\
 & Pyr. $25\%$ & \cellcolor{lightgreen}\strut \underline{$30.5${\scriptsize $/1.5$}} & $42.5${\scriptsize $/1.5$} & $53.8${\scriptsize $/1.5$} & \cellcolor{lightgreen}\strut \underline{$55.4${\scriptsize $/1.5$}} & $44.4${\scriptsize $/1.5$} & \underline{$31.6${\scriptsize $/1.5$}} & $30.1${\scriptsize $/1.5$} & \underline{$24.9${\scriptsize $/1.5$}} & $23.6${\scriptsize $/1.5$} & $67.5${\scriptsize $/1.5$} & \underline{$91.5${\scriptsize $/1.5$}} & \underline{$43.6${\scriptsize $/1.5$}} & \underline{$6.4${\scriptsize $/1.5$}} & \underline{$99.5${\scriptsize $/1.5$}} & \underline{$62.8${\scriptsize $/1.5$}} & \underline{$56.9${\scriptsize $/1.5$}} \\
 & Quest & \underline{$30.5${\scriptsize $/1.1$}} & \underline{$45.1${\scriptsize $/1.1$}} & \underline{$55.5${\scriptsize $/1.1$}} & \underline{$55.3${\scriptsize $/1.1$}} & \cellcolor{lightgreen}\strut \underline{$45.9${\scriptsize $/1.1$}} & \cellcolor{lightgreen}\strut \underline{$32.8${\scriptsize $/1.1$}} & \cellcolor{lightgreen}\strut \underline{$35.8${\scriptsize $/1.1$}} & \underline{$25.8${\scriptsize $/1.1$}} & \underline{$27.2${\scriptsize $/1.1$}} & $72.0${\scriptsize $/1.1$} & \underline{$91.5${\scriptsize $/1.1$}} & \underline{$43.1${\scriptsize $/1.1$}} & \underline{$6.9${\scriptsize $/1.1$}} & \underline{$99.5${\scriptsize $/1.1$}} & \underline{$62.9${\scriptsize $/1.1$}} & \underline{$56.4${\scriptsize $/1.1$}} \\
 & Double & \cellcolor{lightgreen}\strut \underline{$30.6${\scriptsize $/1.4$}} & $43.8${\scriptsize $/1.4$} & \underline{$55.9${\scriptsize $/1.4$}} & \underline{$54.7${\scriptsize $/1.4$}} & $43.6${\scriptsize $/1.4$} & \underline{$31.4${\scriptsize $/1.4$}} & \underline{$34.6${\scriptsize $/1.4$}} & $24.5${\scriptsize $/1.4$} & \underline{$26.7${\scriptsize $/1.4$}} & $71.5${\scriptsize $/1.4$} & \underline{$91.6${\scriptsize $/1.4$}} & $42.9${\scriptsize $/1.4$} & \underline{$7.0${\scriptsize $/1.4$}} & $99.0${\scriptsize $/1.4$} & \cellcolor{lightgreen}\strut \underline{$63.5${\scriptsize $/1.4$}} & \underline{$55.6${\scriptsize $/1.4$}} \\
 & {\bfseries\boldmath Our $\tau.01$} & $\mathbf{29.2}${\scriptsize $\mathbf{/1.5}$} & \cellcolor{lightgreen}\strut \underline{$\mathbf{45.4}${\scriptsize $\mathbf{/1.6}$}} & \underline{$\mathbf{54.2}${\scriptsize $\mathbf{/1.6}$}} & \underline{$\mathbf{55.2}${\scriptsize $\mathbf{/1.6}$}} & \underline{$\mathbf{45.0}${\scriptsize $\mathbf{/1.6}$}} & \underline{$\mathbf{30.9}${\scriptsize $\mathbf{/1.6}$}} & \cellcolor{lightgreen}\strut \underline{$\mathbf{35.4}${\scriptsize $\mathbf{/1.6}$}} & \underline{$\mathbf{25.4}${\scriptsize $\mathbf{/1.5}$}} & \underline{$\mathbf{27.2}${\scriptsize $\mathbf{/1.5}$}} & \underline{$\mathbf{72.5}${\scriptsize $\mathbf{/1.5}$}} & \underline{$\mathbf{91.6}${\scriptsize $\mathbf{/1.5}$}} & \underline{$\mathbf{44.0}${\scriptsize $\mathbf{/1.5}$}} & \underline{$\mathbf{7.0}${\scriptsize $\mathbf{/1.6}$}} & \underline{$\mathbf{99.5}${\scriptsize $\mathbf{/1.6}$}} & \cellcolor{lightgreen}\strut \underline{$\mathbf{63.0}${\scriptsize $\mathbf{/1.6}$}} & \underline{$\mathbf{56.3}${\scriptsize $\mathbf{/1.6}$}} \\
 & {\bfseries\boldmath Our $\tau.1$} & \underline{$\mathbf{30.1}${\scriptsize $\mathbf{/1.7}$}} & $\mathbf{44.3}${\scriptsize $\mathbf{/1.7}$} & \cellcolor{lightgreen}\strut \underline{$\mathbf{56.3}${\scriptsize $\mathbf{/1.7}$}} & \cellcolor{lightgreen}\strut \underline{$\mathbf{55.3}${\scriptsize $\mathbf{/1.7}$}} & \cellcolor{lightgreen}\strut \underline{$\mathbf{45.3}${\scriptsize $\mathbf{/1.7}$}} & \cellcolor{lightgreen}\strut \underline{$\mathbf{31.7}${\scriptsize $\mathbf{/1.6}$}} & \cellcolor{lightgreen}\strut $\mathbf{34.3}${\scriptsize $\mathbf{/1.7}$} & \cellcolor{lightgreen}\strut \underline{$\mathbf{25.1}${\scriptsize $\mathbf{/1.7}$}} & \cellcolor{lightgreen}\strut \underline{$\mathbf{27.2}${\scriptsize $\mathbf{/1.7}$}} & \cellcolor{lightgreen}\strut $\mathbf{71.5}${\scriptsize $\mathbf{/1.7}$} & \underline{$\mathbf{91.7}${\scriptsize $\mathbf{/1.7}$}} & \cellcolor{lightgreen}\strut \underline{$\mathbf{44.6}${\scriptsize $\mathbf{/1.7}$}} & \underline{$\mathbf{6.4}${\scriptsize $\mathbf{/1.7}$}} & \cellcolor{lightgreen}\strut \underline{$\mathbf{100.0}${\scriptsize $\mathbf{/1.7}$}} & \cellcolor{lightgreen}\strut $\mathbf{62.5}${\scriptsize $\mathbf{/1.7}$} & \cellcolor{lightgreen}\strut \underline{$\mathbf{57.0}${\scriptsize $\mathbf{/1.7}$}} \\
 & {\bfseries\boldmath Our $\tau.1$ Q32} & \underline{$\mathbf{29.7}${\scriptsize $\mathbf{/1.6}$}} & \cellcolor{lightgreen}\strut \underline{$\mathbf{45.3}${\scriptsize $\mathbf{/1.6}$}} & \underline{$\mathbf{55.7}${\scriptsize $\mathbf{/1.6}$}} & \underline{$\mathbf{55.1}${\scriptsize $\mathbf{/1.6}$}} & \underline{$\mathbf{44.9}${\scriptsize $\mathbf{/1.6}$}} & \underline{$\mathbf{31.2}${\scriptsize $\mathbf{/1.6}$}} & \cellcolor{lightgreen}\strut \underline{$\mathbf{34.6}${\scriptsize $\mathbf{/1.6}$}} & \cellcolor{lightgreen}\strut \underline{$\mathbf{25.8}${\scriptsize $\mathbf{/1.6}$}} & \cellcolor{lightgreen}\strut \underline{$\mathbf{27.4}${\scriptsize $\mathbf{/1.6}$}} & \cellcolor{lightgreen}\strut \underline{$\mathbf{72.5}${\scriptsize $\mathbf{/1.6}$}} & \underline{$\mathbf{91.3}${\scriptsize $\mathbf{/1.6}$}} & \underline{$\mathbf{43.7}${\scriptsize $\mathbf{/1.6}$}} & \cellcolor{lightgreen}\strut \underline{$\mathbf{7.6}${\scriptsize $\mathbf{/1.6}$}} & \cellcolor{lightgreen}\strut \underline{$\mathbf{100.0}${\scriptsize $\mathbf{/1.6}$}} & \cellcolor{lightgreen}\strut \underline{$\mathbf{62.8}${\scriptsize $\mathbf{/1.6}$}} & \underline{$\mathbf{56.5}${\scriptsize $\mathbf{/1.6}$}} \\
\midrule
\multirow{12}{*}{\rotatebox[origin=c]{90}{\textbf{Qwen}}} & Full & $28.6${\scriptsize $/1.0$} & $44.6${\scriptsize $/1.0$} & $51.4${\scriptsize $/1.0$} & $58.3${\scriptsize $/1.0$} & $46.5${\scriptsize $/1.0$} & $29.6${\scriptsize $/1.0$} & $33.3${\scriptsize $/1.0$} & $24.3${\scriptsize $/1.0$} & $25.8${\scriptsize $/1.0$} & $73.5${\scriptsize $/1.0$} & $89.4${\scriptsize $/1.0$} & $46.9${\scriptsize $/1.0$} & $9.0${\scriptsize $/1.0$} & $100.0${\scriptsize $/1.0$} & $61.2${\scriptsize $/1.0$} & $66.9${\scriptsize $/1.0$} \\
 & Snap $10\%$ & \cellcolor{lightgreen}\strut \underline{$29.2${\scriptsize $/1.3$}} & $36.1${\scriptsize $/1.3$} & $48.8${\scriptsize $/1.3$} & \cellcolor{lightgreen}\strut $57.4${\scriptsize $/1.3$} & \cellcolor{lightgreen}\strut \underline{$46.4${\scriptsize $/1.3$}} & \cellcolor{lightgreen}\strut \underline{$29.8${\scriptsize $/1.3$}} & $24.8${\scriptsize $/1.3$} & $22.7${\scriptsize $/1.3$} & $16.4${\scriptsize $/1.3$} & $54.5${\scriptsize $/1.3$} & \cellcolor{lightgreen}\strut $88.4${\scriptsize $/1.3$} & $43.8${\scriptsize $/1.3$} & \underline{$9.0${\scriptsize $/1.3$}} & \cellcolor{lightgreen}\strut \underline{$100.0${\scriptsize $/1.3$}} & $53.4${\scriptsize $/1.3$} & $65.5${\scriptsize $/1.3$} \\
 & Snap $25\%$ & \cellcolor{lightgreen}\strut \underline{$29.9${\scriptsize $/1.3$}} & $41.1${\scriptsize $/1.3$} & $50.7${\scriptsize $/1.3$} & \cellcolor{lightgreen}\strut \underline{$58.3${\scriptsize $/1.3$}} & \underline{$46.1${\scriptsize $/1.3$}} & \cellcolor{lightgreen}\strut \underline{$30.6${\scriptsize $/1.3$}} & $28.4${\scriptsize $/1.3$} & $23.3${\scriptsize $/1.3$} & $20.1${\scriptsize $/1.3$} & $65.0${\scriptsize $/1.3$} & \cellcolor{lightgreen}\strut $88.8${\scriptsize $/1.3$} & $44.9${\scriptsize $/1.3$} & \underline{$9.0${\scriptsize $/1.3$}} & \underline{$100.0${\scriptsize $/1.3$}} & $57.2${\scriptsize $/1.3$} & \cellcolor{lightgreen}\strut \underline{$66.7${\scriptsize $/1.3$}} \\
 & Str. $10\%$ & $23.4${\scriptsize $/1.3$} & $27.1${\scriptsize $/1.3$} & $28.2${\scriptsize $/1.3$} & $43.9${\scriptsize $/1.3$} & $38.0${\scriptsize $/1.3$} & $19.7${\scriptsize $/1.3$} & $23.8${\scriptsize $/1.3$} & $19.3${\scriptsize $/1.3$} & $16.0${\scriptsize $/1.3$} & $58.5${\scriptsize $/1.3$} & $86.6${\scriptsize $/1.3$} & $43.3${\scriptsize $/1.3$} & \cellcolor{lightgreen}\strut \underline{$9.5${\scriptsize $/1.3$}} & $37.0${\scriptsize $/1.3$} & $49.9${\scriptsize $/1.3$} & $57.7${\scriptsize $/1.3$} \\
 & Str. $25\%$ & $25.9${\scriptsize $/1.3$} & $29.9${\scriptsize $/1.3$} & $31.3${\scriptsize $/1.3$} & $47.5${\scriptsize $/1.3$} & $40.7${\scriptsize $/1.3$} & $22.8${\scriptsize $/1.3$} & $27.7${\scriptsize $/1.3$} & $20.1${\scriptsize $/1.3$} & $19.5${\scriptsize $/1.3$} & $64.5${\scriptsize $/1.3$} & \cellcolor{lightgreen}\strut \underline{$89.1${\scriptsize $/1.3$}} & $44.0${\scriptsize $/1.3$} & \underline{$9.0${\scriptsize $/1.3$}} & $57.2${\scriptsize $/1.3$} & $53.4${\scriptsize $/1.3$} & $62.3${\scriptsize $/1.3$} \\
 & Pyr. $10\%$ & \cellcolor{lightgreen}\strut \underline{$29.0${\scriptsize $/1.3$}} & $37.7${\scriptsize $/1.3$} & $48.6${\scriptsize $/1.3$} & $55.5${\scriptsize $/1.3$} & $44.7${\scriptsize $/1.3$} & $27.4${\scriptsize $/1.3$} & $25.7${\scriptsize $/1.3$} & $22.8${\scriptsize $/1.3$} & $18.0${\scriptsize $/1.3$} & $55.0${\scriptsize $/1.3$} & $85.4${\scriptsize $/1.3$} & $45.2${\scriptsize $/1.3$} & \cellcolor{lightgreen}\strut \underline{$9.0${\scriptsize $/1.3$}} & \cellcolor{lightgreen}\strut \underline{$100.0${\scriptsize $/1.3$}} & $56.5${\scriptsize $/1.3$} & $63.5${\scriptsize $/1.3$} \\
 & Pyr. $25\%$ & \underline{$29.9${\scriptsize $/1.3$}} & $41.5${\scriptsize $/1.3$} & $49.3${\scriptsize $/1.3$} & $56.9${\scriptsize $/1.3$} & $45.6${\scriptsize $/1.3$} & \underline{$29.5${\scriptsize $/1.3$}} & $28.1${\scriptsize $/1.3$} & $23.2${\scriptsize $/1.3$} & $21.5${\scriptsize $/1.3$} & $67.0${\scriptsize $/1.3$} & $87.0${\scriptsize $/1.3$} & \underline{$46.6${\scriptsize $/1.3$}} & \underline{$9.0${\scriptsize $/1.3$}} & \underline{$100.0${\scriptsize $/1.3$}} & $60.3${\scriptsize $/1.3$} & $63.9${\scriptsize $/1.3$} \\
 & Quest & \underline{$29.0${\scriptsize $/0.9$}} & $43.7${\scriptsize $/0.9$} & \underline{$52.3${\scriptsize $/0.9$}} & $57.4${\scriptsize $/0.9$} & \underline{$46.3${\scriptsize $/0.9$}} & $28.9${\scriptsize $/0.9$} & \underline{$33.1${\scriptsize $/0.9$}} & $23.4${\scriptsize $/0.9$} & $24.3${\scriptsize $/0.9$} & $72.0${\scriptsize $/0.9$} & \underline{$89.6${\scriptsize $/0.9$}} & $45.9${\scriptsize $/0.9$} & $7.5${\scriptsize $/0.9$} & $99.5${\scriptsize $/0.9$} & $59.5${\scriptsize $/0.9$} & $65.5${\scriptsize $/0.9$} \\
 & Double & \underline{$28.5${\scriptsize $/1.1$}} & $43.8${\scriptsize $/1.1$} & $50.7${\scriptsize $/1.1$} & \underline{$58.3${\scriptsize $/1.1$}} & $44.7${\scriptsize $/1.1$} & $28.2${\scriptsize $/1.1$} & \underline{$33.1${\scriptsize $/1.1$}} & $23.3${\scriptsize $/1.1$} & \underline{$25.5${\scriptsize $/1.1$}} & \underline{$73.5${\scriptsize $/1.1$}} & $88.8${\scriptsize $/1.1$} & \underline{$46.8${\scriptsize $/1.1$}} & \underline{$9.0${\scriptsize $/1.1$}} & $98.0${\scriptsize $/1.1$} & $60.6${\scriptsize $/1.1$} & $64.2${\scriptsize $/1.1$} \\
 & {\bfseries\boldmath Our $\tau.01$} & \underline{$\mathbf{28.6}${\scriptsize $\mathbf{/1.2}$}} & \underline{$\mathbf{44.8}${\scriptsize $\mathbf{/1.2}$}} & \underline{$\mathbf{51.1}${\scriptsize $\mathbf{/1.2}$}} & \underline{$\mathbf{58.3}${\scriptsize $\mathbf{/1.2}$}} & $\mathbf{45.6}${\scriptsize $\mathbf{/1.2}$} & \underline{$\mathbf{29.5}${\scriptsize $\mathbf{/1.2}$}} & \underline{$\mathbf{33.4}${\scriptsize $\mathbf{/1.2}$}} & \underline{$\mathbf{24.2}${\scriptsize $\mathbf{/1.2}$}} & \cellcolor{lightgreen}\strut \underline{$\mathbf{25.8}${\scriptsize $\mathbf{/1.2}$}} & \cellcolor{lightgreen}\strut \underline{$\mathbf{73.5}${\scriptsize $\mathbf{/1.2}$}} & \underline{$\mathbf{88.9}${\scriptsize $\mathbf{/1.2}$}} & \underline{$\mathbf{47.0}${\scriptsize $\mathbf{/1.2}$}} & \underline{$\mathbf{9.0}${\scriptsize $\mathbf{/1.2}$}} & \underline{$\mathbf{100.0}${\scriptsize $\mathbf{/1.2}$}} & \underline{$\mathbf{61.3}${\scriptsize $\mathbf{/1.2}$}} & \underline{$\mathbf{66.7}${\scriptsize $\mathbf{/1.2}$}} \\
 & {\bfseries\boldmath Our $\tau.1$} & \underline{$\mathbf{29.1}${\scriptsize $\mathbf{/1.3}$}} & \cellcolor{lightgreen}\strut \underline{$\mathbf{45.0}${\scriptsize $\mathbf{/1.3}$}} & \cellcolor{lightgreen}\strut $\mathbf{50.8}${\scriptsize $\mathbf{/1.3}$} & $\mathbf{57.7}${\scriptsize $\mathbf{/1.3}$} & $\mathbf{45.6}${\scriptsize $\mathbf{/1.3}$} & \underline{$\mathbf{29.7}${\scriptsize $\mathbf{/1.3}$}} & \cellcolor{lightgreen}\strut \underline{$\mathbf{33.8}${\scriptsize $\mathbf{/1.3}$}} & \cellcolor{lightgreen}\strut \underline{$\mathbf{24.4}${\scriptsize $\mathbf{/1.3}$}} & \cellcolor{lightgreen}\strut \underline{$\mathbf{25.6}${\scriptsize $\mathbf{/1.3}$}} & \cellcolor{lightgreen}\strut $\mathbf{73.0}${\scriptsize $\mathbf{/1.3}$} & \underline{$\mathbf{89.0}${\scriptsize $\mathbf{/1.3}$}} & \cellcolor{lightgreen}\strut \underline{$\mathbf{47.1}${\scriptsize $\mathbf{/1.3}$}} & \underline{$\mathbf{9.0}${\scriptsize $\mathbf{/1.3}$}} & $\mathbf{99.5}${\scriptsize $\mathbf{/1.3}$} & \cellcolor{lightgreen}\strut \underline{$\mathbf{61.0}${\scriptsize $\mathbf{/1.3}$}} & \cellcolor{lightgreen}\strut $\mathbf{66.1}${\scriptsize $\mathbf{/1.3}$} \\
 & {\bfseries\boldmath Our $\tau.1$ Q32} & \underline{$\mathbf{28.7}${\scriptsize $\mathbf{/1.2}$}} & \underline{$\mathbf{45.0}${\scriptsize $\mathbf{/1.2}$}} & \cellcolor{lightgreen}\strut \underline{$\mathbf{51.1}${\scriptsize $\mathbf{/1.2}$}} & \cellcolor{lightgreen}\strut \underline{$\mathbf{58.4}${\scriptsize $\mathbf{/1.2}$}} & \underline{$\mathbf{46.1}${\scriptsize $\mathbf{/1.2}$}} & \underline{$\mathbf{29.7}${\scriptsize $\mathbf{/1.2}$}} & \cellcolor{lightgreen}\strut \underline{$\mathbf{33.8}${\scriptsize $\mathbf{/1.2}$}} & \cellcolor{lightgreen}\strut \underline{$\mathbf{24.5}${\scriptsize $\mathbf{/1.2}$}} & \underline{$\mathbf{25.5}${\scriptsize $\mathbf{/1.2}$}} & \cellcolor{lightgreen}\strut \underline{$\mathbf{73.5}${\scriptsize $\mathbf{/1.2}$}} & \underline{$\mathbf{88.9}${\scriptsize $\mathbf{/1.2}$}} & \underline{$\mathbf{46.9}${\scriptsize $\mathbf{/1.2}$}} & \underline{$\mathbf{9.0}${\scriptsize $\mathbf{/1.2}$}} & \underline{$\mathbf{100.0}${\scriptsize $\mathbf{/1.2}$}} & \cellcolor{lightgreen}\strut \underline{$\mathbf{61.5}${\scriptsize $\mathbf{/1.2}$}} & \cellcolor{lightgreen}\strut \underline{$\mathbf{66.7}${\scriptsize $\mathbf{/1.2}$}} \\
\bottomrule
\end{tabular}%
}
\end{table*}

\providecolor{lightgreen}{RGB}{239,248,239}
\begin{table*}[t]
\centering
\caption{RULER per-task quality--latency comparison.}
\label{tab:ruler-per-task-acc-accl}
\setlength{\tabcolsep}{1.0pt}
\renewcommand{\arraystretch}{1.10}
\small
\resizebox{\textwidth}{!}{%
\begin{tabular}{@{}c@{\hspace{0.5pt}}l@{\hspace{0.5pt}}c@{\hspace{0.5pt}}*{12}{c}@{}}
\toprule
 & Method & \multicolumn{3}{c}{S-NIAH} & \multicolumn{3}{c}{MK-NIAH} & \multicolumn{2}{c}{MV/MQ-NIAH} & \multicolumn{3}{c}{A\&T} & \multicolumn{2}{c}{QA} \\[-2pt]
\cmidrule(lr){3-5}\cmidrule(lr){6-8}\cmidrule(lr){9-10}\cmidrule(lr){11-13}\cmidrule(l){14-15}
 & & \rotatebox{60}{\scriptsize NIAH-S1} & \rotatebox{60}{\scriptsize NIAH-S2} & \rotatebox{60}{\scriptsize NIAH-S3} & \rotatebox{60}{\scriptsize NIAH-MK1} & \rotatebox{60}{\scriptsize NIAH-MK2} & \rotatebox{60}{\scriptsize NIAH-MK3} & \rotatebox{60}{\scriptsize NIAH-MV} & \rotatebox{60}{\scriptsize NIAH-MQ} & \rotatebox{60}{\scriptsize CWE} & \rotatebox{60}{\scriptsize FWE} & \rotatebox{60}{\scriptsize VT} & \rotatebox{60}{\scriptsize QA-1} & \rotatebox{60}{\scriptsize QA-2} \\
\midrule
\multirow{12}{*}{\rotatebox[origin=c]{90}{\textbf{Llama}}} & Full & $100.0${\scriptsize $/1.0$} & $100.0${\scriptsize $/1.0$} & $100.0${\scriptsize $/1.0$} & $90.0${\scriptsize $/1.0$} & $70.0${\scriptsize $/1.0$} & $55.0${\scriptsize $/1.0$} & $93.8${\scriptsize $/1.0$} & $100.0${\scriptsize $/1.0$} & $0.0${\scriptsize $/1.0$} & $70.0${\scriptsize $/1.0$} & $65.0${\scriptsize $/1.0$} & $65.0${\scriptsize $/1.0$} & $45.0${\scriptsize $/1.0$} \\
 & Snap $10\%$ & \cellcolor{lightgreen}\strut \underline{$100.0${\scriptsize $/1.7$}} & \cellcolor{lightgreen}\strut \underline{$100.0${\scriptsize $/1.7$}} & $70.0${\scriptsize $/1.7$} & \cellcolor{lightgreen}\strut \underline{$90.0${\scriptsize $/1.7$}} & $60.0${\scriptsize $/1.7$} & $30.0${\scriptsize $/1.7$} & \cellcolor{lightgreen}\strut $88.8${\scriptsize $/1.7$} & \cellcolor{lightgreen}\strut $97.5${\scriptsize $/1.7$} & \cellcolor{lightgreen}\strut \underline{$0.5${\scriptsize $/1.7$}} & $61.7${\scriptsize $/1.7$} & $60.0${\scriptsize $/1.7$} & \cellcolor{lightgreen}\strut \underline{$65.0${\scriptsize $/1.7$}} & \cellcolor{lightgreen}\strut \underline{$45.0${\scriptsize $/1.7$}} \\
 & Snap $25\%$ & \underline{$100.0${\scriptsize $/1.5$}} & \underline{$100.0${\scriptsize $/1.5$}} & $85.0${\scriptsize $/1.5$} & \underline{$90.0${\scriptsize $/1.5$}} & \cellcolor{lightgreen}\strut $65.0${\scriptsize $/1.5$} & $40.0${\scriptsize $/1.5$} & \cellcolor{lightgreen}\strut \underline{$93.8${\scriptsize $/1.5$}} & \cellcolor{lightgreen}\strut \underline{$100.0${\scriptsize $/1.5$}} & \underline{$0.0${\scriptsize $/1.5$}} & $65.0${\scriptsize $/1.5$} & $59.0${\scriptsize $/1.5$} & \underline{$65.0${\scriptsize $/1.5$}} & \underline{$45.0${\scriptsize $/1.5$}} \\
 & Str. $10\%$ & $5.0${\scriptsize $/1.7$} & $20.0${\scriptsize $/1.7$} & $5.0${\scriptsize $/1.7$} & $10.0${\scriptsize $/1.7$} & $10.0${\scriptsize $/1.7$} & $20.0${\scriptsize $/1.7$} & $10.0${\scriptsize $/1.7$} & $10.0${\scriptsize $/1.7$} & \cellcolor{lightgreen}\strut \underline{$0.5${\scriptsize $/1.7$}} & \cellcolor{lightgreen}\strut \underline{$80.0${\scriptsize $/1.7$}} & $9.0${\scriptsize $/1.7$} & $50.0${\scriptsize $/1.7$} & $25.0${\scriptsize $/1.7$} \\
 & Str. $25\%$ & $35.0${\scriptsize $/1.5$} & $45.0${\scriptsize $/1.5$} & $20.0${\scriptsize $/1.5$} & $40.0${\scriptsize $/1.5$} & $30.0${\scriptsize $/1.5$} & $25.0${\scriptsize $/1.5$} & $25.0${\scriptsize $/1.5$} & $25.0${\scriptsize $/1.5$} & \cellcolor{lightgreen}\strut \underline{$1.0${\scriptsize $/1.5$}} & \underline{$76.7${\scriptsize $/1.5$}} & $25.0${\scriptsize $/1.5$} & \underline{$65.0${\scriptsize $/1.5$}} & $35.0${\scriptsize $/1.5$} \\
 & Pyr. $10\%$ & \cellcolor{lightgreen}\strut \underline{$100.0${\scriptsize $/1.7$}} & \cellcolor{lightgreen}\strut \underline{$100.0${\scriptsize $/1.7$}} & $70.0${\scriptsize $/1.7$} & \cellcolor{lightgreen}\strut \underline{$90.0${\scriptsize $/1.7$}} & $60.0${\scriptsize $/1.7$} & $35.0${\scriptsize $/1.7$} & $83.8${\scriptsize $/1.7$} & \cellcolor{lightgreen}\strut $98.8${\scriptsize $/1.7$} & \cellcolor{lightgreen}\strut \underline{$0.5${\scriptsize $/1.7$}} & $55.0${\scriptsize $/1.7$} & \cellcolor{lightgreen}\strut $62.0${\scriptsize $/1.7$} & \cellcolor{lightgreen}\strut \underline{$65.0${\scriptsize $/1.7$}} & \cellcolor{lightgreen}\strut \underline{$45.0${\scriptsize $/1.7$}} \\
 & Pyr. $25\%$ & \underline{$100.0${\scriptsize $/1.5$}} & \underline{$100.0${\scriptsize $/1.5$}} & $85.0${\scriptsize $/1.5$} & \underline{$90.0${\scriptsize $/1.5$}} & $60.0${\scriptsize $/1.5$} & \cellcolor{lightgreen}\strut $50.0${\scriptsize $/1.5$} & \cellcolor{lightgreen}\strut \underline{$93.8${\scriptsize $/1.5$}} & \cellcolor{lightgreen}\strut \underline{$100.0${\scriptsize $/1.5$}} & \underline{$0.0${\scriptsize $/1.5$}} & $63.3${\scriptsize $/1.5$} & $62.0${\scriptsize $/1.5$} & \underline{$65.0${\scriptsize $/1.5$}} & \underline{$45.0${\scriptsize $/1.5$}} \\
 & Quest & \underline{$100.0${\scriptsize $/1.1$}} & \underline{$100.0${\scriptsize $/1.1$}} & $75.0${\scriptsize $/1.1$} & \underline{$90.0${\scriptsize $/1.1$}} & $45.0${\scriptsize $/1.1$} & $0.0${\scriptsize $/1.1$} & $77.5${\scriptsize $/1.1$} & $87.5${\scriptsize $/1.1$} & \underline{$0.0${\scriptsize $/1.1$}} & $66.7${\scriptsize $/1.1$} & $64.0${\scriptsize $/1.1$} & \underline{$65.0${\scriptsize $/1.1$}} & $40.0${\scriptsize $/1.1$} \\
 & Double & $95.0${\scriptsize $/1.4$} & $80.0${\scriptsize $/1.4$} & $95.0${\scriptsize $/1.4$} & $85.0${\scriptsize $/1.4$} & $20.0${\scriptsize $/1.4$} & $5.0${\scriptsize $/1.4$} & $42.5${\scriptsize $/1.4$} & $53.8${\scriptsize $/1.4$} & \underline{$0.5${\scriptsize $/1.4$}} & $68.3${\scriptsize $/1.4$} & $34.0${\scriptsize $/1.4$} & \underline{$65.0${\scriptsize $/1.4$}} & \underline{$45.0${\scriptsize $/1.4$}} \\
 & {\bfseries\boldmath Our $\tau.01$} & \underline{$\mathbf{100.0}${\scriptsize $\mathbf{/1.6}$}} & \underline{$\mathbf{100.0}${\scriptsize $\mathbf{/1.5}$}} & \underline{$\mathbf{100.0}${\scriptsize $\mathbf{/1.5}$}} & \cellcolor{lightgreen}\strut \underline{$\mathbf{95.0}${\scriptsize $\mathbf{/1.6}$}} & $\mathbf{55.0}${\scriptsize $\mathbf{/1.5}$} & $\mathbf{25.0}${\scriptsize $\mathbf{/1.6}$} & \cellcolor{lightgreen}\strut $\mathbf{90.0}${\scriptsize $\mathbf{/1.5}$} & $\mathbf{97.5}${\scriptsize $\mathbf{/1.5}$} & \underline{$\mathbf{0.0}${\scriptsize $\mathbf{/1.6}$}} & \underline{$\mathbf{70.0}${\scriptsize $\mathbf{/1.6}$}} & \cellcolor{lightgreen}\strut \underline{$\mathbf{70.0}${\scriptsize $\mathbf{/1.6}$}} & \underline{$\mathbf{65.0}${\scriptsize $\mathbf{/1.5}$}} & \underline{$\mathbf{45.0}${\scriptsize $\mathbf{/1.5}$}} \\
 & {\bfseries\boldmath Our $\tau.1$} & \cellcolor{lightgreen}\strut \underline{$\mathbf{100.0}${\scriptsize $\mathbf{/1.7}$}} & $\mathbf{95.0}${\scriptsize $\mathbf{/1.7}$} & \cellcolor{lightgreen}\strut \underline{$\mathbf{100.0}${\scriptsize $\mathbf{/1.7}$}} & \cellcolor{lightgreen}\strut \underline{$\mathbf{90.0}${\scriptsize $\mathbf{/1.7}$}} & $\mathbf{30.0}${\scriptsize $\mathbf{/1.7}$} & $\mathbf{5.0}${\scriptsize $\mathbf{/1.7}$} & $\mathbf{71.2}${\scriptsize $\mathbf{/1.7}$} & $\mathbf{86.2}${\scriptsize $\mathbf{/1.7}$} & \underline{$\mathbf{0.0}${\scriptsize $\mathbf{/1.7}$}} & $\mathbf{60.0}${\scriptsize $\mathbf{/1.7}$} & $\mathbf{61.0}${\scriptsize $\mathbf{/1.7}$} & \cellcolor{lightgreen}\strut \underline{$\mathbf{70.0}${\scriptsize $\mathbf{/1.7}$}} & \cellcolor{lightgreen}\strut \underline{$\mathbf{45.0}${\scriptsize $\mathbf{/1.7}$}} \\
 & {\bfseries\boldmath Our $\tau.1$ Q32} & \cellcolor{lightgreen}\strut \underline{$\mathbf{100.0}${\scriptsize $\mathbf{/1.7}$}} & \cellcolor{lightgreen}\strut \underline{$\mathbf{100.0}${\scriptsize $\mathbf{/1.6}$}} & \cellcolor{lightgreen}\strut \underline{$\mathbf{100.0}${\scriptsize $\mathbf{/1.6}$}} & \cellcolor{lightgreen}\strut \underline{$\mathbf{90.0}${\scriptsize $\mathbf{/1.7}$}} & $\mathbf{60.0}${\scriptsize $\mathbf{/1.7}$} & $\mathbf{20.0}${\scriptsize $\mathbf{/1.7}$} & $\mathbf{83.8}${\scriptsize $\mathbf{/1.6}$} & $\mathbf{93.8}${\scriptsize $\mathbf{/1.6}$} & \underline{$\mathbf{0.0}${\scriptsize $\mathbf{/1.7}$}} & $\mathbf{63.3}${\scriptsize $\mathbf{/1.7}$} & \cellcolor{lightgreen}\strut \underline{$\mathbf{65.0}${\scriptsize $\mathbf{/1.7}$}} & \underline{$\mathbf{65.0}${\scriptsize $\mathbf{/1.6}$}} & \cellcolor{lightgreen}\strut \underline{$\mathbf{45.0}${\scriptsize $\mathbf{/1.6}$}} \\
\midrule
\multirow{12}{*}{\rotatebox[origin=c]{90}{\textbf{Qwen}}} & Full & $100.0${\scriptsize $/1.0$} & $100.0${\scriptsize $/1.0$} & $90.0${\scriptsize $/1.0$} & $95.0${\scriptsize $/1.0$} & $45.0${\scriptsize $/1.0$} & $10.0${\scriptsize $/1.0$} & $73.8${\scriptsize $/1.0$} & $85.0${\scriptsize $/1.0$} & $38.0${\scriptsize $/1.0$} & $51.7${\scriptsize $/1.0$} & $83.0${\scriptsize $/1.0$} & $60.0${\scriptsize $/1.0$} & $25.0${\scriptsize $/1.0$} \\
 & Snap $10\%$ & \cellcolor{lightgreen}\strut \underline{$100.0${\scriptsize $/1.3$}} & \cellcolor{lightgreen}\strut $95.0${\scriptsize $/1.3$} & $0.0${\scriptsize $/1.3$} & $75.0${\scriptsize $/1.3$} & $0.0${\scriptsize $/1.3$} & $0.0${\scriptsize $/1.3$} & $20.0${\scriptsize $/1.3$} & $72.5${\scriptsize $/1.3$} & \cellcolor{lightgreen}\strut $35.5${\scriptsize $/1.3$} & $43.3${\scriptsize $/1.3$} & $82.0${\scriptsize $/1.3$} & \cellcolor{lightgreen}\strut $55.0${\scriptsize $/1.3$} & $20.0${\scriptsize $/1.3$} \\
 & Snap $25\%$ & \cellcolor{lightgreen}\strut \underline{$100.0${\scriptsize $/1.3$}} & \cellcolor{lightgreen}\strut \underline{$100.0${\scriptsize $/1.3$}} & $5.0${\scriptsize $/1.3$} & $70.0${\scriptsize $/1.3$} & $5.0${\scriptsize $/1.3$} & $0.0${\scriptsize $/1.3$} & $43.8${\scriptsize $/1.3$} & \cellcolor{lightgreen}\strut \underline{$86.2${\scriptsize $/1.3$}} & \cellcolor{lightgreen}\strut \underline{$44.0${\scriptsize $/1.3$}} & $46.7${\scriptsize $/1.3$} & \underline{$83.0${\scriptsize $/1.3$}} & \cellcolor{lightgreen}\strut $55.0${\scriptsize $/1.3$} & \cellcolor{lightgreen}\strut \underline{$25.0${\scriptsize $/1.3$}} \\
 & Str. $10\%$ & $5.0${\scriptsize $/1.3$} & $20.0${\scriptsize $/1.3$} & $5.0${\scriptsize $/1.3$} & $5.0${\scriptsize $/1.3$} & $0.0${\scriptsize $/1.3$} & $0.0${\scriptsize $/1.3$} & $10.0${\scriptsize $/1.3$} & $10.0${\scriptsize $/1.3$} & $18.0${\scriptsize $/1.3$} & \cellcolor{lightgreen}\strut $48.3${\scriptsize $/1.3$} & $15.0${\scriptsize $/1.3$} & $25.0${\scriptsize $/1.3$} & $20.0${\scriptsize $/1.3$} \\
 & Str. $25\%$ & $35.0${\scriptsize $/1.3$} & $50.0${\scriptsize $/1.3$} & $15.0${\scriptsize $/1.3$} & $15.0${\scriptsize $/1.3$} & $10.0${\scriptsize $/1.3$} & \cellcolor{lightgreen}\strut \underline{$10.0${\scriptsize $/1.3$}} & $17.5${\scriptsize $/1.3$} & $23.8${\scriptsize $/1.3$} & $18.5${\scriptsize $/1.3$} & $43.3${\scriptsize $/1.3$} & $29.0${\scriptsize $/1.3$} & \cellcolor{lightgreen}\strut \underline{$65.0${\scriptsize $/1.3$}} & $15.0${\scriptsize $/1.3$} \\
 & Pyr. $10\%$ & \cellcolor{lightgreen}\strut $95.0${\scriptsize $/1.3$} & $80.0${\scriptsize $/1.3$} & $0.0${\scriptsize $/1.3$} & $50.0${\scriptsize $/1.3$} & $5.0${\scriptsize $/1.3$} & $0.0${\scriptsize $/1.3$} & $18.8${\scriptsize $/1.3$} & $56.2${\scriptsize $/1.3$} & $21.0${\scriptsize $/1.3$} & $45.0${\scriptsize $/1.3$} & \cellcolor{lightgreen}\strut \underline{$90.0${\scriptsize $/1.3$}} & $45.0${\scriptsize $/1.3$} & \cellcolor{lightgreen}\strut \underline{$25.0${\scriptsize $/1.3$}} \\
 & Pyr. $25\%$ & \underline{$100.0${\scriptsize $/1.3$}} & \cellcolor{lightgreen}\strut \underline{$100.0${\scriptsize $/1.3$}} & $0.0${\scriptsize $/1.3$} & $65.0${\scriptsize $/1.3$} & $5.0${\scriptsize $/1.3$} & $0.0${\scriptsize $/1.3$} & $36.2${\scriptsize $/1.3$} & $83.8${\scriptsize $/1.3$} & $31.0${\scriptsize $/1.3$} & $50.0${\scriptsize $/1.3$} & \underline{$83.0${\scriptsize $/1.3$}} & $55.0${\scriptsize $/1.3$} & \underline{$25.0${\scriptsize $/1.3$}} \\
 & Quest & \underline{$100.0${\scriptsize $/0.9$}} & $90.0${\scriptsize $/0.9$} & $50.0${\scriptsize $/0.9$} & $70.0${\scriptsize $/0.9$} & $25.0${\scriptsize $/0.9$} & $0.0${\scriptsize $/0.9$} & $68.8${\scriptsize $/0.9$} & $71.2${\scriptsize $/0.9$} & $21.5${\scriptsize $/0.9$} & $43.3${\scriptsize $/0.9$} & \underline{$86.0${\scriptsize $/0.9$}} & $30.0${\scriptsize $/0.9$} & \underline{$25.0${\scriptsize $/0.9$}} \\
 & Double & $95.0${\scriptsize $/1.1$} & $75.0${\scriptsize $/1.1$} & $75.0${\scriptsize $/1.1$} & $60.0${\scriptsize $/1.1$} & $15.0${\scriptsize $/1.1$} & $0.0${\scriptsize $/1.1$} & $47.5${\scriptsize $/1.1$} & $53.8${\scriptsize $/1.1$} & \underline{$42.0${\scriptsize $/1.1$}} & $50.0${\scriptsize $/1.1$} & $70.0${\scriptsize $/1.1$} & \underline{$60.0${\scriptsize $/1.1$}} & $20.0${\scriptsize $/1.1$} \\
 & {\bfseries\boldmath Our $\tau.01$} & \underline{$\mathbf{100.0}${\scriptsize $\mathbf{/1.2}$}} & \underline{$\mathbf{100.0}${\scriptsize $\mathbf{/1.2}$}} & \underline{$\mathbf{90.0}${\scriptsize $\mathbf{/1.2}$}} & \cellcolor{lightgreen}\strut \underline{$\mathbf{95.0}${\scriptsize $\mathbf{/1.2}$}} & \cellcolor{lightgreen}\strut $\mathbf{40.0}${\scriptsize $\mathbf{/1.2}$} & \underline{$\mathbf{10.0}${\scriptsize $\mathbf{/1.2}$}} & \cellcolor{lightgreen}\strut $\mathbf{70.0}${\scriptsize $\mathbf{/1.2}$} & $\mathbf{81.2}${\scriptsize $\mathbf{/1.2}$} & \underline{$\mathbf{39.0}${\scriptsize $\mathbf{/1.2}$}} & \underline{$\mathbf{53.3}${\scriptsize $\mathbf{/1.2}$}} & \underline{$\mathbf{85.0}${\scriptsize $\mathbf{/1.2}$}} & \underline{$\mathbf{60.0}${\scriptsize $\mathbf{/1.2}$}} & \underline{$\mathbf{25.0}${\scriptsize $\mathbf{/1.2}$}} \\
 & {\bfseries\boldmath Our $\tau.1$} & \cellcolor{lightgreen}\strut \underline{$\mathbf{100.0}${\scriptsize $\mathbf{/1.3}$}} & \cellcolor{lightgreen}\strut $\mathbf{95.0}${\scriptsize $\mathbf{/1.3}$} & \cellcolor{lightgreen}\strut \underline{$\mathbf{90.0}${\scriptsize $\mathbf{/1.3}$}} & $\mathbf{85.0}${\scriptsize $\mathbf{/1.3}$} & $\mathbf{30.0}${\scriptsize $\mathbf{/1.3}$} & $\mathbf{0.0}${\scriptsize $\mathbf{/1.3}$} & \cellcolor{lightgreen}\strut $\mathbf{68.8}${\scriptsize $\mathbf{/1.3}$} & $\mathbf{68.8}${\scriptsize $\mathbf{/1.3}$} & \cellcolor{lightgreen}\strut \underline{$\mathbf{39.0}${\scriptsize $\mathbf{/1.3}$}} & \cellcolor{lightgreen}\strut \underline{$\mathbf{55.0}${\scriptsize $\mathbf{/1.3}$}} & $\mathbf{78.0}${\scriptsize $\mathbf{/1.3}$} & $\mathbf{55.0}${\scriptsize $\mathbf{/1.3}$} & $\mathbf{20.0}${\scriptsize $\mathbf{/1.2}$} \\
 & {\bfseries\boldmath Our $\tau.1$ Q32} & \cellcolor{lightgreen}\strut \underline{$\mathbf{100.0}${\scriptsize $\mathbf{/1.3}$}} & \cellcolor{lightgreen}\strut \underline{$\mathbf{100.0}${\scriptsize $\mathbf{/1.2}$}} & \underline{$\mathbf{90.0}${\scriptsize $\mathbf{/1.2}$}} & \cellcolor{lightgreen}\strut $\mathbf{90.0}${\scriptsize $\mathbf{/1.2}$} & \cellcolor{lightgreen}\strut $\mathbf{40.0}${\scriptsize $\mathbf{/1.2}$} & \cellcolor{lightgreen}\strut \underline{$\mathbf{10.0}${\scriptsize $\mathbf{/1.2}$}} & $\mathbf{65.0}${\scriptsize $\mathbf{/1.2}$} & \cellcolor{lightgreen}\strut \underline{$\mathbf{87.5}${\scriptsize $\mathbf{/1.2}$}} & \underline{$\mathbf{43.5}${\scriptsize $\mathbf{/1.2}$}} & \cellcolor{lightgreen}\strut \underline{$\mathbf{55.0}${\scriptsize $\mathbf{/1.3}$}} & $\mathbf{81.0}${\scriptsize $\mathbf{/1.3}$} & \underline{$\mathbf{65.0}${\scriptsize $\mathbf{/1.2}$}} & \cellcolor{lightgreen}\strut \underline{$\mathbf{30.0}${\scriptsize $\mathbf{/1.2}$}} \\
\bottomrule
\end{tabular}%
}
\end{table*}

\providecolor{lightgreen}{RGB}{239,248,239}
\begin{table*}[t]
\centering
\caption{HELMET per-task quality--latency comparison.}
\label{tab:helmet-per-task-acc-accl}
\setlength{\tabcolsep}{1.0pt}
\renewcommand{\arraystretch}{1.10}
\small
\resizebox{\textwidth}{!}{%
\begin{tabular}{@{}c@{\hspace{0.5pt}}l@{\hspace{0.5pt}}c@{\hspace{0.5pt}}*{15}{c}@{}}
\toprule
 & Method & \multicolumn{4}{c}{RAG} & \multicolumn{2}{c}{LC-QA} & \multicolumn{5}{c}{ICL} & \multicolumn{4}{c}{Recall} & \multicolumn{1}{c}{Rerank.} \\[-2pt]
\cmidrule(lr){3-6}\cmidrule(lr){7-8}\cmidrule(lr){9-13}\cmidrule(lr){14-17}\cmidrule(l){18-18}
 & & \rotatebox{60}{\scriptsize KILT-Hotpot} & \rotatebox{60}{\scriptsize KILT-NQ} & \rotatebox{60}{\scriptsize KILT-PopQA} & \rotatebox{60}{\scriptsize KILT-Trivia} & \rotatebox{60}{\scriptsize InfB-Choice} & \rotatebox{60}{\scriptsize InfB-QA} & \rotatebox{60}{\scriptsize ICL-Bank77} & \rotatebox{60}{\scriptsize ICL-CLINC} & \rotatebox{60}{\scriptsize ICL-NLU} & \rotatebox{60}{\scriptsize ICL-TREC-C} & \rotatebox{60}{\scriptsize ICL-TREC-F} & \rotatebox{60}{\scriptsize JSON-KV} & \rotatebox{60}{\scriptsize NIAH-MK2} & \rotatebox{60}{\scriptsize NIAH-MK3} & \rotatebox{60}{\scriptsize NIAH-MV} & \rotatebox{60}{\scriptsize MSMARCO} \\
\midrule
\multirow{12}{*}{\rotatebox[origin=c]{90}{\textbf{Llama}}} & Full & $60.0${\scriptsize $/1.0$} & $47.5${\scriptsize $/1.0$} & $54.2${\scriptsize $/1.0$} & $80.0${\scriptsize $/1.0$} & $50.0${\scriptsize $/1.0$} & $47.1${\scriptsize $/1.0$} & $90.0${\scriptsize $/1.0$} & $95.0${\scriptsize $/1.0$} & $85.0${\scriptsize $/1.0$} & $80.0${\scriptsize $/1.0$} & $35.0${\scriptsize $/1.0$} & $90.0${\scriptsize $/1.0$} & $90.0${\scriptsize $/1.0$} & $90.0${\scriptsize $/1.0$} & $96.2${\scriptsize $/1.0$} & $15.5${\scriptsize $/1.0$} \\
 & Snap $10\%$ & \cellcolor{lightgreen}\strut \underline{$61.7${\scriptsize $/1.7$}} & $44.2${\scriptsize $/1.7$} & \cellcolor{lightgreen}\strut \underline{$55.0${\scriptsize $/1.6$}} & \cellcolor{lightgreen}\strut \underline{$83.3${\scriptsize $/1.7$}} & \cellcolor{lightgreen}\strut \underline{$50.0${\scriptsize $/1.6$}} & \cellcolor{lightgreen}\strut \underline{$46.8${\scriptsize $/1.6$}} & \cellcolor{lightgreen}\strut \underline{$90.0${\scriptsize $/1.6$}} & \cellcolor{lightgreen}\strut \underline{$95.0${\scriptsize $/1.6$}} & \cellcolor{lightgreen}\strut \underline{$85.0${\scriptsize $/1.6$}} & \cellcolor{lightgreen}\strut \underline{$80.0${\scriptsize $/1.6$}} & \cellcolor{lightgreen}\strut \underline{$35.0${\scriptsize $/1.6$}} & $75.0${\scriptsize $/1.5$} & \cellcolor{lightgreen}\strut \underline{$95.0${\scriptsize $/1.5$}} & $35.0${\scriptsize $/1.5$} & \cellcolor{lightgreen}\strut $95.0${\scriptsize $/1.6$} & $3.3${\scriptsize $/1.5$} \\
 & Snap $25\%$ & \underline{$61.7${\scriptsize $/1.5$}} & $45.0${\scriptsize $/1.5$} & \underline{$55.0${\scriptsize $/1.5$}} & \underline{$80.8${\scriptsize $/1.5$}} & \underline{$50.0${\scriptsize $/1.5$}} & \cellcolor{lightgreen}\strut \underline{$48.9${\scriptsize $/1.5$}} & \underline{$90.0${\scriptsize $/1.5$}} & \underline{$95.0${\scriptsize $/1.5$}} & \underline{$85.0${\scriptsize $/1.5$}} & \underline{$80.0${\scriptsize $/1.5$}} & \cellcolor{lightgreen}\strut \underline{$40.0${\scriptsize $/1.4$}} & \cellcolor{lightgreen}\strut \underline{$90.0${\scriptsize $/1.4$}} & \underline{$95.0${\scriptsize $/1.4$}} & $70.0${\scriptsize $/1.4$} & \cellcolor{lightgreen}\strut \underline{$98.8${\scriptsize $/1.5$}} & $5.0${\scriptsize $/1.4$} \\
 & Str. $10\%$ & $46.7${\scriptsize $/1.7$} & $35.8${\scriptsize $/1.7$} & $53.3${\scriptsize $/1.6$} & $78.3${\scriptsize $/1.7$} & \cellcolor{lightgreen}\strut \underline{$50.0${\scriptsize $/1.6$}} & $34.8${\scriptsize $/1.6$} & \cellcolor{lightgreen}\strut \underline{$90.0${\scriptsize $/1.6$}} & $90.0${\scriptsize $/1.6$} & \cellcolor{lightgreen}\strut \underline{$85.0${\scriptsize $/1.6$}} & \cellcolor{lightgreen}\strut \underline{$90.0${\scriptsize $/1.6$}} & $20.0${\scriptsize $/1.6$} & $0.0${\scriptsize $/1.5$} & $25.0${\scriptsize $/1.5$} & $5.0${\scriptsize $/1.5$} & $10.0${\scriptsize $/1.6$} & $2.8${\scriptsize $/1.5$} \\
 & Str. $25\%$ & $55.0${\scriptsize $/1.5$} & $37.5${\scriptsize $/1.5$} & \cellcolor{lightgreen}\strut \underline{$58.3${\scriptsize $/1.5$}} & \underline{$80.8${\scriptsize $/1.5$}} & \underline{$50.0${\scriptsize $/1.5$}} & $33.0${\scriptsize $/1.5$} & \underline{$90.0${\scriptsize $/1.5$}} & \underline{$95.0${\scriptsize $/1.5$}} & \underline{$85.0${\scriptsize $/1.5$}} & $75.0${\scriptsize $/1.5$} & $20.0${\scriptsize $/1.4$} & $10.0${\scriptsize $/1.4$} & $40.0${\scriptsize $/1.4$} & $20.0${\scriptsize $/1.4$} & $15.0${\scriptsize $/1.5$} & $1.4${\scriptsize $/1.4$} \\
 & Pyr. $10\%$ & \underline{$60.0${\scriptsize $/1.7$}} & \cellcolor{lightgreen}\strut $45.8${\scriptsize $/1.7$} & \cellcolor{lightgreen}\strut \underline{$55.0${\scriptsize $/1.6$}} & \cellcolor{lightgreen}\strut \underline{$83.3${\scriptsize $/1.7$}} & \cellcolor{lightgreen}\strut \underline{$50.0${\scriptsize $/1.6$}} & \cellcolor{lightgreen}\strut \underline{$48.1${\scriptsize $/1.6$}} & \cellcolor{lightgreen}\strut \underline{$90.0${\scriptsize $/1.6$}} & \cellcolor{lightgreen}\strut \underline{$95.0${\scriptsize $/1.6$}} & \cellcolor{lightgreen}\strut \underline{$85.0${\scriptsize $/1.6$}} & \underline{$80.0${\scriptsize $/1.6$}} & $25.0${\scriptsize $/1.6$} & $80.0${\scriptsize $/1.5$} & \cellcolor{lightgreen}\strut \underline{$95.0${\scriptsize $/1.5$}} & $45.0${\scriptsize $/1.5$} & \cellcolor{lightgreen}\strut \underline{$97.5${\scriptsize $/1.6$}} & $2.1${\scriptsize $/1.5$} \\
 & Pyr. $25\%$ & \underline{$60.0${\scriptsize $/1.5$}} & $44.2${\scriptsize $/1.5$} & \underline{$54.2${\scriptsize $/1.4$}} & \underline{$83.3${\scriptsize $/1.5$}} & \underline{$50.0${\scriptsize $/1.5$}} & \cellcolor{lightgreen}\strut \underline{$48.9${\scriptsize $/1.5$}} & \underline{$90.0${\scriptsize $/1.5$}} & \underline{$95.0${\scriptsize $/1.5$}} & \underline{$85.0${\scriptsize $/1.5$}} & \underline{$80.0${\scriptsize $/1.5$}} & $30.0${\scriptsize $/1.4$} & \cellcolor{lightgreen}\strut \underline{$95.0${\scriptsize $/1.4$}} & \underline{$90.0${\scriptsize $/1.4$}} & $80.0${\scriptsize $/1.4$} & \cellcolor{lightgreen}\strut \underline{$98.8${\scriptsize $/1.5$}} & $8.5${\scriptsize $/1.4$} \\
 & Quest & $51.7${\scriptsize $/1.1$} & $40.8${\scriptsize $/1.1$} & $47.5${\scriptsize $/1.1$} & $76.7${\scriptsize $/1.1$} & \underline{$50.0${\scriptsize $/1.1$}} & \underline{$47.5${\scriptsize $/1.1$}} & \cellcolor{lightgreen}\strut \underline{$95.0${\scriptsize $/1.1$}} & \underline{$95.0${\scriptsize $/1.1$}} & $80.0${\scriptsize $/1.1$} & \underline{$80.0${\scriptsize $/1.1$}} & \underline{$35.0${\scriptsize $/1.1$}} & $5.0${\scriptsize $/1.1$} & \underline{$90.0${\scriptsize $/1.1$}} & $0.0${\scriptsize $/1.1$} & $93.8${\scriptsize $/1.1$} & \cellcolor{lightgreen}\strut \underline{$16.4${\scriptsize $/1.0$}} \\
 & Double & $58.3${\scriptsize $/1.4$} & $46.7${\scriptsize $/1.4$} & \underline{$55.0${\scriptsize $/1.3$}} & \cellcolor{lightgreen}\strut \underline{$84.2${\scriptsize $/1.4$}} & \underline{$50.0${\scriptsize $/1.3$}} & \underline{$47.0${\scriptsize $/1.3$}} & \underline{$90.0${\scriptsize $/1.3$}} & \underline{$95.0${\scriptsize $/1.3$}} & \underline{$85.0${\scriptsize $/1.3$}} & \underline{$85.0${\scriptsize $/1.3$}} & $30.0${\scriptsize $/1.3$} & $25.0${\scriptsize $/1.3$} & $60.0${\scriptsize $/1.2$} & $15.0${\scriptsize $/1.2$} & $57.5${\scriptsize $/1.3$} & $6.1${\scriptsize $/1.2$} \\
 & {\bfseries\boldmath Our $\tau.01$} & \underline{$\mathbf{60.0}${\scriptsize $\mathbf{/1.5}$}} & \cellcolor{lightgreen}\strut \underline{$\mathbf{47.5}${\scriptsize $\mathbf{/1.5}$}} & \underline{$\mathbf{55.8}${\scriptsize $\mathbf{/1.4}$}} & \underline{$\mathbf{83.3}${\scriptsize $\mathbf{/1.5}$}} & \underline{$\mathbf{50.0}${\scriptsize $\mathbf{/1.4}$}} & \underline{$\mathbf{47.8}${\scriptsize $\mathbf{/1.5}$}} & \underline{$\mathbf{90.0}${\scriptsize $\mathbf{/1.4}$}} & \underline{$\mathbf{95.0}${\scriptsize $\mathbf{/1.4}$}} & \underline{$\mathbf{85.0}${\scriptsize $\mathbf{/1.4}$}} & \underline{$\mathbf{80.0}${\scriptsize $\mathbf{/1.4}$}} & $\mathbf{25.0}${\scriptsize $\mathbf{/1.4}$} & \underline{$\mathbf{90.0}${\scriptsize $\mathbf{/1.4}$}} & \underline{$\mathbf{95.0}${\scriptsize $\mathbf{/1.4}$}} & $\mathbf{80.0}${\scriptsize $\mathbf{/1.4}$} & \underline{$\mathbf{96.2}${\scriptsize $\mathbf{/1.5}$}} & $\mathbf{14.2}${\scriptsize $\mathbf{/1.3}$} \\
 & {\bfseries\boldmath Our $\tau.1$} & $\mathbf{55.0}${\scriptsize $\mathbf{/1.6}$} & $\mathbf{40.8}${\scriptsize $\mathbf{/1.6}$} & $\mathbf{53.3}${\scriptsize $\mathbf{/1.6}$} & \underline{$\mathbf{81.7}${\scriptsize $\mathbf{/1.6}$}} & \cellcolor{lightgreen}\strut \underline{$\mathbf{50.0}${\scriptsize $\mathbf{/1.6}$}} & $\mathbf{44.0}${\scriptsize $\mathbf{/1.6}$} & \cellcolor{lightgreen}\strut \underline{$\mathbf{90.0}${\scriptsize $\mathbf{/1.6}$}} & \cellcolor{lightgreen}\strut \underline{$\mathbf{95.0}${\scriptsize $\mathbf{/1.6}$}} & $\mathbf{75.0}${\scriptsize $\mathbf{/1.6}$} & \underline{$\mathbf{80.0}${\scriptsize $\mathbf{/1.6}$}} & \cellcolor{lightgreen}\strut \underline{$\mathbf{40.0}${\scriptsize $\mathbf{/1.5}$}} & $\mathbf{30.0}${\scriptsize $\mathbf{/1.5}$} & $\mathbf{80.0}${\scriptsize $\mathbf{/1.5}$} & $\mathbf{15.0}${\scriptsize $\mathbf{/1.5}$} & $\mathbf{70.0}${\scriptsize $\mathbf{/1.6}$} & $\mathbf{9.1}${\scriptsize $\mathbf{/1.4}$} \\
 & {\bfseries\boldmath Our $\tau.1$ Q32} & $\mathbf{58.3}${\scriptsize $\mathbf{/1.6}$} & $\mathbf{45.0}${\scriptsize $\mathbf{/1.6}$} & \cellcolor{lightgreen}\strut \underline{$\mathbf{55.0}${\scriptsize $\mathbf{/1.5}$}} & \cellcolor{lightgreen}\strut \underline{$\mathbf{83.3}${\scriptsize $\mathbf{/1.6}$}} & \cellcolor{lightgreen}\strut \underline{$\mathbf{50.0}${\scriptsize $\mathbf{/1.6}$}} & \underline{$\mathbf{48.0}${\scriptsize $\mathbf{/1.6}$}} & \underline{$\mathbf{90.0}${\scriptsize $\mathbf{/1.5}$}} & \underline{$\mathbf{95.0}${\scriptsize $\mathbf{/1.5}$}} & \underline{$\mathbf{85.0}${\scriptsize $\mathbf{/1.5}$}} & $\mathbf{75.0}${\scriptsize $\mathbf{/1.5}$} & $\mathbf{30.0}${\scriptsize $\mathbf{/1.5}$} & \cellcolor{lightgreen}\strut \underline{$\mathbf{90.0}${\scriptsize $\mathbf{/1.5}$}} & \cellcolor{lightgreen}\strut \underline{$\mathbf{95.0}${\scriptsize $\mathbf{/1.5}$}} & $\mathbf{60.0}${\scriptsize $\mathbf{/1.5}$} & $\mathbf{87.5}${\scriptsize $\mathbf{/1.6}$} & \cellcolor{lightgreen}\strut \underline{$\mathbf{15.3}${\scriptsize $\mathbf{/1.4}$}} \\
\midrule
\multirow{12}{*}{\rotatebox[origin=c]{90}{\textbf{Qwen}}} & Full & $41.7${\scriptsize $/1.0$} & $30.0${\scriptsize $/1.0$} & $48.3${\scriptsize $/1.0$} & $74.2${\scriptsize $/1.0$} & $70.0${\scriptsize $/1.0$} & $31.7${\scriptsize $/1.0$} & $85.0${\scriptsize $/1.0$} & $95.0${\scriptsize $/1.0$} & $80.0${\scriptsize $/1.0$} & $95.0${\scriptsize $/1.0$} & $25.0${\scriptsize $/1.0$} & $40.0${\scriptsize $/1.0$} & $50.0${\scriptsize $/1.0$} & $35.0${\scriptsize $/1.0$} & $65.0${\scriptsize $/1.0$} & $7.3${\scriptsize $/1.0$} \\
 & Snap $10\%$ & \underline{$41.7${\scriptsize $/1.3$}} & \cellcolor{lightgreen}\strut $29.2${\scriptsize $/1.3$} & $37.5${\scriptsize $/1.3$} & \underline{$78.3${\scriptsize $/1.3$}} & \cellcolor{lightgreen}\strut \underline{$70.0${\scriptsize $/1.3$}} & \cellcolor{lightgreen}\strut $29.8${\scriptsize $/1.3$} & \cellcolor{lightgreen}\strut $80.0${\scriptsize $/1.3$} & $90.0${\scriptsize $/1.3$} & \cellcolor{lightgreen}\strut \underline{$85.0${\scriptsize $/1.3$}} & \cellcolor{lightgreen}\strut $90.0${\scriptsize $/1.3$} & \underline{$25.0${\scriptsize $/1.3$}} & $25.0${\scriptsize $/1.3$} & $0.0${\scriptsize $/1.3$} & $0.0${\scriptsize $/1.3$} & $37.5${\scriptsize $/1.3$} & $0.8${\scriptsize $/1.2$} \\
 & Snap $25\%$ & \underline{$41.7${\scriptsize $/1.3$}} & $28.3${\scriptsize $/1.3$} & \cellcolor{lightgreen}\strut $46.7${\scriptsize $/1.2$} & \underline{$80.0${\scriptsize $/1.3$}} & \cellcolor{lightgreen}\strut \underline{$70.0${\scriptsize $/1.2$}} & $28.2${\scriptsize $/1.2$} & \underline{$85.0${\scriptsize $/1.2$}} & \cellcolor{lightgreen}\strut \underline{$95.0${\scriptsize $/1.2$}} & $75.0${\scriptsize $/1.2$} & \cellcolor{lightgreen}\strut \underline{$95.0${\scriptsize $/1.2$}} & \underline{$25.0${\scriptsize $/1.2$}} & $30.0${\scriptsize $/1.3$} & $5.0${\scriptsize $/1.2$} & $15.0${\scriptsize $/1.2$} & \cellcolor{lightgreen}\strut $60.0${\scriptsize $/1.2$} & $1.0${\scriptsize $/1.2$} \\
 & Str. $10\%$ & $40.0${\scriptsize $/1.3$} & $25.8${\scriptsize $/1.3$} & $24.2${\scriptsize $/1.3$} & $62.5${\scriptsize $/1.3$} & \cellcolor{lightgreen}\strut \underline{$70.0${\scriptsize $/1.3$}} & $22.7${\scriptsize $/1.3$} & \cellcolor{lightgreen}\strut $80.0${\scriptsize $/1.3$} & $75.0${\scriptsize $/1.3$} & \underline{$80.0${\scriptsize $/1.3$}} & $85.0${\scriptsize $/1.3$} & \cellcolor{lightgreen}\strut \underline{$30.0${\scriptsize $/1.3$}} & $0.0${\scriptsize $/1.3$} & $20.0${\scriptsize $/1.3$} & $0.0${\scriptsize $/1.3$} & $10.0${\scriptsize $/1.3$} & $1.5${\scriptsize $/1.2$} \\
 & Str. $25\%$ & $38.3${\scriptsize $/1.3$} & $22.5${\scriptsize $/1.3$} & $37.5${\scriptsize $/1.2$} & $60.8${\scriptsize $/1.3$} & \cellcolor{lightgreen}\strut \underline{$70.0${\scriptsize $/1.2$}} & $24.7${\scriptsize $/1.2$} & \cellcolor{lightgreen}\strut \underline{$90.0${\scriptsize $/1.2$}} & $90.0${\scriptsize $/1.2$} & \underline{$80.0${\scriptsize $/1.2$}} & $90.0${\scriptsize $/1.2$} & \underline{$25.0${\scriptsize $/1.2$}} & $0.0${\scriptsize $/1.3$} & $25.0${\scriptsize $/1.2$} & $10.0${\scriptsize $/1.2$} & $13.8${\scriptsize $/1.2$} & $2.2${\scriptsize $/1.2$} \\
 & Pyr. $10\%$ & \cellcolor{lightgreen}\strut \underline{$45.0${\scriptsize $/1.3$}} & \cellcolor{lightgreen}\strut $26.7${\scriptsize $/1.3$} & $36.7${\scriptsize $/1.3$} & \cellcolor{lightgreen}\strut \underline{$80.8${\scriptsize $/1.3$}} & \cellcolor{lightgreen}\strut \underline{$70.0${\scriptsize $/1.3$}} & \cellcolor{lightgreen}\strut $27.9${\scriptsize $/1.3$} & $70.0${\scriptsize $/1.3$} & \cellcolor{lightgreen}\strut \underline{$95.0${\scriptsize $/1.3$}} & \underline{$80.0${\scriptsize $/1.3$}} & $85.0${\scriptsize $/1.3$} & \underline{$25.0${\scriptsize $/1.3$}} & $25.0${\scriptsize $/1.3$} & $0.0${\scriptsize $/1.3$} & $0.0${\scriptsize $/1.3$} & $26.2${\scriptsize $/1.3$} & $0.0${\scriptsize $/1.2$} \\
 & Pyr. $25\%$ & \underline{$45.0${\scriptsize $/1.3$}} & $26.7${\scriptsize $/1.2$} & $35.8${\scriptsize $/1.2$} & \underline{$75.0${\scriptsize $/1.3$}} & \cellcolor{lightgreen}\strut \underline{$70.0${\scriptsize $/1.2$}} & $27.4${\scriptsize $/1.2$} & $80.0${\scriptsize $/1.2$} & $90.0${\scriptsize $/1.2$} & \underline{$80.0${\scriptsize $/1.2$}} & \cellcolor{lightgreen}\strut \underline{$95.0${\scriptsize $/1.2$}} & \underline{$30.0${\scriptsize $/1.2$}} & $25.0${\scriptsize $/1.2$} & $0.0${\scriptsize $/1.2$} & $0.0${\scriptsize $/1.2$} & $42.5${\scriptsize $/1.2$} & $0.0${\scriptsize $/1.2$} \\
 & Quest & $36.7${\scriptsize $/0.9$} & $21.7${\scriptsize $/0.9$} & $36.7${\scriptsize $/0.9$} & $68.3${\scriptsize $/0.9$} & $65.0${\scriptsize $/0.9$} & \underline{$34.4${\scriptsize $/0.9$}} & \underline{$85.0${\scriptsize $/0.9$}} & $70.0${\scriptsize $/0.9$} & $75.0${\scriptsize $/0.9$} & $90.0${\scriptsize $/0.9$} & $20.0${\scriptsize $/0.9$} & $0.0${\scriptsize $/0.9$} & $25.0${\scriptsize $/0.9$} & $0.0${\scriptsize $/0.9$} & $56.2${\scriptsize $/0.9$} & $5.5${\scriptsize $/0.9$} \\
 & Double & $38.3${\scriptsize $/1.1$} & \cellcolor{lightgreen}\strut \underline{$37.5${\scriptsize $/1.1$}} & $37.5${\scriptsize $/1.0$} & \underline{$75.8${\scriptsize $/1.1$}} & \underline{$70.0${\scriptsize $/1.1$}} & $30.8${\scriptsize $/1.0$} & $80.0${\scriptsize $/1.0$} & \underline{$95.0${\scriptsize $/1.0$}} & $75.0${\scriptsize $/1.0$} & $85.0${\scriptsize $/1.0$} & \cellcolor{lightgreen}\strut \underline{$40.0${\scriptsize $/1.0$}} & $0.0${\scriptsize $/1.1$} & $25.0${\scriptsize $/1.0$} & $0.0${\scriptsize $/1.1$} & $33.8${\scriptsize $/1.0$} & \cellcolor{lightgreen}\strut \underline{$7.6${\scriptsize $/1.0$}} \\
 & {\bfseries\boldmath Our $\tau.01$} & \underline{$\mathbf{41.7}${\scriptsize $\mathbf{/1.2}$}} & \underline{$\mathbf{31.7}${\scriptsize $\mathbf{/1.1}$}} & $\mathbf{46.7}${\scriptsize $\mathbf{/1.1}$} & \underline{$\mathbf{76.7}${\scriptsize $\mathbf{/1.2}$}} & \underline{$\mathbf{70.0}${\scriptsize $\mathbf{/1.2}$}} & \cellcolor{lightgreen}\strut \underline{$\mathbf{31.7}${\scriptsize $\mathbf{/1.1}$}} & \underline{$\mathbf{85.0}${\scriptsize $\mathbf{/1.1}$}} & \underline{$\mathbf{95.0}${\scriptsize $\mathbf{/1.1}$}} & $\mathbf{75.0}${\scriptsize $\mathbf{/1.1}$} & \underline{$\mathbf{95.0}${\scriptsize $\mathbf{/1.1}$}} & \underline{$\mathbf{30.0}${\scriptsize $\mathbf{/1.1}$}} & $\mathbf{35.0}${\scriptsize $\mathbf{/1.2}$} & \underline{$\mathbf{50.0}${\scriptsize $\mathbf{/1.1}$}} & \cellcolor{lightgreen}\strut $\mathbf{30.0}${\scriptsize $\mathbf{/1.2}$} & \cellcolor{lightgreen}\strut \underline{$\mathbf{73.8}${\scriptsize $\mathbf{/1.1}$}} & $\mathbf{4.7}${\scriptsize $\mathbf{/1.1}$} \\
 & {\bfseries\boldmath Our $\tau.1$} & $\mathbf{40.0}${\scriptsize $\mathbf{/1.3}$} & \cellcolor{lightgreen}\strut \underline{$\mathbf{32.5}${\scriptsize $\mathbf{/1.2}$}} & \cellcolor{lightgreen}\strut $\mathbf{47.5}${\scriptsize $\mathbf{/1.2}$} & \underline{$\mathbf{77.5}${\scriptsize $\mathbf{/1.3}$}} & \cellcolor{lightgreen}\strut \underline{$\mathbf{70.0}${\scriptsize $\mathbf{/1.2}$}} & \cellcolor{lightgreen}\strut \underline{$\mathbf{31.5}${\scriptsize $\mathbf{/1.2}$}} & \underline{$\mathbf{85.0}${\scriptsize $\mathbf{/1.2}$}} & \underline{$\mathbf{95.0}${\scriptsize $\mathbf{/1.2}$}} & $\mathbf{75.0}${\scriptsize $\mathbf{/1.2}$} & \cellcolor{lightgreen}\strut \underline{$\mathbf{95.0}${\scriptsize $\mathbf{/1.2}$}} & \underline{$\mathbf{30.0}${\scriptsize $\mathbf{/1.2}$}} & \cellcolor{lightgreen}\strut $\mathbf{35.0}${\scriptsize $\mathbf{/1.3}$} & \cellcolor{lightgreen}\strut \underline{$\mathbf{50.0}${\scriptsize $\mathbf{/1.2}$}} & $\mathbf{15.0}${\scriptsize $\mathbf{/1.3}$} & \cellcolor{lightgreen}\strut $\mathbf{61.2}${\scriptsize $\mathbf{/1.2}$} & \cellcolor{lightgreen}\strut $\mathbf{5.1}${\scriptsize $\mathbf{/1.2}$} \\
 & {\bfseries\boldmath Our $\tau.1$ Q32} & \underline{$\mathbf{41.7}${\scriptsize $\mathbf{/1.2}$}} & \underline{$\mathbf{30.0}${\scriptsize $\mathbf{/1.2}$}} & \cellcolor{lightgreen}\strut \underline{$\mathbf{50.0}${\scriptsize $\mathbf{/1.2}$}} & \underline{$\mathbf{75.8}${\scriptsize $\mathbf{/1.2}$}} & \underline{$\mathbf{70.0}${\scriptsize $\mathbf{/1.2}$}} & \underline{$\mathbf{31.5}${\scriptsize $\mathbf{/1.2}$}} & \underline{$\mathbf{85.0}${\scriptsize $\mathbf{/1.1}$}} & \underline{$\mathbf{95.0}${\scriptsize $\mathbf{/1.1}$}} & $\mathbf{75.0}${\scriptsize $\mathbf{/1.1}$} & \underline{$\mathbf{95.0}${\scriptsize $\mathbf{/1.1}$}} & \underline{$\mathbf{25.0}${\scriptsize $\mathbf{/1.1}$}} & \cellcolor{lightgreen}\strut \underline{$\mathbf{40.0}${\scriptsize $\mathbf{/1.2}$}} & \cellcolor{lightgreen}\strut \underline{$\mathbf{50.0}${\scriptsize $\mathbf{/1.2}$}} & $\mathbf{25.0}${\scriptsize $\mathbf{/1.2}$} & \cellcolor{lightgreen}\strut \underline{$\mathbf{65.0}${\scriptsize $\mathbf{/1.2}$}} & $\mathbf{5.0}${\scriptsize $\mathbf{/1.1}$} \\
\bottomrule
\end{tabular}%
}
\end{table*}

\newpage
\subsection{Risk Score Analysis and Ablations}
\begin{equation}
\lVert \boldsymbol{o}-\widehat{\boldsymbol{o}} \rVert_2
\leq
2 \sum_{C\in\mathcal C}
\underbrace{\widehat p_C \left( V_{\max} \frac{\cosh(\delta_C^{\mathrm{bound}})-1}{\sum_{C'\in\mathcal C} \widehat p_{C'} \cosh(\delta_{C'}^{\mathrm{bound}})}
+ V_C \tanh\left(\frac{\delta_C^{\mathrm{bound}}}{2}\right) \right)}_{\text{Risk Score $R_C$ for cluster \(C\)}}.
\end{equation}

Our risk score is derived by regrouping the attention-output error into cluster-wise terms and bounding their magnitudes. The resulting score captures the joint effect of global importance, measured by $\widehat p_C$, and local distortion, characterized by 
\[\delta_C = \max_{i\in C} |\boldsymbol{q}^\top(\boldsymbol{k}_i-\bar{\boldsymbol{k}}_C)/\sqrt d|.\]

Computing $\delta_C$ exactly requires accessing all keys within a cluster, so we replace it with an efficiently computable upper bound $\delta_C^{\mathrm{bound}}$ in practice. This raises three questions:
\begin{itemize}
    \item How does a large $\delta_C^{\mathrm{bound}}$ affect the risk score, and can the score be simplified?
    \item How does using the exact, oracle $\delta_C$ affect selection quality?
    \item How does our risk score compare with simpler selection rules?
\end{itemize}
We investigate these questions in the following subsections.

\subsubsection{Score Simplification}

\paragraph{Simplified score.}
When $\delta_C$ is large, $\cosh(x) \approx e^{x}/2$ and $\tanh(x) \approx 1$, so $R_C$ simplifies to
\begin{equation}
  \tilde R_C \;=\; V_{\max}\,\operatorname{softmax}_C\!\left(\log \hat p_C + \delta_C\right) \;+\; \hat p_C V_C .
  \label{eq:simplified-rc}
\end{equation}

\paragraph{Setting.}
Llama-3.1-8B, RULER-128K, $Q=0$, GQA-any, block size 32.
We record every layer and query head of 26 prompts (2 per task), at $\tau=.01$ and $\tau=.1$.
\emph{Same set} is the fraction of KV heads for which $\tilde R_C$ selects exactly the same clusters as $R_C$ at the same $\tau$.
\emph{Cluster Jaccard} is $|A\cap B|/|A\cup B|$ between the two selected cluster sets, averaged over KV heads.
\emph{Eq.~4 $> 2V_{\max}$} is the fraction of heads for which Eq.~4 exceeds the trivial bound $2V_{\max}$.

\paragraph{Results.}
Table~\ref{tab:simplified-rc} shows that, with the oracle deviation, Eq.~4 is well below the trivial bound.
In practice, with the box or ellipsoid bound, $R_C$ and $\tilde R_C$ select the same clusters on 96--98\% of KV heads.

\begin{table}[ht]
  \centering
  \caption{Agreement between $R_C$ and the simplified score $\tilde R_C$ (Eq.~\ref{eq:simplified-rc}), and how often Eq.~4 exceeds the trivial bound $2V_{\max}$.
  Llama-3.1-8B, RULER-128K, $Q=0$, GQA-any, block size 32, 26 prompts.
  $\delta_C$ is reported as mean $\pm$ std.}
  \label{tab:simplified-rc}
  \small
  \begin{tabular}{lccccccc}
    \toprule
    & & \multicolumn{2}{c}{Same set (\%)} & \multicolumn{2}{c}{Cluster Jaccard} & \multicolumn{2}{c}{Eq.~4 $> 2V_{\max}$ (\%)} \\
    \cmidrule(lr){3-4} \cmidrule(lr){5-6} \cmidrule(lr){7-8}
    Deviation $\delta_C$ & $\delta_C$ & $\tau=.01$ & $\tau=.1$ & $\tau=.01$ & $\tau=.1$ & $\tau=.01$ & $\tau=.1$ \\
    \midrule
    Box            & $21.92 \pm 5.31$ & 96.4 & 97.7 & 0.9999 & 0.9991 & 99.5 & 99.7 \\
    Diag.\ ellipsoid & $23.48 \pm 7.59$ & 96.4 & 97.7 & 0.9998 & 0.9982 & 99.7 & 99.8 \\
    Exact (oracle) & $\phantom{0}4.17 \pm 1.79$ & \phantom{0}2.2 & 38.4 & 0.918\phantom{0} & 0.945\phantom{0} & \phantom{0}1.7 & \phantom{0}5.0 \\
    \bottomrule
  \end{tabular}
\end{table}

\paragraph{Simplified $R_C$ does not invalidate RefineKV.} RefineKV uses $R_C$ only through the ranking it over clusters. The simplified form, keeps the intuition behind RefineKV: we want to jointly capture Importance $\hat p_C$ and Distortion $\delta_C$. Both signals enter jointly through $\mathrm{softmax}_C(\log \hat{p}_C + \delta_C)$, and a mass-weighted value term $\hat{p}_C V_C$ is added.

\subsubsection{Selection with Oracle $\delta$ and simplified risk score}

To separate the deviation estimate from the form of the score, we fix the per-head token budget to that selected by $R_C$ with the box $\delta_C^{\mathrm{bound}}$ (Ours) and change only the cluster ranking (Table~\ref{tab:oracle-simplified}).
A tighter deviation ranks clusters better: at the low budget ($\tau=0.1$), the box summary improves on the ellipsoid by $5.3$ points (Table~\ref{tab:ablation-score}), and the non-deployable oracle adds another $6.5$; at $\tau=0.01$ the oracle comes within $1.4$ points of Full Attention.
In contrast, replacing $R_C$ by its large-$\delta$ simplification $V_{\max}\,\mathrm{softmax}_C(\log\hat p_C+\delta_C)+\hat p_C V_C$ changes accuracy by at most $1.1$ points, and no paired 95\% CI excludes zero.

\begin{table}[ht]
\centering
\caption{Selection with oracle and simplified $R_C$ at a fixed per-head token budget
(Llama-3.1-8B-Instruct, RULER-128K, $Q{=}0$).
Every row expands exactly as many tokens per KV head as $R_C$ with the box $\delta_C^{\mathrm{bound}}$ (Ours) selects at the same decoding state; only the cluster selection differs.}
\label{tab:oracle-simplified}
\begin{tabular}{llcc}
\toprule
$R_C$ & $\delta_C$ & $\tau=0.01$ & $\tau=0.1$ \\
\midrule
original   & box    & 70.27 & 62.58 \\
simplified & box    & 70.19 & 62.67 \\
\midrule
original   & oracle & 72.00 & \textbf{69.06} \\
simplified & oracle & \textbf{72.13} & 68.01 \\
\midrule
\multicolumn{2}{l}{\textit{Full Attention}} & \multicolumn{2}{c}{73.37} \\
\bottomrule
\end{tabular}
\end{table}

\subsection{Results on Qwen2.5-14B-Instruct}
\label{app:qwen14b}
To test whether RefineKV and its thresholds carry over to a larger model, we evaluate Qwen2.5-14B-Instruct
(48 layers, 40 query / 8 KV heads) on RULER-128K (13 tasks $\times$ 20 samples).
We change nothing in the method: we use the same thresholds $\tau\in\{0.01, 0.1\}$, $|C|{=}32$, the box summary, GQA-any and $Q{=}0$,
and apply no retuning.
As for Qwen2.5-7B, the context is extended with static YaRN (factor 4, original length 32{,}768).
Full Attention and both RefineKV settings are timed in the same batch.
SnapKV-25\% uses the same CUDA Graph static-cache port as in the main experiments, and its same-batch Full Attention control measures 40.99\,ms (versus 40.98\,ms).
$\Delta$ is paired against Full Attention on the same 260 samples, with task-stratified bootstrap 95\% CIs.

\begin{table}[ht]
\centering
\caption{Qwen2.5-14B-Instruct on RULER-128K, with Qwen2.5-7B-Instruct for reference.
RefineKV uses the same $\tau$ on both models. $\Delta$ is relative to Full Attention, with 95\% paired bootstrap CIs.
Qwen2.5-7B numbers are those of the main results;}
\label{tab:qwen14b}
\small
\setlength{\tabcolsep}{4pt}
\begin{tabular}{l cccc ccc}
\toprule
 & \multicolumn{4}{c}{Qwen2.5-14B} & \multicolumn{3}{c}{Qwen2.5-7B} \\
\cmidrule(lr){2-5}\cmidrule(lr){6-8}
Method & Acc.$\uparrow$ & $\Delta$ [95\% CI] & TPOT (ms)$\downarrow$ & Speedup$\uparrow$ & Acc.$\uparrow$ & $\Delta$ & Speedup$\uparrow$ \\
\midrule
Full Attention          & 72.96 & ---                      & 40.98 & 1.00$\times$ & 65.88 & ---       & 1.00$\times$ \\
\textbf{RefineKV} ($\tau{=}0.01$) & \textbf{71.97} & $-0.99$ [$-2.88$, $+1.04$] & 29.56 & 1.39$\times$ & 65.28 & $-0.60$ & 1.19$\times$ \\
\textbf{RefineKV} ($\tau{=}0.1$)  & 69.19 & $-3.78$ [$-6.72$, $-0.91$] & 26.49 & \textbf{1.55$\times$} & 60.35 & $-5.53$ & 1.27$\times$ \\
SnapKV (25\%)           & 68.44 & $-4.52$ [$-7.50$, $-1.64$] & 28.33 & 1.45$\times$ & 51.05 & $-14.83$ & 1.27$\times$ \\
\bottomrule
\end{tabular}
\end{table}

\subsection{Speedup without CUDA Graph}
\label{app:eager}

The main results time Full Attention, RefineKV and the eviction baselines with CUDA Graph.
To show that the speedups do not depend on it, and to compare with methods whose custom kernels do not support CUDA Graph (Quest),
we also time RefineKV and Full Attention in eager mode with the same static cache and protocol.
RefineKV produces token-identical outputs in both modes, so quality is unchanged.
The eager speedups are 3--6\% lower than the CUDA Graph ones, because CUDA Graph removes more kernel-launch overhead from RefineKV
(eager/Graph TPOT 1.10--1.12) than from Full Attention (1.06--1.07).

\begin{table}[ht]
\centering
\caption{Quality and decode speedup of RefineKV over Full Attention, with both methods using CUDA Graph or both in eager mode.}
\label{tab:eager}
\small
\setlength{\tabcolsep}{3.5pt}
\begin{tabular}{ll cccccc}
\toprule
 & & \multicolumn{3}{c}{Llama-3.1-8B} & \multicolumn{3}{c}{Qwen2.5-7B} \\
\cmidrule(lr){3-5}\cmidrule(lr){6-8}
 & & RULER & HELMET & LongBench & RULER & HELMET & LongBench \\
\midrule
Full Attention & Quality & 73.37 & 69.10 & 48.86 & 65.88 & 54.57 & 49.34 \\
\midrule
\multirow{3}{*}{RefineKV, $\tau{=}0.01$}
 & Quality              & 70.19 & 68.43 & 48.87 & 65.28 & 54.48 & 49.23 \\
 & Speedup, CUDA  & 1.56$\times$ & 1.43$\times$ & 1.55$\times$ & 1.19$\times$ & 1.13$\times$ & 1.18$\times$ \\
 & Speedup, eager       & 1.49$\times$ & 1.37$\times$ & 1.49$\times$ & 1.15$\times$ & 1.10$\times$ & 1.15$\times$ \\
\midrule
\multirow{3}{*}{RefineKV, $\tau{=}0.1$}
 & Quality              & 62.58 & 56.80 & 48.95 & 60.35 & 52.83 & 49.13 \\
 & Speedup, CUDA  & 1.67$\times$ & 1.56$\times$ & 1.67$\times$ & 1.27$\times$ & 1.23$\times$ & 1.27$\times$ \\
 & Speedup, eager       & 1.59$\times$ & 1.47$\times$ & 1.59$\times$ & 1.23$\times$ & 1.18$\times$ & 1.22$\times$ \\
\bottomrule
\end{tabular}
\end{table}

\subsection{Context Length Ablation}
\label{app:length}

We hold the model (Llama-3.1-8B-Instruct), the task set (RULER, 13 tasks $\times$ 20 samples) and $\tau$ fixed, and vary the input length from 16K to 128K.
Full Attention and RefineKV see identical prompts at each length.
``Expanded'' is the fraction of visible KV entries loaded exactly at each decode step.

\begin{table}[h]
\centering
\caption{RULER accuracy(Acc.), expanded fraction(Exp.) and decode speed versus input length (Llama-3.1-8B-Instruct, fixed $\tau$).
$\Delta$ is relative to Full Attention.}
\label{tab:length}
\small
\setlength{\tabcolsep}{3pt}
\begin{tabular}{l cc cccc cccc}
\toprule
 & \multicolumn{2}{c}{Full Attention} & \multicolumn{4}{c}{RefineKV ($\tau{=}0.01$)} & \multicolumn{4}{c}{RefineKV ($\tau{=}0.1$)} \\
\cmidrule(lr){2-3}\cmidrule(lr){4-7}\cmidrule(lr){8-11}
Length & Acc. & TPOT & Acc. & $\Delta$ & Exp. & Speedup & Acc. & $\Delta$ & Exp. & Speedup \\
\midrule
16K  & 89.03 & 14.46 & 89.22 & $+0.19$ & 52.8\% & 1.01$\times$ & 85.34 & $-3.69$ & 10.9\% & 1.05$\times$ \\
32K  & 83.41 & 15.86 & 83.42 & $+0.01$ & 32.8\% & 1.07$\times$ & 75.12 & $-8.29$ & 6.7\% & 1.14$\times$ \\
64K  & 82.80 & 18.71 & 81.94 & $-0.86$ & 18.7\% & 1.24$\times$ & 74.74 & $-8.06$ & 4.6\% & 1.33$\times$ \\
128K & 73.35 & 24.43 & 70.27 & $-3.08$ & 11.2\% & 1.56$\times$ & 62.58 & $-10.77$ & 4.0\% & 1.67$\times$ \\
\bottomrule
\end{tabular}
\end{table}

At fixed $\tau$, the expanded fraction falls steadily with length.
At fixed $\tau$, the expanded fraction falls steadily with length, from 52.8\% at 16K to 11.2\% at 128K for $\tau{=}0.01$, and from 10.9\% to 4.0\% for $\tau{=}0.1$.
This is because $\tau$ bounds the per-step output error rather than fixing a budget, and on RULER most of the added context is filler that the cluster summaries approximate well.
As a result, RefineKV's TPOT stays nearly flat (14.4--15.7\,ms) while Full Attention's grows with the cache, and the speedup rises from about 1.0$\times$ at 16K to 1.56$\times$ at 128K.
The quality gap widens with length: at $\tau{=}0.01$, RefineKV matches Full Attention up to 32K, is within noise at 64K, and loses 3.08 points at 128K.

\paragraph{Limited speedup at short lengths.}
The decode speedup is small at 16K--32K (1.01--1.07$\times$ at $\tau{=}0.01$) and becomes substantial only from 64K onward.
RefineKV reduces only the attention term of a decode step, while the remaining kernels, mainly the weight GEMVs, cost about 12\,ms per step regardless of context length (Figure~5).
At 16K, attention therefore accounts for only about 2\,ms of Full Attention's 14.46\,ms TPOT, which caps the end-to-end speedup of any attention-side method at roughly 1.2$\times$; in addition, more than half of the KV cache is still expanded at this length.
RefineKV targets long-context decoding, where attention dominates the cost. Current LLMs commonly support 128K to 1M-token contexts, which fall within the regime RefineKV targets.

\subsection{Paired results with full attention}
At $\tau{=}0.01$, the CI of $\Delta$ contains zero on four of six panels.
The two exceptions are Qwen LongBench, whose CI barely excludes zero ($-0.11$ points), and Llama RULER ($-3.17$), whose loss is concentrated in two tasks.
\begin{table}[ht]
\centering
\caption{RefineKV versus Full Attention, paired on identical samples.
$\Delta$ is RefineKV minus Full Attention, with its 95\% bootstrap CI.}
\label{tab:paired}
\small
\setlength{\tabcolsep}{3pt}
\begin{tabular}{ll c cc cc}
\toprule
 & & & \multicolumn{2}{c}{$\tau{=}0.01$} & \multicolumn{2}{c}{$\tau{=}0.1$} \\
\cmidrule(lr){4-5}\cmidrule(lr){6-7}
Model & Benchmark & Full & Acc. & $\Delta$ [95\% CI] & Acc. & $\Delta$ [95\% CI] \\
\midrule
\multirow{3}{*}{Llama-3.1-8B}
 & RULER     & 73.37 & 70.19 & $-3.17$ [$-5.44$, $-1.02$] & 62.58 & $-10.79$ [$-13.67$, $-7.90$] \\
 & HELMET    & 69.10 & 68.43 & $-0.67$ [$-2.16$, $+0.77$] & 56.80 & $-12.29$ [$-15.34$, $-9.43$] \\
 & LongBench & 48.86 & 48.87 & $+0.01$ [$-0.11$, $+0.13$] & 48.95 & $+0.09$ [$-0.15$, $+0.35$]   \\
\midrule
\multirow{3}{*}{Qwen2.5-7B}
 & RULER     & 65.88 & 65.28 & $-0.60$ [$-2.23$, $+1.07$] & 60.35 & $-5.53$ [$-8.66$, $-2.47$]   \\
 & HELMET    & 54.57 & 54.48 & $-0.09$ [$-1.68$, $+1.52$] & 52.83 & $-1.74$ [$-3.86$, $+0.39$]   \\
 & LongBench & 49.34 & 49.23 & $-0.11$ [$-0.23$, $-0.00$] & 49.13 & $-0.21$ [$-0.39$, $-0.04$]   \\
\bottomrule
\end{tabular}
\end{table}

\end{document}